\PassOptionsToPackage{xcdraw,dvipsnames,table}{xcolor}
\documentclass[11pt]{article}

\usepackage[preprint]{acl}

\usepackage{times}
\usepackage{latexsym}
\usepackage[T1]{fontenc}
\usepackage[utf8]{inputenc}
\usepackage{microtype}
\usepackage{inconsolata}
\usepackage{graphicx}
\usepackage{amsmath}
\usepackage{amssymb}
\usepackage{booktabs}
\usepackage{multirow} 
\usepackage{tabularx}
\usepackage{array}
\usepackage{stfloats}
\usepackage{tcolorbox}
\usepackage{fvextra}
\usepackage{subcaption}
\usepackage{ragged2e}
\usepackage{tcolorbox}
\usepackage{hyperref}
\usepackage{fontawesome5}
\usepackage[export]{adjustbox}

\title{MM-JudgeBias: A Benchmark for Evaluating Compositional Biases \\ in MLLM-as-a-Judge}

\author{
 \textbf{Sua Lee\textsuperscript{1*}},
 \textbf{Sanghee Park\textsuperscript{2,3}},
  \textbf{Jinbae Im\textsuperscript{2,3*$\dagger$}}
\\
 \textsuperscript{1}Seoul National University,
 \textsuperscript{2}NAVER Cloud AI,
 \textsuperscript{3}KAIST AI
}

\begin{document}
\maketitle

\renewcommand{\thefootnote}{\fnsymbol{footnote}}
\footnotetext[1]{Equal contribution.}
\footnotetext[2]{Corresponding Author: \href{jinbae.im@navercorp.com.}{jinbae.im@navercorp.com.}}

\begin{abstract}
Multimodal Large Language Models (MLLMs) have been increasingly used as automatic evaluators—a paradigm known as \textit{MLLM-as-a-Judge}. 
However, their reliability and vulnerabilities to biases remain underexplored. 
We find that many MLLM judges fail to reliably integrate key visual or textual cues, yielding unreliable evaluations when evidence is missing or mismatched, and exhibiting instability under semantically irrelevant perturbations.
To address this, we systematically define \textit{Compositional Bias} in MLLM-as-a-Judge systems and introduce \textbf{MM-JudgeBias}, a benchmark for evaluating it.
MM-JudgeBias introduces controlled perturbations across Query, Image, and Response, and evaluates model behavior via two complementary metrics: \textit{Bias-Deviation (BD)} for sensitivity and \textit{Bias-Conformity (BC)} for stability. 
Our dataset of over 1,800 curated and refined multimodal samples, drawn from 29 source benchmarks, enables a fine-grained diagnosis of nine bias types across diverse tasks and domains.
Experiments on 26 state-of-the-art MLLMs reveal systematic modality neglect and asymmetric evaluation tendencies, underscoring the need for more reliable judges.

\end{abstract}

\definecolor{linkblue}{RGB}{0,92,175}
\definecolor{softmagenta}{RGB}{170,60,120}

\begin{center}
    \vspace{-0.8em}
    {\small
    \href{https://mm-judgebias.github.io/}{\textcolor{softmagenta}{\faGlobe\hspace{0.3em}Project Page}}
    \quad|\quad
    \href{https://github.com/naver-ai/MM-JudgeBias}{\textcolor{softmagenta}{\faGithub\hspace{0.3em}Code}}
    \quad|\quad
    \href{https://huggingface.co/datasets/naver-ai/MM-JudgeBias}{\textcolor{softmagenta}{\faDatabase\hspace{0.3em}Dataset}}
    }
    \vspace{0.2em}
\end{center}

\begin{figure}[t!]
  \centering
  \includegraphics[width=\columnwidth]{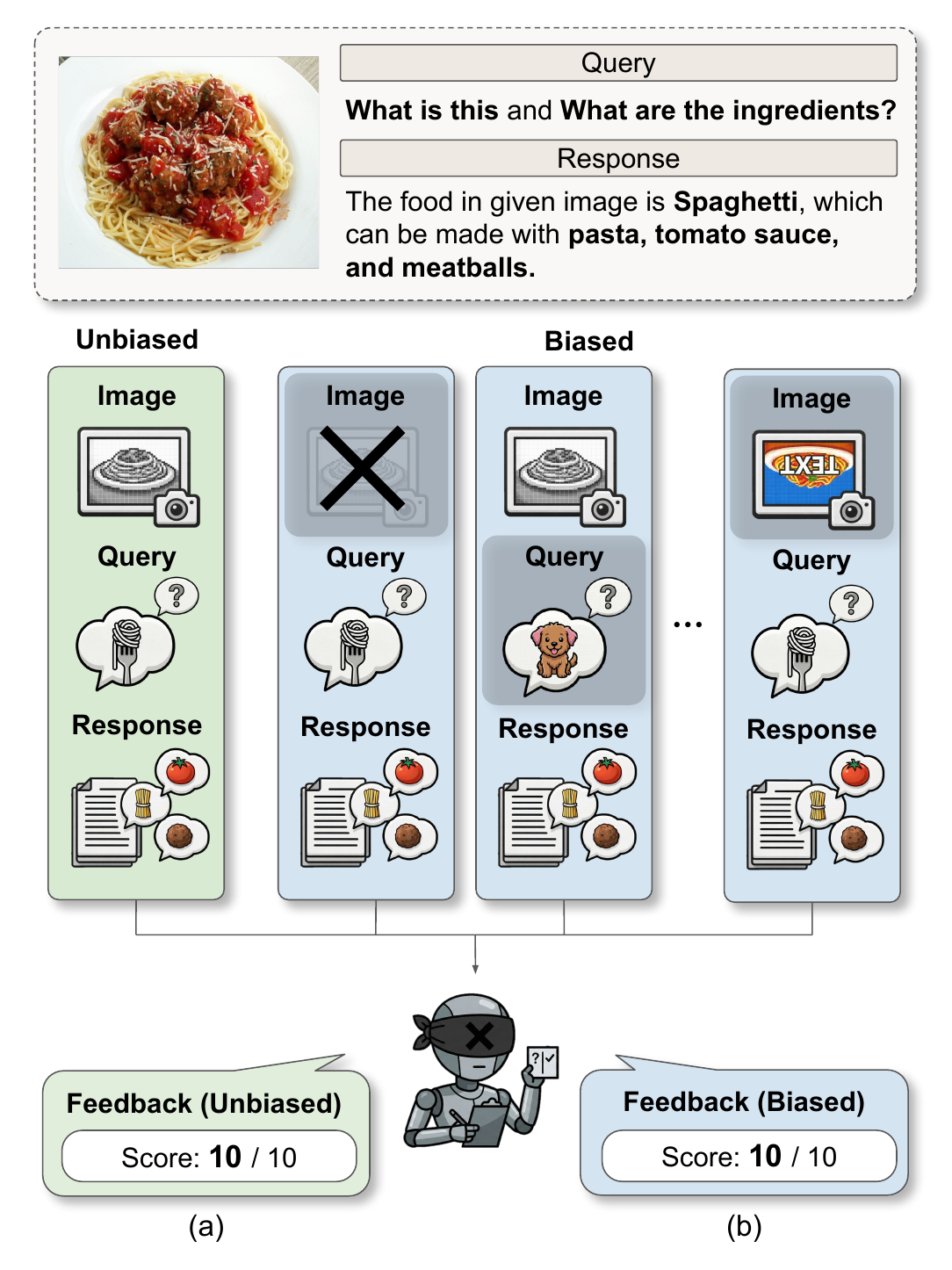}
  \caption{\textbf{Illustration of the Compositional Bias.} (a) Unbiased: A valid evaluation necessitates joint reasoning across all components (Image, Query, and Response); for example, answering \textit{``What is this?''} inherently requires the judge to verify the response against the specific image provided. (b) Biased: Scenarios where essential grounding evidence such as the image or the original query is removed or replaced with irrelevant content. 
  We observe that the \textit{Compositional Bias} is prevalent in many MLLMs, leading them to assign a near-perfect score to biased scenarios despite the fundamental lack of context or semantic misalignment. (See Appendix~\ref{appendix:judgment-examples} for actual examples.)
  }
  \label{fig:introduction}
  \vspace{-4mm}
\end{figure}

\section{Introduction}

Multimodal Large Language Models (MLLMs) ~\citep{caffagni2024revolution, achiam2023gpt, comanici2025gemini, bai2025qwen2} have recently demonstrated remarkable capabilities in integrating vision and language, enabling complex perception and reasoning across modalities.
Beyond serving as task solvers, MLLMs have increasingly been used as automatic judges—a paradigm known as \textit{MLLM-as-a-Judge}—to assess multimodal generations such as captioning, visual question answering, and visual reasoning~\citep{chen2024mllm, wang2025judge, li2024llms}.
While early studies~\citep{zhang2023gpt} have primarily focused on employing generalist MLLMs as general-purpose judges, recent efforts have shifted toward developing judge-specialized (critic) models~\citep{lee-etal-2024-prometheus, Xiong_2025_CVPR}, which are fine-tuned to provide feedback and judgments.
However, despite substantial progress in benchmarking MLLMs in judging scenarios~\cite{chen2024mllm, wang2025judge, zeng-etal-2025-mm, Li_2025_CVPR, Ruan_2025_ICCV},
the reliability and fundamental biases of MLLM-based judges remain underexplored, in contrast to the extensive studies on bias in LLM-as-a-Judge~\cite{ye2024justice, shi2024judging, li2024llms}.

Recent studies have increasingly revealed that MLLMs often struggle to balance and integrate information across modalities, primarily exhibiting ``modality biases'' where models rely on spurious textual correlations to reach a correct answer~\citep{chen2024quantifying, zhao-etal-2025-looking, zheng2025mllms, lin2025unveiling, zheng2025unveiling}. 
While these biases are typically viewed as limitations in task-solving accuracy, we argue that they evolve into a more insidious failure of verification integrity within the MLLM-as-a-Judge paradigm. 
Unlike a task-solver, the primary mandate of a judge is not mere probabilistic prediction, but the conditional verification of alignment between textual queries, visual evidence, and candidate responses. 

We identify that judgment failures go beyond modality bias and stem from \textbf{Compositional Bias}: a systematic tendency where a judge fails to correctly integrate and reason over multiple components (query, image, and response), instead relying on partial, misaligned, or spurious compositions, as illustrated in Figure~\ref{fig:introduction}.
Compositional bias does not merely reflect incorrect guessing by the judge; rather, it indicates situations where the judge actively endorses erroneous or ungrounded responses by failing to properly ground its judgment in the provided context, compromising the reliability of the entire evaluation pipeline~\citep{xiong2025multi, laskar-etal-2025-judging}.
This motivates the need for a systematic benchmark that explicitly measures how reliably MLLM judges attend to and integrate visual, textual, and response information as a judge.

To address this, we introduce \textbf{MM-JudgeBias}, which stands for \textbf{M}LL\textbf{M}-as-a\textbf{-Judge} \textbf{Bias} Evaluation Benchmark, a comprehensive benchmark designed to measure the compositional biases in MLLM-as-a-Judge.
We define a taxonomy of nine bias types across three critical dimensions: 
(1) \textit{Integrality}: Does the judge consider all components holistically?
(2) \textit{Congruity}: Does the judge detect and appropriately address semantic contradictions between components?
(3) \textit{Robustness}: Does the judge maintain a stable judgment despite semantic-preserving variations?
Motivated by~\citet{ye2024justice}, our framework systematically injects controlled bias perturbations and quantifies the model's vulnerability to biases.
Specifically, we utilize two complementary metrics: \textit{Bias-Deviation (BD)} for sensitivity to disruptions in Integrality and Congruity, and \textit{Bias-Conformity (BC)} for stability in Robustness.

To encompass a wide variety of tasks and domains, we structure our benchmark around two axes: four task types to assess different multimodal abilities, and 12 domain types to capture diverse visual contexts.
We curate source data from 29 benchmarks and refine them for the MLLM-as-a-Judge setting, resulting in a benchmark set of 1,804 samples that cover diverse query tasks and difficulty levels.
To comprehensively analyze biases in MLLM-as-a-Judge, we assess 26 MLLMs including closed-source, open-source, and critic models.
Our evaluation results demonstrate that even state-of-the-art models struggle to jointly consider image inputs and text queries.

In summary, this work makes the following three key contributions.
First, we demonstrate the emergence of the \textit{Compositional Bias} that arises from the interaction between components of the judgment, such as text queries, image inputs, and responses.
We also introduce \textit{MM-JudgeBias}, a systematic benchmark that evaluates the compositional biases in MLLM-as-a-Judge through nine bias types and two complementary metrics: Bias-Deviation (BD) and Bias-Conformity (BC). 
Finally, we perform an extensive empirical analysis of \textit{26 state-of-the-art MLLMs}, uncovering that the compositional bias is a systemic issue pervasive even in the most advanced reasoning-heavy models. 

\begin{figure*}[t]
    \centering
    \includegraphics[width=\textwidth]{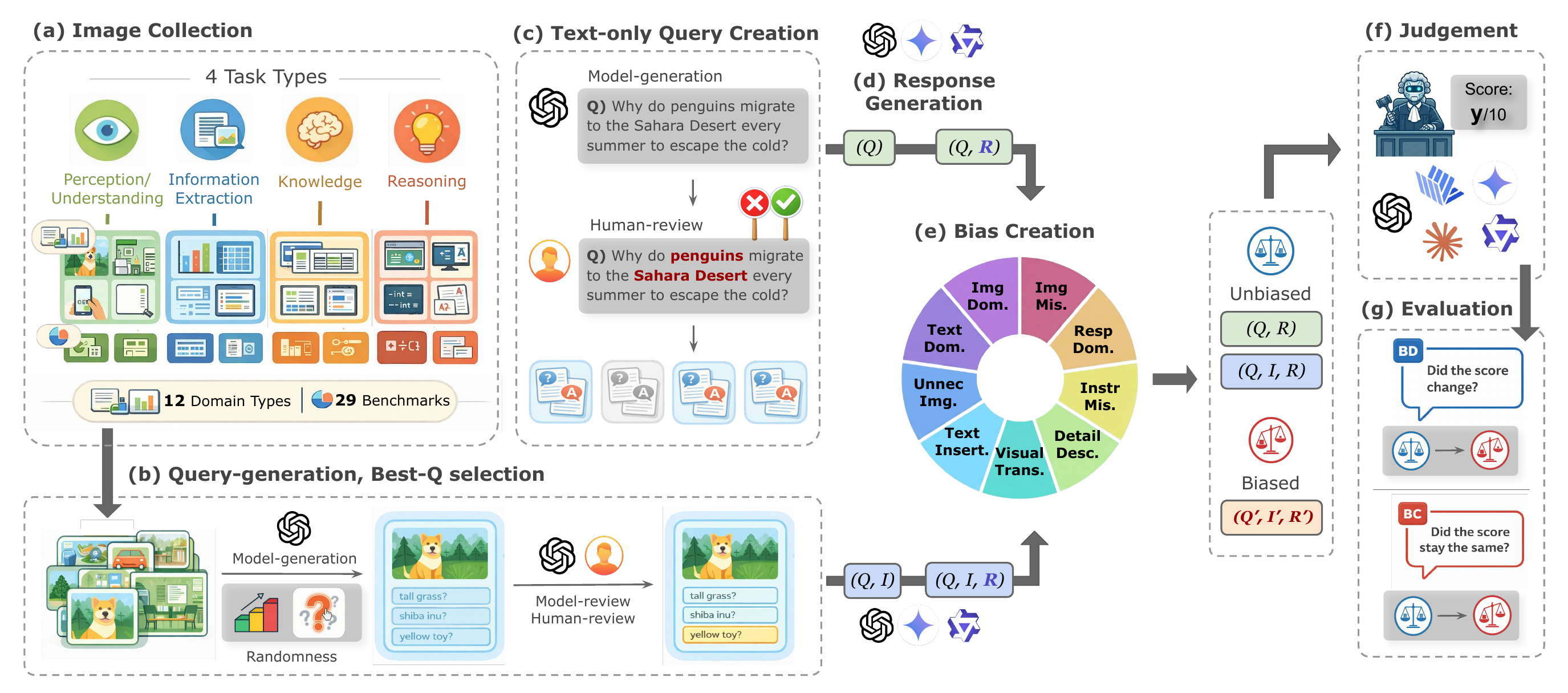}
    \caption{\textbf{Overview of the MM-JudgeBias construction and evaluation pipeline.} (a--b) We construct image sets from 29 source benchmarks covering four task types and 12 domain types, and then generate queries tailored to our bias evaluation setting through a comprehensive model-and-human verification framework to ensure high quality. (c) A text-only evaluation set is independently constructed via a parallel generation and verification process. (d--e) Based on nine defined bias types, we apply input perturbations to transform original unbiased samples—triplets $(Q, I, R)$ or text-only pairs $(Q, R)$—into their corresponding biased variants $(Q', I', R')$. (f--g) Finally, we evaluate 26 MLLM judges using our proposed Bias-Deviation (BD) and Bias-Conformity (BC) metrics to systematically quantify the impact of Compositional Bias on their judgment reliability.}
    \label{fig:wide}
\end{figure*}

\section{Related Work}

\subsection{Bias and Reliability in LLM-as-a-Judge}
The LLM-as-a-Judge framework has become central to automatic evaluation in generation tasks \citep{zheng2023judging, gu2024survey, li2024llms}.
While early benchmarks showed high agreement with human judgments, subsequent analyses revealed systematic biases that undermine reliability.
Recent studies demonstrate that LLM judges are influenced by superficial cues such as position, verbosity, and stylistic patterns, rather than actual content quality \citep{ye2024justice, shi2024judging}.
\citet{ye2024justice} identified 12 distinct bias types in large-scale evaluation; \citet{shi2024judging} and~\citet{li2025evaluating} showed that prompt order and rubric phrasing significantly alter scores.
Bias also arises from self-similarity between judge and generator models (preference leakage) and model-specific favoritism~\citep{li2025preference, spiliopoulou2025play}.
Cross-lingual and domain analyses confirm that such biases persist even under controlled conditions~\citep{chen2024humans, gao2025evaluating}.
Collectively, these findings indicate that detecting and quantifying bias is crucial to maintain credibility in automated evaluation.

However, almost all prior studies remain limited to text-only scenarios.
Existing approaches measure linguistic robustness but overlook multimodal evaluation, where visual and textual cues jointly determine correctness.
This gap motivates the need to extend bias analysis into multimodal judging, where biases may manifest through modality dependence or neglect.

\subsection{Bias and Reliability in MLLM-as-a-Judge}
Recent work extends evaluation to MLLMs, aiming to assess their judgment capability and reliability on multimodal contexts~\citep{chen2024mllm, wang2025judge, laskar-etal-2025-judging, zeng-etal-2025-mm, xiong2025multi, Li_2025_CVPR, yasunaga2025multimodal, Ruan_2025_ICCV}.
However, biases in MLLM-as-a-Judge remain underexplored.
\citet{chen2024mllm} analyzed egocentric, position, and length biases in MLLMs; however, such biases have been examined in LLM-as-a-Judge, leaving unique biases arising in multimodal contexts unaddressed.
\citet{hwang2025fooling} defined several visual biases and evaluated the vulnerability of MLLM judges to the biases; however, their scope was limited to biases related to visual transformations in text-to-image evaluation.
In this work, we introduce compositional biases in general multimodal judgment settings from the perspective of integrality, congruity, and robustness, and propose two metrics to systematically measure these biases.
Further details on the positioning of our work are presented in Appendix~\ref{appendix:novelty_positioning}.

\section{MM-JudgeBias: MLLM-as-a-Judge Bias Evaluation Benchmark}

We introduce MM-JudgeBias, a benchmark built to quantify vulnerabilities to compositional biases in MLLM-as-a-Judge.
MM-JudgeBias is designed around the central hypothesis that a reliable judge should attend to and integrate all available components---\textit{Query, Image, and Response}---when producing its judgment.
In this section, we introduce task formulation (\ref{subsection:task-formulation}), evaluation metrics (\ref{subsection:evaluation-metrics}), bias taxonomy (\ref{subsection:bias-taxonomy}), and data construction pipeline (\ref{subsection:data-construction}) of MM-JudgeBias.

\begin{table*}[t]
\centering
\small
\renewcommand{\arraystretch}{1.3}
\begin{tabular}{m{0.5cm}m{2.8cm}m{5.2cm}m{4.3cm}m{1.0cm}}
\toprule
 & \textbf{Bias Type} & \textbf{Targeted Reliability (Purpose)} & \textbf{Perturbation Strategy} & \textbf{Metric} \\
\midrule

\parbox[t]{2mm}{\multirow{6}{*}{\rotatebox[origin=c]{90}{\textbf{Integrality}}}} 
& \cellcolor{gray!12} \textbf{Text-Dominance} & 
\cellcolor{gray!12} Assesses over-reliance on linguistic cues when visual grounding is absent. & 
\cellcolor{gray!12} Replaces the image with a null image (black image). & 
\cellcolor{gray!12} BD \\

& \textbf{Image-Dominance} & 
Tests if the judge ignores the original query in favor of visual content. & 
Replaces the query with a null text (\texttt{``''}). & 
BD \\

& \cellcolor{gray!12} \textbf{Response-Dominance} & 
\cellcolor{gray!12} Evaluates if scores are assigned based on response fluency alone, ignoring all context. & 
\cellcolor{gray!12} Replaces both the query and image with null inputs (``'' and a black image).& 
\cellcolor{gray!12} BD \\

\midrule

\parbox[t]{2mm}{\multirow{4}{*}{\rotatebox[origin=c]{90}{\textbf{Congruity}}}} 
& \textbf{Instruction-Misalignment} & 
Probes query-response conflict by checking for semantic mismatch between the query and response. & 
Replaces the query with a random, unrelated sample. & 
BD \\

& \cellcolor{gray!12} \textbf{Image-Misalignment} & 
\cellcolor{gray!12} Probes image-response conflict by checking for semantic mismatch between the image and response. & 
\cellcolor{gray!12} Replaces the image with a random, unrelated sample. & 
\cellcolor{gray!12} BD \\

\midrule

\parbox[t]{2mm}{\multirow{8}{*}{\rotatebox[origin=c]{90}{\textbf{Robustness}}}} 
& \textbf{Detail-Description} & 
Examines if providing redundant, aligned visual information inflates the score. & 
Appends a detailed caption of the given image to the query. & 
BC \\

& \cellcolor{gray!12} \textbf{Unnecessary-Image} & 
\cellcolor{gray!12} Tests the ability to recognize and ignore irrelevant visual inputs in text-only tasks. & 
\cellcolor{gray!12} Adds a random, unrelated image to a text-only task. & 
\cellcolor{gray!12} BC \\

& \textbf{Visual-Transformation} & 
Measures invariance to low-level visual transformations that preserve semantics. & 
Applies a diverse set of semantic-preserving augmentations to the image. & 
BC \\

& \cellcolor{gray!12} \textbf{Texture-Insertion} & 
\cellcolor{gray!12} Probes over-sensitivity to textual cues embedded directly within the visual modality. & 
\cellcolor{gray!12} Overlays query-related keywords or the query text itself onto the image. & 
\cellcolor{gray!12} BC \\

\bottomrule
\end{tabular}
\caption{
\textbf{The MM-JudgeBias Taxonomy.} Biases are grouped into three dimensions: \textit{Integrality} and \textit{Congruity} evaluate the model's sensitivity to semantic disruptions, while \textit{Robustness} assesses judgment stability against semantic-preserving perturbations. (See Appendix~\ref{appendix:bias-examples} and~\ref{appendix:bias_perturbation} for detailed visual examples and perturbation strategies).
}
\label{tab:bias-taxonomy}
\end{table*}

\subsection{Task Formulation}
\label{subsection:task-formulation}

Let a judgment instance be a multimodal triplet $(Q, I, R)$, where $Q$ denotes the text-based query (used interchangeably with question or instruction), $I$ represents the associated visual context (image), and $R$ is the model-generated response being evaluated. 
An MLLM-as-a-Judge model, denoted as $f_\theta$, serves as a scoring function:
\[
f_\theta : (Q, I, R) \rightarrow y, \quad y \in \{1, 2, \dots, S\},
\]
where $y$ is a scalar score representing the quality of $R$ given $Q$ and $I$, and $S$ is the maximum scale (e.g., $S=10$). 
We adopt the 10-point scoring convention following recent open-ended evaluation protocols \citep{liu2023visual, yu2023mm, chen2024mllm}. We provide detailed prompts in Appendix~\ref{appendix:judgement_prompts}.

To evaluate the susceptibility of $f_\theta$ to biases, MM-JudgeBias pairs each unbiased triplet $(Q, I, R)$ with a corresponding biased triplet $(Q', I', R')$. 
The biased triplet is generated by injecting a controlled perturbation into one or more components to evaluate whether the model's judgments arise from a holistic integration of $Q$, $I$, and $R$, rather than from biased or partial consideration of these factors.

The judge model thus produces two comparative scores: the reference score $y = f_\theta(Q, I, R)$ and the perturbed score $\hat{y} = f_\theta(Q', I', R')$. 
We categorize the expected behavior of a reliable judge into two paradigms:
\begin{enumerate}
    \item \textbf{Sensitivity to Semantic Distortion (Integrality \& Congruity):} When a perturbation disrupts the essential semantic alignment, such as modality removal (integrality) or inter-modal mismatching (congruity), a reliable judge should reflect this deficiency by decreasing the score ($\hat{y} < y$).
    \item \textbf{Invariance to Semantic-Preserving Perturbations (Robustness):} When a perturbation introduces changes that do not affect the objective quality of the response such as minor visual transformations or redundant captions, the score should remain stable ($\hat{y} \approx y$).
\end{enumerate}
Based on these principles, we develop quantitative measures to evaluate the degree of score deviation and stability across different bias categories.

\subsection{Evaluation Metrics}
\label{subsection:evaluation-metrics}

To effectively measure robustness under biased conditions, we propose two complementary metrics, \textit{Bias-Deviation (BD)} and \textit{Bias-Conformity (BC)}, which quantify the model's behavioral response to semantic distortions and semantic-preserving perturbations, respectively.

\paragraph{Bias-Deviation (BD).}

The BD metric is designed to assess whether a judge responds appropriately to semantic distortions (e.g., modality removal or misalignment) by lowering its judgment score.
For a dataset $\mathcal{D}$ consisting of $(y, \hat{y})$ pairs, BD is defined as follows:
\[
\text{BD} = \mathbb{E}_{(y,\hat{y}) \sim \mathcal{D}} \left[ \frac{(y - \hat{y})_+}{y-1} \;\middle|\; y > 1 \right],
\]
where the positive deviation between $y$ and $\hat{y}$ is normalized by the maximum positive deviation. Note that samples with $y = 1$ are excluded from the BD computation, as no further score reduction is possible.
A higher BD indicates that the model is sensitive to the injected bias and appropriately adjusts its judgment. For bias types associated with Integrality and Congruity, we use BD to measure the reliability of MLLM judges.

\paragraph{Bias-Conformity (BC).}
In contrast, the BC metric is devised to evaluate the stability of the model's judgment against perturbations that should not affect the objective quality of the response (e.g., visual style changes or redundant captions). We define BC as the degree of score invariance:
\[
\text{BC} = \mathbb{E}_{(y,\hat{y}) \sim \mathcal{D}} \left[1 - \frac{|y - \hat{y}|}{\text{max}(y-1, S-y)} \right],
\]
where the absolute difference between $y$ and $\hat{y}$ is normalized by the upper bound on the score change. 
A high BC score signifies that the model is robust to irrelevant variations, maintaining a consistent evaluative standard. 
For bias types related to Robustness, an ideal evaluator should achieve a BC score near 1.0, indicating that its judgment is not swayed by superficial or non-essential modifications.
As these perturbations are designed to be semantically invariant, the theoretical expectation for score deviation is minimal, inherently leading to a relatively high-value distribution of BC scores that characterizes a robust judge. 

\paragraph{Metric Selection and Interpretation.}
It is important to note that BD and BC are not competing metrics but are applied selectively based on the nature of the bias. We define a reliability score for a judge model $f_\theta$ by aggregating BD across semantic-disruptive categories and BC across semantic-preserving categories. This dual-lens approach allows us to distinguish between a model that is simply ``volatile'' (high deviation in all cases) and one that is ``discerning'' (high deviation only when semantically justified).

\subsection{Bias Taxonomy}
\label{subsection:bias-taxonomy}

To evaluate the multifaceted reliability of MLLM-as-a-Judge, we define nine bias types categorized into three functional dimensions based on the expected behavior of a reliable judge: \textit{Integrality}, \textit{Congruity}, and \textit{Robustness}. Table~\ref{tab:bias-taxonomy} provides a detailed overview of the taxonomy, including the perturbation strategies and the metrics.

\textbf{Integrality-targeted Biases} test the model's awareness of modality necessity—whether it recognizes that each input is strictly required for evaluation. We exclude modalities to probe modality dominance such as \textit{Text-}, \textit{Image-}, and \textit{Response-Dominance}. A reliable judge is expected to penalize these information-deficient cases (high BD).

\textbf{Congruity-targeted Biases} probe cross-modal reasoning by checking whether the judge detects semantic contradictions and appropriately reflects them in its judgment. This involves cases where the query or image is mismatched with the response (\textit{Instruction/Image-Misalignment}). Similar to Integrality, a logical judge should decrease its score when encountering such conflicts (high BD).

\textbf{Robustness-targeted Biases} assess the model's stability against perturbations that do not alter the objective quality of a response. These involve redundant information (\textit{Detail-Desc.}), irrelevant visual inputs (\textit{Unnecessary-Img.}), or superficial variations (\textit{Visual-Trans.}, \textit{Texture-Insertion}). An ideal judge should maintain scoring stability regardless of these changes (high BC).

\begin{table*}[t]
\centering
\small
\renewcommand{\arraystretch}{1.25}

\resizebox{\textwidth}{!}{%
\begin{tabular}{l|l|c|ccc|cc|cccc|c|cc}
\toprule
\textbf{Category} & \textbf{Model} & \textbf{Think} & 
\multicolumn{3}{c|}{\textbf{Integrality (BD$\uparrow$)}} & 
\multicolumn{2}{c|}{\textbf{Congruity (BD$\uparrow$)}} & 
\multicolumn{4}{c|}{\textbf{Robustness (BC$\uparrow$)}} & 
\textbf{Avg.} & 
\multicolumn{2}{c}{\textbf{Variance}} \\
& & & 
\rotatebox{90}{TextDom.} & \rotatebox{90}{ImgDom.} & \rotatebox{90}{RespDom.} & 
\rotatebox{90}{InstrMis.} & \rotatebox{90}{ImgMis.} & 
\rotatebox{90}{DetailDesc.} & \rotatebox{90}{UnnecImg.} & \rotatebox{90}{VisualTrans.} & \rotatebox{90}{TextInsert.} &  &
\rotatebox{90}{Inter-run} & \rotatebox{90}{Inter-sample} \\
\midrule

\multirow{14}{*}{\textbf{Closed-source}} 
& Gemini-3-Pro (high) & $\checkmark$ & \cellcolor{blue!15}\textbf{0.912} & 0.278 & \cellcolor{blue!15}\textbf{0.988} & 0.982 & 0.955 & \cellcolor{blue!15}\textbf{0.936} & 0.947 & 0.889 & 0.933 & \cellcolor{blue!15}\textbf{0.869} & 0.6 & 9.3\\
& Gemini-2.5-Pro & $\checkmark$ & 0.751 & 0.535 & 0.978 & \cellcolor{blue!15}\textbf{1.0} & 0.904 & 0.935 & 0.924 & 0.861 & 0.930 & \cellcolor{blue!15}\textbf{0.869} & 0.5 & 8.7 \\
& Gemini-2.5-Flash & $\checkmark$ & 0.423 & 0.292 & 0.743 & 0.997 & 0.760 & 0.905 & 0.944 & 0.864 & 0.918 & 0.761 & 0.6 & 7.0\\
& Gemini-2.5-Flash-Lite & $\checkmark$ & 0.391 & 0.335 & 0.890 & 0.993 & 0.634 & 0.866 & 0.940 & 0.801 & 0.875 & 0.747 & 1.1 & 7.9\\
& Gemini-2.5-Flash-Lite &  & 0.113 & 0.367 & 0.544 & 0.978 & 0.585 & 0.845 & 0.937 & 0.787 & 0.873 & 0.670 & 1.3 & 6.5 \\
& Gemini-2.0-Flash-Lite &  & 0.162 & 0.358 & 0.570 & 0.924 & 0.217 & 0.897 & 0.958 & 0.862 & 0.878 & 0.647 & 0.5 & 4.5 \\
& o3 (high) & $\checkmark$ & 0.276 & 0.354 & 0.596 & 0.986 & 0.337 & 0.879 & 0.937 & 0.822 & 0.884 & 0.675 & 0.7 & 6.3 \\
& o4-mini (high) & $\checkmark$ & 0.141 & 0.446 & 0.518 & 0.997 & 0.184 & 0.886 & 0.950 & 0.854 & 0.909 & 0.654 & 0.9 & 8.5 \\
& GPT-5.1 (high) & $\checkmark$ & 0.192 & 0.211 & 0.200 & \cellcolor{blue!15}\textbf{1.0} & 0.296 & 0.911 & 0.961 & 0.867 & 0.908 & 0.616 & 0.6 & 7.3 \\
& GPT-5 mini (high) & $\checkmark$ & 0.112 & 0.236 & 0.337 & 0.991 & 0.185 & 0.898 & 0.954 & 0.877 & 0.927 & 0.613 & 0.4 & 5.8 \\
& GPT-4.1 mini &  & 0.049 & 0.148 & 0.138 & 0.992 & 0.160 & 0.882 & 0.972 & 0.887 & 0.904 & 0.570 & 0.5 & 4.0 \\
& Claude-Opus-4.5 & $\checkmark$ & 0.680 & \cellcolor{blue!15}\textbf{0.589} & 0.917 & 0.987 & 0.959 & 0.917 & 0.925 & 0.842 & 0.903 & 0.858 & 0.2 & 5.3 \\
& Claude-Sonnet-4.5 & $\checkmark$ & 0.556 & 0.540 & 0.831 & 0.987 & 0.905 & 0.884 & 0.921 & 0.825 & 0.893 & 0.816 & 0.5 & 5.6 \\
& Claude-Haiku-4.5 & $\checkmark$ & 0.291 & 0.530 & 0.735 & 0.981 & 0.809 & 0.870 & 0.943 & 0.785 & 0.869 & 0.757 & 0.6 & 5.7 \\
& Claude-Opus-4.5 &  & 0.539 & 0.571 & 0.838 & 0.990 & \cellcolor{blue!15}\textbf{0.973} & 0.902 & 0.814 & 0.820 & 0.891 & 0.815 & 0.2 & 5.5 \\
& Claude-Sonnet-4.5 &  & 0.445 & 0.385 & 0.716 & 0.982 & 0.862 & 0.837 & 0.834 & 0.798 & 0.866 & 0.747 & 0.5 & 6.3 \\
& Claude-Haiku-4.5 &  & 0.278 & 0.370 & 0.603 & 0.981 & 0.818 & 0.827 & 0.888 & 0.747 & 0.844 & 0.706 & 0.7 & 6.3 \\

\midrule
\multirow{7}{*}{\textbf{Open-source}} 
& Qwen3-VL-30B-A3B-Thinking & $\checkmark$ & 0.293 & 0.343 & 0.806 & 0.983 & 0.476 & 0.886 & 0.905 & 0.850 & 0.874 & 0.713 & 0.9 & 6.7 \\
& Qwen3-VL-8B-Thinking & $\checkmark$ & 0.177 & 0.210 & 0.412 & 0.991 & 0.529 & 0.883 & 0.927 & 0.865 & 0.898 & 0.655 & 0.9 & 6.2 \\
& Qwen3-VL-30B-A3B-Instruct &  & 0.237 & 0.174 & 0.642 & 0.988 & 0.648 & 0.840 & 0.837 & 0.854 & 0.883 & 0.678 & 0.6 & 4.6 \\
& Qwen3-VL-8B-Instruct &  & 0.336 & 0.266 & 0.913 & \cellcolor{blue!15}\textbf{1.0} & 0.622 & 0.803 & 0.952 & 0.870 & 0.880 & 0.738 & 0.8 & 7.3\\
& Qwen2.5-VL-72B-Instruct &  & 0.082 & 0.158 & 0.223 & 0.989 & 0.208 & 0.822 & 0.903 & 0.855 & 0.853 & 0.566 & 0.6 & 2.3 \\
& Qwen2.5-VL-7B-Instruct &  & 0.141 & 0.188 & 0.194 & 0.991 & 0.277 & 0.735 & 0.815 & 0.739 & 0.767 & 0.539 & 1.7 & 4.6 \\
& InternVL3.5-30B-A3B &  & 0.073 & 0.225 & 0.179 & 0.964 & 0.377 & 0.800 & 0.837 & 0.801 & 0.804 & 0.562 & 1.4 & 4.1 \\
& InternVL3.5-14B &  & 0.137 & 0.243 & 0.273 & 0.982 & 0.464 & 0.803 & 0.850 & 0.797 & 0.814 & 0.596 & 1.0 & 3.7 \\
& InternVL3.5-8B &  & 0.099 & 0.178 & 0.215 & 0.926 & 0.289 & 0.798 & 0.886 & 0.791 & 0.814 & 0.555 & 1.0 & 3.5 \\

\midrule
\multirow{4}{*}{\textbf{Critic}} 
& Prometheus-Vision-13B &  & 0.163 & 0.340 & 0.362 & 0.890 & 0.166 & 0.738 & 0.781 & 0.804 & 0.818 & 0.563 & 2.4 & 10.3 \\
& Prometheus-Vision-7B &  & 0.167 & 0.242 & 0.246 & 0.869 & 0.165 & 0.750 & 0.793 & 0.821 & 0.806 & 0.540 & 2.4 & 10.6 \\
& LLaVA-Critic-72B &  & 0.147 & 0.121 & 0.373 & 0.989 & 0.250 & 0.926 & \cellcolor{blue!15}\textbf{0.974} & \cellcolor{blue!15}\textbf{0.931} & \cellcolor{blue!15}\textbf{0.942} & 0.628 & 0.0 & 2.6 \\
& LLaVA-Critic-7B &  & 0.238 & 0.266 & 0.420 & 0.958 & 0.452 & 0.824 & 0.929 & 0.869 & 0.864 & 0.647 & 0.0 & 6.1 \\

\midrule
\multicolumn{3}{r|}{\textbf{Avg.}} 
& 0.287 & 0.317 & 0.547 & 0.976 & 0.516 & 0.856 & 0.905 & 0.835 & 0.874 & 0.679 & 0.8 & 6.1 \\

\bottomrule
\end{tabular}
}
\caption{\textbf{Main experimental results on MM-JudgeBias.} Biases are categorized into three dimensions: \textbf{Integrality}, \textbf{Congruity}, and \textbf{Robustness}. Higher BD in Integrality and Congruity indicates better sensitivity to semantic disruptions, while higher BC in Robustness reflects superior judgment stability. The `Avg.' column shows the average reliability scores aggregated over all bias types. We also include an `Avg.' row reporting the mean across all models to show overall trends across bias types. Models using internal reasoning are checked in the `Think' column, and (high) indicates the reasoning effort. Given the intrinsic consistency issue of MLLMs, three experimental runs are performed. We also report variance across runs and samples as supporting information. In particular, inter-sample variance is added to address that models producing identical scores for all samples can easily achieve a BC score of 1.0.}
\label{tab:main-results}
\end{table*}

\subsection{Data Construction}
\label{subsection:data-construction}

The construction of \textit{MM-JudgeBias} follows a systematic pipeline designed to ensure task diversity, multimodal dependency, and rigorous quality control, as shown in Figure~\ref{fig:wide}. 
We consolidate the process into four primary phases, yielding a final dataset of 1,804 high-difficulty evaluation pairs that require sophisticated multimodal reasoning. Detailed statistics are summarized in Appendix~\ref{appendix:data_statistics}.

\paragraph{Taxonomy Definition and Source Data Selection.} 
To ensure a comprehensive evaluation of multimodal judgment, we define a taxonomy consisting of four core cognitive tasks and 12 domains.
From the taxonomy, we sample base instances from 29 diverse representative benchmarks. 
To ensure data quality and maintain a balanced distribution, we apply oversampling followed by filtering.
A detailed list is provided in Appendix~\ref{appendix:task_domain_types}.

\paragraph{Human-in-the-loop Data Synthesis.} 

Although source benchmarks provide instruction queries, we synthesize new queries to ensure that evaluation requires grounding in all relevant modalities and remains sufficiently challenging for MLLM judges.
For each instance, we employ Gemini-2.5-Pro~\cite{comanici2025gemini} to generate three complex queries. 
To guarantee the multimodal dependency and quality of the synthetic queries, they are filtered and revised via a meticulously designed human-in-the-loop pipeline.
Furthermore, we separately create text-only queries to construct data appropriate for the unnecessary-image bias type.
Detailed information about the data synthesis is provided in Appendix~\ref{appendix:data_synthesis}.

\paragraph{Multimodal Triplet Synthesis.} 

For each qualified data instance, we synthesize a reference triplet $(Q, I, R)$.
We generate a response $R$ conditioned on $Q$ and $I$, which serves as the target output for evaluation. 
To ensure diversity in judgment scores, we employ GPT-5 mini~\citep{gpt5}, Gemini-2.0-Flash-Lite, and Qwen2.5-VL-7B-Instruct~\citep{bai2025qwen2} as response generation models in a balanced manner.

\paragraph{Systematic Bias Perturbation.} 
Starting from the unbiased triplets, we apply the nine perturbation strategies defined in Section~\ref{subsection:bias-taxonomy} to generate biased counterparts $(Q', I', R')$. 
Each bias type is generated using its corresponding perturbation strategy: visual transformations are implemented with predefined transform pipeline combinations, while texture modifications are model-generated.
For each biased and unbiased triplet, MLLMs perform judgment, and we evaluate their reliability using measures tailored to each bias type.
Detailed perturbation strategies are provided in Appendix~\ref{appendix:bias_perturbation}.

\section{Empirical Results and Analysis}

In this section, we present a comprehensive evaluation of state-of-the-art MLLMs as judges. 
We analyze their susceptibility to various biases and investigate whether advanced reasoning capabilities or specialized training for judgment can mitigate these issues.

\subsection{Experimental Setup}

\paragraph{Judge Models and Operational Modes.}
We evaluate 26 MLLMs as judges, categorized into: (1) Closed-source models (Gemini~\citep{comanici2025gemini}, GPT~\citep{o3, gpt5}, and Claude~\citep{sonnet4.5, opus4.5, haiku4.5} families), (2) Open-source models (Qwen-VL~\cite{Qwen3-VL, bai2025qwen2} and InternVL~\citep{wang2025internvl3} series), and (3) Specialized Critic models (Prometheus-Vision~\citep{lee-etal-2024-prometheus} and LLaVA-Critic~\cite{Xiong_2025_CVPR}). 
For models supporting extended reasoning (think), we evaluate them in both standard and reasoning modes. This allows us to assess whether internal deliberation improves cross-modal judgment.

\paragraph{Implementation Details.}

To fully leverage MLLMs' capabilities as judges, we set the \texttt{max\_tokens} parameter to 16384.
For reasoning models, we configure \texttt{reasoning\_effort} as ``high'' and \texttt{budget\_tokens} as 16384 for Claude families.
For other parameters such as \texttt{top\_p} and \texttt{temperature}, we use the models' default values.

\subsection{Main Results}

Table~\ref{tab:main-results} summarizes the reliability of all evaluated MLLMs on MM-JudgeBias. 
Although each model exhibits varying strengths and weaknesses across different bias types, the overall reliability---computed by aggregating BD or BC scores across bias types---generally aligns with the model's overall capability.
The internal reasoning process is found to generally improve reliability, with the Qwen3-VL-8B as the sole exception.
It appears to facilitate reliable judgments by thoroughly examining all components provided in the context.
Prometheus-Vision and LLaVA-Critic, fine-tuned for use as critics from LLaVA-1.5~\citep{Liu_2024_CVPR} and LLaVA-OneVision~\citep{li2024llava}, respectively, demonstrate reliability higher than the base models' expected performance.
Nonetheless, vulnerabilities to biases are not largely resolved through critic instruction-following datasets, indicating that the underlying capabilities of the base models remain fundamentally important.

Analysis by bias type reveals that models are especially susceptible to the three integrality-related bias types and the image-misalignment bias type.
Notably, substantial weaknesses are observed in response-dominance bias scenarios, despite the removal of both the image and the text query.
Gemini-3-Pro and Claude-Opus-4.5 exhibit notably strong performance on the response-dominance and image-misalignment bias types; however, o3, despite being a strong reasoning model, shows relatively low performance.
These observations imply that a model's training recipe can substantially affect the reliability of the judgments, even when this is not evident from its general capability.
Although Gemini-3-Pro performs reasonably well on the text-dominance bias, most models still rely on the context of either image or text alone.
While modality bias has been the subject of growing research, it appears particularly important to address modality bias in MLLM-as-a-Judge.

A category-level analysis further shows that integrality is particularly difficult for most models, with substantial performance variance across them.
In contrast, performance on congruity is relatively higher, indicating that judges are better at detecting semantic contradictions than at recognizing the absence of required information.
Although robustness is generally high across models, open-source and critic models are more vulnerable to robustness-related biases involving visual and textual noise.

Notably, model scale does not linearly correlate with compositional integrity; increasing parameters within the same family does not consistently improve BD/BC scores, reinforcing that judgment reliability is distinct from general reasoning capability. For instance, LLaVA-Critic-72B shows lower overall reliability than the 7B model, as its gains on robustness are outweighed by large drops on bias types to which most models are generally vulnerable, indicating potential trade-offs between bias types.
Finally, the weak correlation between inter-run and inter-sample variance suggests that execution-level stochasticity and sample-level differentiation are distinct phenomena.
The overall high variance further indicates that models are not operating in a trivial constant-scoring mode, but are genuinely struggling under biased conditions.

\begin{figure}[t]
    \centering
    \includegraphics[width=\columnwidth]{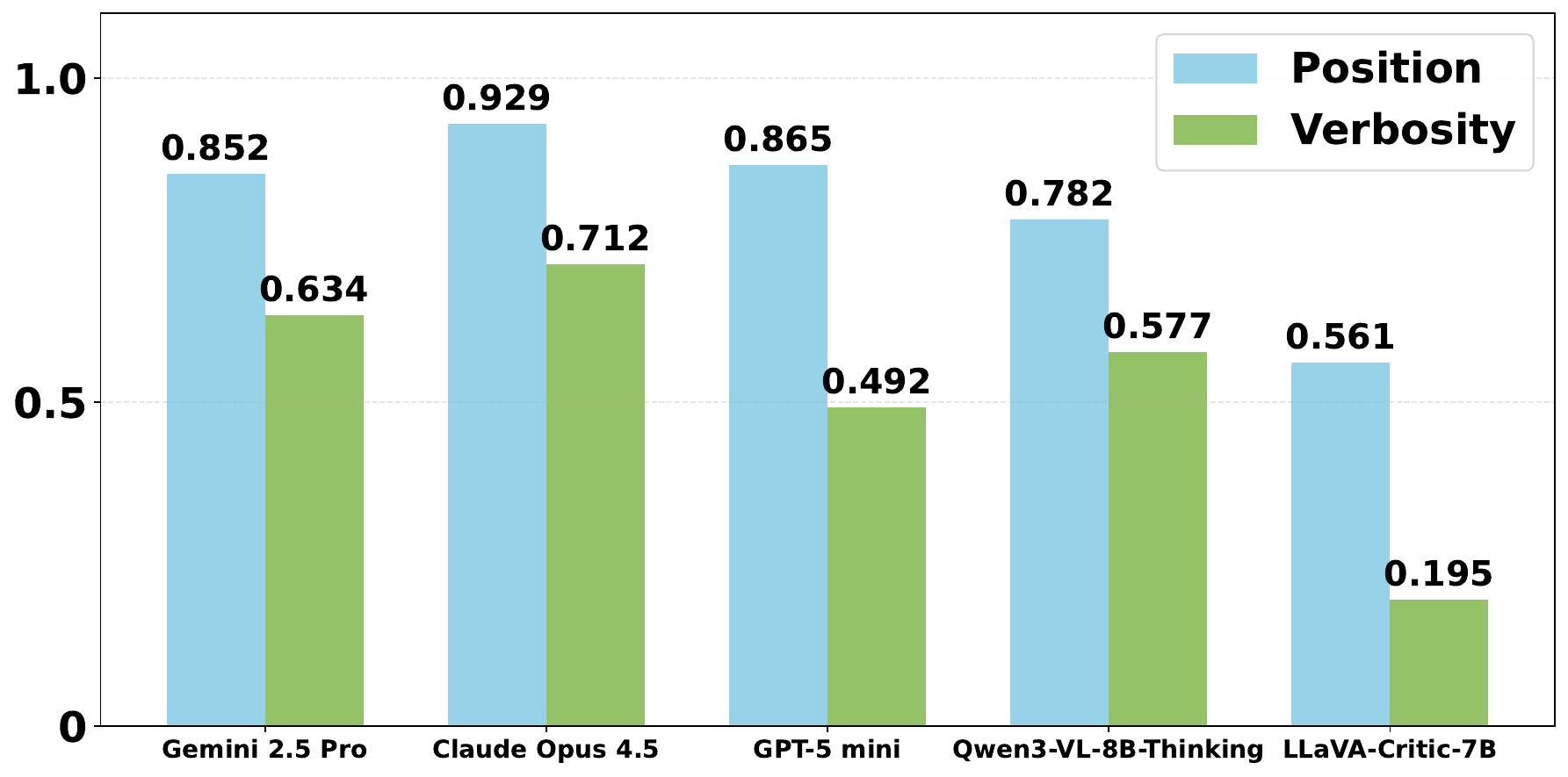}
    \caption{\textbf{Robustness on position and verbosity bias.} Robustness of five representative MLLMs to two representative biases in LLM-as-a-Judge, showing correlated trends and pronounced vulnerability to verbosity bias.}
    \label{fig:existing_bias}
\end{figure}

\subsection{Analysis on Existing Biases}

We extend our study to three existing biases that are especially important in LLM-as-a-Judge scenarios: \textit{position bias, verbosity bias, and self-enhancement bias}. 
Through this analysis, we examine whether judge biases previously observed in LLMs remain relevant in recent MLLMs and establish a complementary baseline for the general evaluative reliability of each judge. 
Detailed protocols are provided in Appendix~\ref{appendix:existing_biases}.

\paragraph{Position Bias} 
We examine whether judges preserve the same preference when the order of two candidate responses is reversed. 
As shown in Figure~\ref{fig:existing_bias}, MLLMs exhibit a wide range of robustness scores against position bias, indicating that order sensitivity observed in LLM-as-a-Judge still persists in MLLM-as-a-Judge settings.

\paragraph{Verbosity Bias} 
We analyze whether judges prefer a semantically equivalent but more verbose response over the original one. 
Figure~\ref{fig:existing_bias} also shows that MLLMs are even more vulnerable to verbosity bias than to position bias. 
In particular, some judges strongly favor verbose responses, indicating that superficial response length remains a major source of evaluative distortion.

\begin{figure}[t]
    \centering
    \includegraphics[width=\columnwidth]{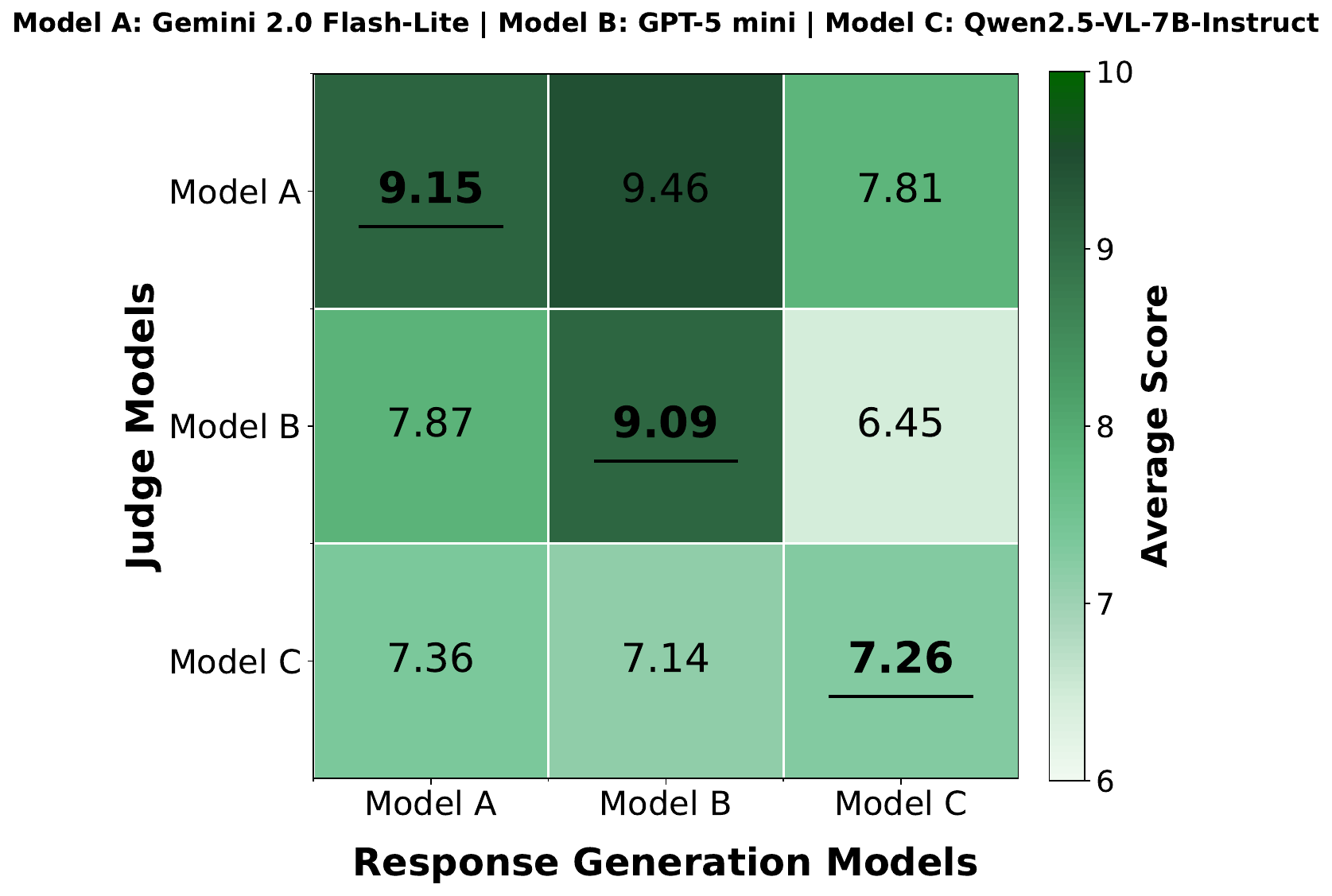}
    \caption{\textbf{Self-enhancement bias analysis.} Average judge-assigned scores for responses from three models, showing a tendency to favor self-generated outputs.}
    \label{fig:self_enhancement_bias}
\end{figure}

\paragraph{Self-Enhancement Bias} 
Finally, we analyze whether judges favor responses generated by themselves over those produced by other models. 
Figure~\ref{fig:self_enhancement_bias} shows that self-enhancement bias is also present, with judges tending to assign more favorable evaluations to their own outputs. 
This suggests that evaluator impartiality remains fragile when the judge is also part of the models to be evaluated.

These results show that, beyond the compositional biases studied in this work, conventional judge biases from LLM-as-a-Judge remain clearly relevant in MLLM-as-a-Judge as well.

\subsection{Analyses on Practical Validity}
We conduct additional analyses to examine the practical reliability of our protocol.
These analyses are conducted on three representative judges from the closed-source, open-source, and critic categories: GPT-5 mini (with reasoning high effort), Qwen3-VL-8B-Thinking, and LLaVA-Critic-7B.

\paragraph{Abstention-Aware Evaluation.}
One of the practical concerns is that some perturbed inputs may become effectively unjudgeable once critical evidence is removed or contradicted.
If such cases are still forced to receive numeric scores in our protocol, the resulting score change may reflect not only compositional bias but the lack of an abstention mechanism, raising the concern that our setting could overestimate bias on inherently invalid cases.
To examine this possibility, we repeat the evaluation with an explicit \texttt{N/A} option that allows judges to abstain when the evidence is insufficient or fundamentally inconsistent.
As shown in Table~\ref{tab:practical_abstention}, the \textit{Ratio} column denotes the proportion of samples excluded from evaluation due to abstention under the \texttt{N/A}-enabled setting. 
This ratio remains limited across all three judges, and even after excluding such samples, the overall scores change only marginally while similar trends are preserved across the three bias categories.
This supports the validity of our protocol as a reliable setting for analyzing compositional bias beyond inherently invalid cases.

\begin{table}[t]
\centering
\small
\renewcommand{\arraystretch}{1.08}
\setlength{\tabcolsep}{3.5pt}
\begin{tabular*}{\columnwidth}{@{\extracolsep{\fill}}l|c|ccc|cc@{}}
\toprule
Model & \texttt{N/A} & Int. & Cong. & Rob. & Overall & Ratio \\
\midrule
\multirow{2}{*}{GPT-5 mini}
&  & 0.223 & 0.589 & 0.916 & 0.612 & -- \\
& $\checkmark$ & 0.153 & 0.603 & 0.923 & 0.595 & 0.172 \\
\midrule
\multirow{2}{*}{Qwen3-VL}
&  & 0.377 & 0.604 & 0.900 & 0.660 & -- \\
& $\checkmark$ & 0.176 & 0.675 & 0.897 & 0.607 & 0.261 \\
\midrule
\multirow{2}{*}{LLaVA-Critic}
&  & 0.308 & 0.705 & 0.872 & 0.647 & -- \\
& $\checkmark$ & 0.176 & 0.628 & 0.915 & 0.605 & 0.065 \\
\bottomrule
\end{tabular*}
\caption{\textbf{Abstention-aware evaluation.} Comparison of the original and \texttt{N/A}-enabled settings for Integrality, Congruity, and Robustness, with the overall score and abstention ratio. (See Appendix~\ref{appendix:abstention_analysis} for detailed results and experimental setting.)}
\label{tab:practical_abstention}
\vspace{-2mm}
\end{table}

\paragraph{Score-guided Prompting.}
Another practical concern is that differences in how judges interpret the 1--10 score range may distort the measured bias patterns.
To examine this possibility, we augment the evaluation prompt with qualitative guidance for the score scale and compare the results with the original setting.
As shown in Table~\ref{tab:practical_prompting}, the overall scores change only modestly across all three judges.
This indicates that our main findings are not driven solely by model-specific score calibration, and supports the practical reliability of our protocol under realistic evaluation settings where judges may not share identical interpretations of the score range.

\paragraph{Modality-enforcing Prompting.}
Motivated by the results in Table~\ref{tab:main-results}, we investigate whether prompting strategies that explicitly enforce reasoning over all input modalities can mitigate the compositional biases identified in our benchmark.
To test this, we apply two prompting strategies: \textit{Modality Constraints}, which explicitly instructs the judge to verify all of $Q$, $I$, and $R$, and \textit{Modality Reasoning}, which further enforces a structured process of rubric decomposition, modality-wise inspection, and final synthesis.
As shown in Table~\ref{tab:practical_prompting}, both strategies improve the performance of strong general-purpose judges, with larger gains generally observed for the structured prompting strategy.
This suggests that explicitly encouraging judges to decompose the task and inspect each modality can help mitigate compositional bias.
At the same time, the trends differ across models, indicating that the effectiveness of such interventions is model-dependent and requires further analysis.

\begin{table}[t]
\centering
\small
\renewcommand{\arraystretch}{1.08}
\setlength{\tabcolsep}{4pt}
\begin{tabular}{l|c|cccc}
\toprule
Model & Orig. & +Guide & +Const. & +Reason. \\
\midrule
GPT-5 mini   & 0.612 & 0.609 & 0.644 & \textbf{0.645} \\
Qwen3-VL     & 0.660 & 0.674 & 0.694 & \textbf{0.726} \\
LLaVA-Critic & 0.647 & 0.631 & 0.600 & \textbf{0.653} \\
\bottomrule
\end{tabular}
\caption{\textbf{Prompt-level intervention.} Comparison of prompt variants: score-guided prompting (\textit{+Guide}), modality constraints (\textit{+Const.}), and modality reasoning (\textit{+Reason.}). (See Appendix~\ref{appendix:scale_sensitivity} and~\ref{appendix:prompt_intervention} for detailed results and prompting settings.)}
\label{tab:practical_prompting}
\vspace{-2mm}
\end{table}

\section{Conclusion}

In this work, we identify and formalize \textit{Compositional Bias} in MLLM-as-a-Judge, a failure mode in which the judge does not reliably ground its evaluation in the full multimodal context, but instead relies on partial, misaligned, or semantically irrelevant evidence. 
To systematically study this problem, we introduce \textit{MM-JudgeBias}, a benchmark designed to evaluate the reliability of multimodal judges under controlled perturbations. 
We organize compositional bias into three functional dimensions---\textit{Integrality}, \textit{Congruity}, and \textit{Robustness}---and instantiate them as nine fine-grained bias types. 
We further propose two complementary metrics, \textit{Bias-Deviation} (BD) and \textit{Bias-Conformity} (BC), to quantify sensitivity to semantically disruptive perturbations and stability under semantically neutral variations, respectively.

Through extensive experiments on 26 state-of-the-art MLLMs, including closed-source, open-source, and critic-specialized models, we show that compositional bias is pervasive in current MLLM judges. 
Our results reveal that many models struggle to holistically integrate query, image, and response, and often fail either to penalize missing or mismatched evidence or to remain stable under irrelevant perturbations. 
We also find that stronger reasoning ability or larger model scale does not necessarily translate into more reliable multimodal judgment, highlighting that judgment reliability is a distinct capability from general task performance.

\paragraph{Acknowledgements.}
This work was initiated during the first author's internship at NAVER Cloud. We thank MyungIn You for coordinating the human review process.

\clearpage

\section*{Limitations}
While this study establishes a robust framework for evaluating MLLM judges, certain limitations remain that provide avenues for future research. 
First, although the landscape of bias in large-scale models is multifaceted—encompassing cultural, social, and geographic dimensions—our investigation intentionally focuses on multimodality-specific biases. 
By narrowing our scope to distortions arising from visual-textual interplay, such as integrality and congruity, we prioritize the assessment of a judge's fundamental ability to synchronize cross-modal information. 
While our benchmark does not exhaustively cover the broader spectrum of human-centric biases, the proposed BC and BD metrics are designed to be extensible, serving as a foundational methodology that can be naturally adapted to evaluate more nuanced, domain-specific biases in future studies.

Another limitation is that the current evaluation is primarily tailored to vision-language interactions within a pointwise scoring paradigm. 
Although this setup is sufficient to demonstrate the evaluative reliability of current MLLMs, the principles of bias-conformity and deviation are inherently modality-agnostic and generalizable. 
This allows for a significant expansion into ``Any-to-Any'' modality scenarios—including video, audio, or interleaved multimodal sequences—to verify if the observed reliability patterns hold true across a broader spectrum of sensory inputs. 
Future work could also generalize these metrics to alternative judgment protocols, such as pairwise comparisons or batch-ranking scenarios, to provide a more comprehensive understanding of the decision-making boundaries of MLLM judges in diverse evaluative contexts.

\bibliography{custom}

@article{zhang2023gpt,
  title={Gpt-4v (ision) as a generalist evaluator for vision-language tasks},
  author={Zhang, Xinlu and Lu, Yujie and Wang, Weizhi and Yan, An and Yan, Jun and Qin, Lianke and Wang, Heng and Yan, Xifeng and Wang, William Yang and Petzold, Linda Ruth},
  journal={arXiv preprint arXiv:2311.01361},
  year={2023}
}

@inproceedings{lee-etal-2024-prometheus,
    title = "Prometheus-Vision: Vision-Language Model as a Judge for Fine-Grained Evaluation",
    author = "Lee, Seongyun  and
      Kim, Seungone  and
      Park, Sue  and
      Kim, Geewook  and
      Seo, Minjoon",
    editor = "Ku, Lun-Wei  and
      Martins, Andre  and
      Srikumar, Vivek",
    booktitle = "Findings of the Association for Computational Linguistics: ACL 2024",
    month = aug,
    year = "2024",
    address = "Bangkok, Thailand",
    publisher = "Association for Computational Linguistics",
    url = "https://aclanthology.org/2024.findings-acl.672/",
    doi = "10.18653/v1/2024.findings-acl.672",
    pages = "11286--11315"
}

@inproceedings{Xiong_2025_CVPR,
    author    = {Xiong, Tianyi and Wang, Xiyao and Guo, Dong and Ye, Qinghao and Fan, Haoqi and Gu, Quanquan and Huang, Heng and Li, Chunyuan},
    title     = {LLaVA-Critic: Learning to Evaluate Multimodal Models},
    booktitle = {Proceedings of the IEEE/CVF Conference on Computer Vision and Pattern Recognition (CVPR)},
    month     = {June},
    year      = {2025},
    pages     = {13618-13628}
}

@article{gu2024survey,
  title={A Survey on LLM-as-a-Judge},
  author={Gu, Jiawei and Jiang, Xuhui and Shi, Zhichao and Tan, Hexiang and Zhai, Xuehao and Xu, Chengjin and Li, Wei and Shen, Yinghan and Ma, Shengjie and Liu, Honghao and others},
  journal={arXiv preprint arXiv:2411.15594},
  year={2024}
}

@inproceedings{
ye2024justice,
title={Justice or Prejudice? Quantifying Biases in {LLM}-as-a-Judge},
author={Jiayi Ye and Yanbo Wang and Yue Huang and Dongping Chen and Qihui Zhang and Nuno Moniz and Tian Gao and Werner Geyer and Chao Huang and Pin-Yu Chen and Nitesh V Chawla and Xiangliang Zhang},
booktitle={The Thirteenth International Conference on Learning Representations},
year={2025},
url={https://openreview.net/forum?id=3GTtZFiajM}
}

@inproceedings{shi2024judging,
    title = "Judging the Judges: A Systematic Study of Position Bias in {LLM}-as-a-Judge",
    author = "Shi, Lin  and
      Ma, Chiyu  and
      Liang, Wenhua  and
      Diao, Xingjian  and
      Ma, Weicheng  and
      Vosoughi, Soroush",
    editor = "Inui, Kentaro  and
      Sakti, Sakriani  and
      Wang, Haofen  and
      Wong, Derek F.  and
      Bhattacharyya, Pushpak  and
      Banerjee, Biplab  and
      Ekbal, Asif  and
      Chakraborty, Tanmoy  and
      Singh, Dhirendra Pratap",
    booktitle = "Proceedings of the 14th International Joint Conference on Natural Language Processing and the 4th Conference of the Asia-Pacific Chapter of the Association for Computational Linguistics",
    month = dec,
    year = "2025",
    address = "Mumbai, India",
    publisher = "The Asian Federation of Natural Language Processing and The Association for Computational Linguistics",
    url = "https://aclanthology.org/2025.ijcnlp-long.18/",
    doi = "10.18653/v1/2025.ijcnlp-long.18",
    pages = "292--314",
    ISBN = "979-8-89176-298-5"
}

@inproceedings{li2025evaluating,
  title     = {Evaluating Scoring Bias in {LLM}-as-a-Judge},
  author    = {Li, Qingquan and Dou, Shaoyu and Shao, Kailai and Chen, Chao and Hu, Haixiang},
  booktitle = {Database Systems for Advanced Applications - 31st International Conference, {DASFAA} 2026},
  series    = {Lecture Notes in Computer Science},
  publisher = {Springer},
  year      = {2026}
}

@inproceedings{
li2025preference,
title={Preference Leakage: A Contamination Problem in {LLM}-as-a-judge},
author={Dawei Li and Renliang Sun and Yue Huang and Ming Zhong and Bohan Jiang and Jiawei Han and Xiangliang Zhang and Wei Wang and huan liu},
booktitle={The Fourteenth International Conference on Learning Representations},
year={2026},
url={https://openreview.net/forum?id=grIvSXVJ65}
}

@article{spiliopoulou2025play,
  title={Play favorites: A statistical method to measure self-bias in llm-as-a-judge},
  author={Spiliopoulou, Evangelia and Fogliato, Riccardo and Burnsky, Hanna and Soliman, Tamer and Ma, Jie and Horwood, Graham and Ballesteros, Miguel},
  journal={arXiv preprint arXiv:2508.06709},
  year={2025}
}

@inproceedings{chen2024humans,
    title = "Humans or {LLM}s as the Judge? A Study on Judgement Bias",
    author = "Chen, Guiming Hardy  and
      Chen, Shunian  and
      Liu, Ziche  and
      Jiang, Feng  and
      Wang, Benyou",
    editor = "Al-Onaizan, Yaser  and
      Bansal, Mohit  and
      Chen, Yun-Nung",
    booktitle = "Proceedings of the 2024 Conference on Empirical Methods in Natural Language Processing",
    month = nov,
    year = "2024",
    address = "Miami, Florida, USA",
    publisher = "Association for Computational Linguistics",
    url = "https://aclanthology.org/2024.emnlp-main.474/",
    doi = "10.18653/v1/2024.emnlp-main.474",
    pages = "8301--8327"
}

@article{gao2025evaluating,
  title={Evaluating and Mitigating LLM-as-a-judge Bias in Communication Systems},
  author={Gao, Jiaxin and Chen, Chen and Jia, Yanwen and Gong, Xueluan and Lam, Kwok-Yan and Wang, Qian},
  journal={arXiv preprint arXiv:2510.12462},
  year={2025}
}

@inproceedings{chen2024mllm,
  title={Mllm-as-a-judge: Assessing multimodal llm-as-a-judge with vision-language benchmark},
  author={Chen, Dongping and Chen, Ruoxi and Zhang, Shilin and Wang, Yaochen and Liu, Yinuo and Zhou, Huichi and Zhang, Qihui and Wan, Yao and Zhou, Pan and Sun, Lichao},
  booktitle={Forty-first International Conference on Machine Learning},
  year={2024}
}

@inproceedings{hwang2025fooling,
  title={Fooling the LVLM Judges: Visual Biases in LVLM-Based Evaluation},
  author={Hwang, Yerin and Lee, Dongryeol and Min, Kyungmin and Kang, Taegwan and Kim, Yongil and Jung, Kyomin},
  booktitle={Proceedings of the 2025 Conference on Empirical Methods in Natural Language Processing},
  pages={23197--23216},
  year={2025}
}

@inproceedings{wang2025judge,
author = {Pu, Shu and Wang, Yaochen and Chen, Dongping and Chen, Yuhang and Wang, Guohao and Qin, Qi and Zhang, Zhongyi and Zhang, Zhiyuan and Zhou, Zetong and Gong, Shuang and Gui, Yi and Wan, Yao and Yu, Philip S.},
title = {Judge Anything: MLLM as a Judge Across Any Modality},
year = {2025},
isbn = {9798400714542},
publisher = {Association for Computing Machinery},
address = {New York, NY, USA},
url = {https://doi.org/10.1145/3711896.3737409},
doi = {10.1145/3711896.3737409},
booktitle = {Proceedings of the 31st ACM SIGKDD Conference on Knowledge Discovery and Data Mining V.2},
pages = {5742–5753},
numpages = {12},
location = {Toronto ON, Canada},
series = {KDD '25}
}

@inproceedings{laskar-etal-2025-judging,
    title = "Judging the Judges: Can Large Vision-Language Models Fairly Evaluate Chart Comprehension and Reasoning?",
    author = "Laskar, Md Tahmid Rahman  and
      Islam, Mohammed Saidul  and
      Mahbub, Ridwan  and
      Masry, Ahmed  and
      Rahman, Mizanur  and
      Bhuiyan, Amran  and
      Nayeem, Mir Tafseer  and
      Joty, Shafiq  and
      Hoque, Enamul  and
      Huang, Jimmy",
    editor = "Rehm, Georg  and
      Li, Yunyao",
    booktitle = "Proceedings of the 63rd Annual Meeting of the Association for Computational Linguistics (Volume 6: Industry Track)",
    month = jul,
    year = "2025",
    address = "Vienna, Austria",
    publisher = "Association for Computational Linguistics",
    url = "https://aclanthology.org/2025.acl-industry.83/",
    doi = "10.18653/v1/2025.acl-industry.83",
    pages = "1203--1216",
    ISBN = "979-8-89176-288-6"
}

@inproceedings{xiong2025multi,
  title     = {Multi-Crit: Benchmarking Multimodal Judges on Pluralistic Criteria-Following},
  author    = {Xiong, Tianyi and Ge, Yi and Li, Ming and Zhang, Zuolong and Kulkarni, Pranav and Wang, Kaishen and He, Qi and Zhu, Zeying and Liu, Chenxi and Chen, Ruibo and Zheng, Tong and Chen, Yanshuo and Wang, Xiyao and Zhang, Renrui and Chen, Wenhu and Huang, Heng},
  booktitle = {Proceedings of the {IEEE/CVF} Conference on Computer Vision and Pattern Recognition ({CVPR})},
  year      = {2026},
}

@article{yasunaga2025multimodal,
  title={Multimodal rewardbench: Holistic evaluation of reward models for vision language models},
  author={Yasunaga, Michihiro and Zettlemoyer, Luke and Ghazvininejad, Marjan},
  journal={arXiv preprint arXiv:2502.14191},
  year={2025}
}

@inproceedings{Ruan_2025_ICCV,
    author    = {Ruan, Jiacheng and Yuan, Wenzhen and Gao, Xian and Guo, Ye and Zhang, Daoxin and Xu, Zhe and Hu, Yao and Liu, Ting and Fu, Yuzhuo},
    title     = {VLRMBench: A Comprehensive and Challenging Benchmark for Vision-Language Reward Models},
    booktitle = {Proceedings of the IEEE/CVF International Conference on Computer Vision (ICCV)},
    month     = {October},
    year      = {2025},
    pages     = {3163-3173}
}

@inproceedings{Li_2025_CVPR,
    author    = {Li, Lei and Wei, Yuancheng and Xie, Zhihui and Yang, Xuqing and Song, Yifan and Wang, Peiyi and An, Chenxin and Liu, Tianyu and Li, Sujian and Lin, Bill Yuchen and Kong, Lingpeng and Liu, Qi},
    title     = {VL-RewardBench: A Challenging Benchmark for Vision-Language Generative Reward Models},
    booktitle = {Proceedings of the IEEE/CVF Conference on Computer Vision and Pattern Recognition (CVPR)},
    month     = {June},
    year      = {2025},
    pages     = {24657-24668}
}

@inproceedings{zeng-etal-2025-mm,
    title = "{MM}-{CRITIC}: A Holistic Evaluation of Large Multimodal Models as Multimodal Critique",
    author = "Zeng, Gailun  and
      Luo, Ziyang  and
      Lin, Hongzhan  and
      Tian, Yuchen  and
      Li, Kaixin  and
      Gong, Ziyang  and
      Guo, Jianxiong  and
      Ma, Jing",
    editor = "Christodoulopoulos, Christos  and
      Chakraborty, Tanmoy  and
      Rose, Carolyn  and
      Peng, Violet",
    booktitle = "Findings of the Association for Computational Linguistics: EMNLP 2025",
    month = nov,
    year = "2025",
    address = "Suzhou, China",
    publisher = "Association for Computational Linguistics",
    url = "https://aclanthology.org/2025.findings-emnlp.733/",
    doi = "10.18653/v1/2025.findings-emnlp.733",
    pages = "13603--13630",
    ISBN = "979-8-89176-335-7"
}

@inproceedings{chen2024quantifying,
    title = "Quantifying and Mitigating Unimodal Biases in Multimodal Large Language Models: A Causal Perspective",
    author = "Chen, Meiqi  and
      Cao, Yixin  and
      Zhang, Yan  and
      Lu, Chaochao",
    editor = "Al-Onaizan, Yaser  and
      Bansal, Mohit  and
      Chen, Yun-Nung",
    booktitle = "Findings of the Association for Computational Linguistics: EMNLP 2024",
    month = nov,
    year = "2024",
    address = "Miami, Florida, USA",
    publisher = "Association for Computational Linguistics",
    url = "https://aclanthology.org/2024.findings-emnlp.960/",
    doi = "10.18653/v1/2024.findings-emnlp.960",
    pages = "16449--16469"
}

@inproceedings{zhao-etal-2025-looking,
    title = "Looking Beyond Text: Reducing Language Bias in Large Vision-Language Models via Multimodal Dual-Attention and Soft-Image Guidance",
    author = "Zhao, Haozhe  and
      Si, Shuzheng  and
      Chen, Liang  and
      Zhang, Yichi  and
      Sun, Maosong  and
      Chang, Baobao  and
      Zhang, Minjia",
    editor = "Christodoulopoulos, Christos  and
      Chakraborty, Tanmoy  and
      Rose, Carolyn  and
      Peng, Violet",
    booktitle = "Proceedings of the 2025 Conference on Empirical Methods in Natural Language Processing",
    month = nov,
    year = "2025",
    address = "Suzhou, China",
    publisher = "Association for Computational Linguistics",
    url = "https://aclanthology.org/2025.emnlp-main.995/",
    doi = "10.18653/v1/2025.emnlp-main.995",
    pages = "19677--19701",
    ISBN = "979-8-89176-332-6"
}

@article{zheng2025unveiling,
  title={Unveiling Intrinsic Text Bias in Multimodal Large Language Models through Attention Key-Space Analysis},
  author={Zheng, Xinhan and Wu, Huyu and Wang, Xueting and Jiang, Haiyun},
  journal={arXiv preprint arXiv:2510.26721},
  year={2025}
}

@article{lin2025unveiling,
  title={Unveiling Modality Bias: Automated Sample-Specific Analysis for Multimodal Misinformation Benchmarks},
  author={Lin, Hehai and Liu, Hui and Cao, Shilei and Li, Jing and Li, Haoliang and Wang, Wenya},
  journal={arXiv preprint arXiv:2511.05883},
  year={2025}
}

@inproceedings{zheng2025mllms,
  title     = {{MLLMs} are Deeply Affected by Modality Bias},
  author    = {Zheng, Xu and Liao, Chenfei and Fu, Yuqian and Lei, Kaiyu and Lyu, Yuanhuiyi and Jiang, Lutao and Ren, Bin and Chen, Jialei and Wang, Jiawen and Li, Chengxin and Zhang, Linfeng and Paudel, Danda Pani and Huang, Xuanjing and Jiang, Yu-Gang and Sebe, Nicu and Tao, Dacheng and Van Gool, Luc and Hu, Xuming},
  booktitle = {{ICLR} 2026 Workshop on Multimodal Intelligence: Next Token Prediction {\&} Beyond},
  year      = {2026}
}

@article{li2024llava,
  title={Llava-onevision: Easy visual task transfer},
  author={Li, Bo and Zhang, Yuanhan and Guo, Dong and Zhang, Renrui and Li, Feng and Zhang, Hao and Zhang, Kaichen and Zhang, Peiyuan and Li, Yanwei and Liu, Ziwei and others},
  journal={arXiv preprint arXiv:2408.03326},
  year={2024}
}

@InProceedings{Liu_2024_CVPR,
    author    = {Liu, Haotian and Li, Chunyuan and Li, Yuheng and Lee, Yong Jae},
    title     = {Improved Baselines with Visual Instruction Tuning},
    booktitle = {Proceedings of the IEEE/CVF Conference on Computer Vision and Pattern Recognition (CVPR)},
    month     = {June},
    year      = {2024},
    pages     = {26296-26306}
}

@inproceedings{caffagni2024revolution,
    title = "The Revolution of Multimodal Large Language Models: A Survey",
    author = "Caffagni, Davide  and
      Cocchi, Federico  and
      Barsellotti, Luca  and
      Moratelli, Nicholas  and
      Sarto, Sara  and
      Baraldi, Lorenzo  and
      Baraldi, Lorenzo  and
      Cornia, Marcella  and
      Cucchiara, Rita",
    editor = "Ku, Lun-Wei  and
      Martins, Andre  and
      Srikumar, Vivek",
    booktitle = "Findings of the Association for Computational Linguistics: ACL 2024",
    month = aug,
    year = "2024",
    address = "Bangkok, Thailand",
    publisher = "Association for Computational Linguistics",
    url = "https://aclanthology.org/2024.findings-acl.807/",
    doi = "10.18653/v1/2024.findings-acl.807",
    pages = "13590--13618"
}

@article{comanici2025gemini,
  title={Gemini 2.5: Pushing the frontier with advanced reasoning, multimodality, long context, and next generation agentic capabilities},
  author={Comanici, Gheorghe and Bieber, Eric and Schaekermann, Mike and Pasupat, Ice and Sachdeva, Noveen and Dhillon, Inderjit and Blistein, Marcel and Ram, Ori and Zhang, Dan and Rosen, Evan and others},
  journal={arXiv preprint arXiv:2507.06261},
  year={2025}
}

@article{achiam2023gpt,
  title={Gpt-4 technical report},
  author={Achiam, Josh and Adler, Steven and Agarwal, Sandhini and Ahmad, Lama and Akkaya, Ilge and Aleman, Florencia Leoni and Almeida, Diogo and Altenschmidt, Janko and Altman, Sam and Anadkat, Shyamal and others},
  journal={arXiv preprint arXiv:2303.08774},
  year={2023}
}

@article{li2024llms,
  title={Llms-as-judges: a comprehensive survey on llm-based evaluation methods},
  author={Li, Haitao and Dong, Qian and Chen, Junjie and Su, Huixue and Zhou, Yujia and Ai, Qingyao and Ye, Ziyi and Liu, Yiqun},
  journal={arXiv preprint arXiv:2412.05579},
  year={2024}
}

@article{zheng2023judging,
  title={Judging llm-as-a-judge with mt-bench and chatbot arena},
  author={Zheng, Lianmin and Chiang, Wei-Lin and Sheng, Ying and Zhuang, Siyuan and Wu, Zhanghao and Zhuang, Yonghao and Lin, Zi and Li, Zhuohan and Li, Dacheng and Xing, Eric and others},
  journal={Advances in neural information processing systems},
  volume={36},
  pages={46595--46623},
  year={2023}
}

@article{liu2023visual,
  title={Visual instruction tuning},
  author={Liu, Haotian and Li, Chunyuan and Wu, Qingyang and Lee, Yong Jae},
  journal={Advances in neural information processing systems},
  volume={36},
  pages={34892--34916},
  year={2023}
}

@inproceedings{yu2023mm,
author = {Yu, Weihao and Yang, Zhengyuan and Li, Linjie and Wang, Jianfeng and Lin, Kevin and Liu, Zicheng and Wang, Xinchao and Wang, Lijuan},
title = {MM-Vet: evaluating large multimodal models for integrated capabilities},
year = {2024},
publisher = {JMLR.org},
booktitle = {Proceedings of the 41st International Conference on Machine Learning},
articleno = {2381},
numpages = {25},
location = {Vienna, Austria},
series = {ICML'24}
}

@inproceedings{lin2014microsoft,
  title={Microsoft coco: Common objects in context},
  author={Lin, Tsung-Yi and Maire, Michael and Belongie, Serge and Hays, James and Perona, Pietro and Ramanan, Deva and Doll{\'a}r, Piotr and Zitnick, C Lawrence},
  booktitle={European conference on computer vision},
  pages={740--755},
  year={2014},
  organization={Springer}
}

@inproceedings{xiao2010sun,
  title={Sun database: Large-scale scene recognition from abbey to zoo},
  author={Xiao, Jianxiong and Hays, James and Ehinger, Krista A and Oliva, Aude and Torralba, Antonio},
  booktitle={2010 IEEE computer society conference on computer vision and pattern recognition},
  pages={3485--3492},
  year={2010},
  organization={IEEE}
}

@article{zhou2017places,
  title={Places: A 10 million image database for scene recognition},
  author={Zhou, Bolei and Lapedriza, Agata and Khosla, Aditya and Oliva, Aude and Torralba, Antonio},
  journal={IEEE transactions on pattern analysis and machine intelligence},
  volume={40},
  number={6},
  pages={1452--1464},
  year={2017},
  publisher={IEEE}
}

@article{bai2025qwen2,
  title={Qwen2.5-VL Technical Report},
  author={Bai, Shuai and Chen, Keqin and Liu, Xuejing and Wang, Jialin and Ge, Wenbin and Song, Sibo and Dang, Kai and Wang, Peng and Wang, Shijie and Tang, Jun and others},
  journal={arXiv preprint arXiv:2502.13923},
  year={2025}
}

@misc{gpt5,
      title={GPT-5 System Card}, 
      author={OpenAI},
      year={2025},
      url={https://cdn.openai.com/gpt-5-system-card.pdf}, 
}

@misc{o3,
      title={OpenAI o3 and o4-mini System Card}, 
      author={OpenAI},
      year={2025},
      url={https://cdn.openai.com/pdf/2221c875-02dc-4789-800b-e7758f3722c1/o3-and-o4-mini-system-card.pdf}, 
}

@misc{opus4.5,
      title={System Card: Claude Opus 4.5}, 
      author={ANTHROPIC},
      year={2025},
      url={https://assets.anthropic.com/m/64823ba7485345a7/Claude-Opus-4-5-System-Card.pdf}, 
}

@misc{sonnet4.5,
      title={System Card: Claude Sonnet 4.5}, 
      author={ANTHROPIC},
      year={2025},
      url={https://assets.anthropic.com/m/12f214efcc2f457a/original/Claude-Sonnet-4-5-System-Card.pdf}, 
}

@misc{haiku4.5,
      title={System Card: Claude Haiku 4.5}, 
      author={ANTHROPIC},
      year={2025},
      url={https://assets.anthropic.com/m/99128ddd009bdcb/Claude-Haiku-4-5-System-Card.pdf}, 
}

@article{Qwen3-VL,
      title={Qwen3-VL Technical Report}, 
      author={Shuai Bai and Yuxuan Cai and Ruizhe Chen and Keqin Chen and Xionghui Chen and Zesen Cheng and Lianghao Deng and Wei Ding and Chang Gao and Chunjiang Ge and Wenbin Ge and Zhifang Guo and Qidong Huang and Jie Huang and Fei Huang and Binyuan Hui and Shutong Jiang and Zhaohai Li and Mingsheng Li and Mei Li and Kaixin Li and Zicheng Lin and Junyang Lin and Xuejing Liu and Jiawei Liu and Chenglong Liu and Yang Liu and Dayiheng Liu and Shixuan Liu and Dunjie Lu and Ruilin Luo and Chenxu Lv and Rui Men and Lingchen Meng and Xuancheng Ren and Xingzhang Ren and Sibo Song and Yuchong Sun and Jun Tang and Jianhong Tu and Jianqiang Wan and Peng Wang and Pengfei Wang and Qiuyue Wang and Yuxuan Wang and Tianbao Xie and Yiheng Xu and Haiyang Xu and Jin Xu and Zhibo Yang and Mingkun Yang and Jianxin Yang and An Yang and Bowen Yu and Fei Zhang and Hang Zhang and Xi Zhang and Bo Zheng and Humen Zhong and Jingren Zhou and Fan Zhou and Jing Zhou and Yuanzhi Zhu and Ke Zhu},
	  journal={arXiv preprint arXiv:2511.21631},
      year={2025}
}

@article{wang2025internvl3,
  title={Internvl3.5: Advancing open-source multimodal models in versatility, reasoning, and efficiency},
  author={Wang, Weiyun and Gao, Zhangwei and Gu, Lixin and Pu, Hengjun and Cui, Long and Wei, Xingguang and Liu, Zhaoyang and Jing, Linglin and Ye, Shenglong and Shao, Jie and others},
  journal={arXiv preprint arXiv:2508.18265},
  year={2025}
}

@inproceedings{cao2022augmented,
  title={An augmented benchmark dataset for geometric question answering through dual parallel text encoding},
  author={Cao, Jie and Xiao, Jing},
  booktitle={Proceedings of the 29th international conference on computational linguistics},
  pages={1511--1520},
  year={2022}
}

@inproceedings{lu2021inter,
    title = "{I}nter-{GPS}: Interpretable Geometry Problem Solving with Formal Language and Symbolic Reasoning",
    author = "Lu, Pan  and
      Gong, Ran  and
      Jiang, Shibiao  and
      Qiu, Liang  and
      Huang, Siyuan  and
      Liang, Xiaodan  and
      Zhu, Song-Chun",
    editor = "Zong, Chengqing  and
      Xia, Fei  and
      Li, Wenjie  and
      Navigli, Roberto",
    booktitle = "Proceedings of the 59th Annual Meeting of the Association for Computational Linguistics and the 11th International Joint Conference on Natural Language Processing (Volume 1: Long Papers)",
    month = aug,
    year = "2021",
    address = "Online",
    publisher = "Association for Computational Linguistics",
    url = "https://aclanthology.org/2021.acl-long.528/",
    doi = "10.18653/v1/2021.acl-long.528",
    pages = "6774--6786"
}

@inproceedings{chen2022unigeo,
    title = "{U}ni{G}eo: Unifying Geometry Logical Reasoning via Reformulating Mathematical Expression",
    author = "Chen, Jiaqi  and
      Li, Tong  and
      Qin, Jinghui  and
      Lu, Pan  and
      Lin, Liang  and
      Chen, Chongyu  and
      Liang, Xiaodan",
    editor = "Goldberg, Yoav  and
      Kozareva, Zornitsa  and
      Zhang, Yue",
    booktitle = "Proceedings of the 2022 Conference on Empirical Methods in Natural Language Processing",
    month = dec,
    year = "2022",
    address = "Abu Dhabi, United Arab Emirates",
    publisher = "Association for Computational Linguistics",
    url = "https://aclanthology.org/2022.emnlp-main.218/",
    doi = "10.18653/v1/2022.emnlp-main.218",
    pages = "3313--3323"
}

@inproceedings{singh2019towards,
  title={Towards vqa models that can read},
  author={Singh, Amanpreet and Natarajan, Vivek and Shah, Meet and Jiang, Yu and Chen, Xinlei and Batra, Dhruv and Parikh, Devi and Rohrbach, Marcus},
  booktitle={Proceedings of the IEEE/CVF conference on computer vision and pattern recognition},
  pages={8317--8326},
  year={2019}
}

@inproceedings{mathew2021docvqa,
  title={Docvqa: A dataset for vqa on document images},
  author={Mathew, Minesh and Karatzas, Dimosthenis and Jawahar, CV},
  booktitle={Proceedings of the IEEE/CVF winter conference on applications of computer vision},
  pages={2200--2209},
  year={2021}
}

@inproceedings{yang2025scaling,
    title = "Scaling Text-Rich Image Understanding via Code-Guided Synthetic Multimodal Data Generation",
    author = "Yang, Yue  and
      Patel, Ajay  and
      Deitke, Matt  and
      Gupta, Tanmay  and
      Weihs, Luca  and
      Head, Andrew  and
      Yatskar, Mark  and
      Callison-Burch, Chris  and
      Krishna, Ranjay  and
      Kembhavi, Aniruddha  and
      Clark, Christopher",
    editor = "Che, Wanxiang  and
      Nabende, Joyce  and
      Shutova, Ekaterina  and
      Pilehvar, Mohammad Taher",
    booktitle = "Proceedings of the 63rd Annual Meeting of the Association for Computational Linguistics (Volume 1: Long Papers)",
    month = jul,
    year = "2025",
    address = "Vienna, Austria",
    publisher = "Association for Computational Linguistics",
    url = "https://aclanthology.org/2025.acl-long.855/",
    doi = "10.18653/v1/2025.acl-long.855",
    pages = "17486--17505",
    ISBN = "979-8-89176-251-0"
}

@inproceedings{masry2025chartqapro,
    title = "{C}hart{QAP}ro: A More Diverse and Challenging Benchmark for Chart Question Answering",
    author = "Masry, Ahmed  and
      Islam, Mohammed Saidul  and
      Ahmed, Mahir  and
      Bajaj, Aayush  and
      Kabir, Firoz  and
      Kartha, Aaryaman  and
      Laskar, Md Tahmid Rahman  and
      Rahman, Mizanur  and
      Rahman, Shadikur  and
      Shahmohammadi, Mehrad  and
      Thakkar, Megh  and
      Parvez, Md Rizwan  and
      Hoque, Enamul  and
      Joty, Shafiq",
    editor = "Che, Wanxiang  and
      Nabende, Joyce  and
      Shutova, Ekaterina  and
      Pilehvar, Mohammad Taher",
    booktitle = "Findings of the Association for Computational Linguistics: ACL 2025",
    month = jul,
    year = "2025",
    address = "Vienna, Austria",
    publisher = "Association for Computational Linguistics",
    url = "https://aclanthology.org/2025.findings-acl.978/",
    doi = "10.18653/v1/2025.findings-acl.978",
    pages = "19123--19151",
    ISBN = "979-8-89176-256-5"
}

@article{xu2023chartbench,
    title={ChartBench: A Benchmark for Complex Visual Reasoning in Charts},
    author={Zhengzhuo Xu and Sinan Du and Yiyan Qi and Chengjin Xu and Chun Yuan and Jian Guo},
    journal={ArXiv},
    year={2023},
    volume={abs/2312.15915},
    url={https://api.semanticscholar.org/CorpusID:266550948}
}

@inproceedings{kembhavi2016diagram,
  title={A diagram is worth a dozen images},
  author={Kembhavi, Aniruddha and Salvato, Mike and Kolve, Eric and Seo, Minjoon and Hajishirzi, Hannaneh and Farhadi, Ali},
  booktitle={European conference on computer vision},
  pages={235--251},
  year={2016},
  organization={Springer}
}

@inproceedings{kembhavi2017you,
  title={Are you smarter than a sixth grader? textbook question answering for multimodal machine comprehension},
  author={Kembhavi, Aniruddha and Seo, Minjoon and Schwenk, Dustin and Choi, Jonghyun and Farhadi, Ali and Hajishirzi, Hannaneh},
  booktitle={Proceedings of the IEEE Conference on Computer Vision and Pattern recognition},
  pages={4999--5007},
  year={2017}
}

@inproceedings{mathew2022infographicvqa,
  title={Infographicvqa},
  author={Mathew, Minesh and Bagal, Viraj and Tito, Rub{\`e}n and Karatzas, Dimosthenis and Valveny, Ernest and Jawahar, CV},
  booktitle={Proceedings of the IEEE/CVF Winter Conference on Applications of Computer Vision},
  pages={1697--1706},
  year={2022}
}

@inproceedings{zheng2024multimodal,
    title = "Multimodal Table Understanding",
    author = "Zheng, Mingyu  and
      Feng, Xinwei  and
      Si, Qingyi  and
      She, Qiaoqiao  and
      Lin, Zheng  and
      Jiang, Wenbin  and
      Wang, Weiping",
    editor = "Ku, Lun-Wei  and
      Martins, Andre  and
      Srikumar, Vivek",
    booktitle = "Proceedings of the 62nd Annual Meeting of the Association for Computational Linguistics (Volume 1: Long Papers)",
    month = aug,
    year = "2024",
    address = "Bangkok, Thailand",
    publisher = "Association for Computational Linguistics",
    url = "https://aclanthology.org/2024.acl-long.493/",
    doi = "10.18653/v1/2024.acl-long.493",
    pages = "9102--9124"
}

@article{yang2023large,
  title={A large-scale dataset for end-to-end table recognition in the wild},
  author={Yang, Fan and Hu, Lei and Liu, Xinwu and Huang, Shuangping and Gu, Zhenghui},
  journal={Scientific Data},
  volume={10},
  number={1},
  pages={110},
  year={2023},
  publisher={Nature Publishing Group UK London}
}

@inproceedings{cheng2024seeclick,
  title={Seeclick: Harnessing gui grounding for advanced visual gui agents},
  author={Cheng, Kanzhi and Sun, Qiushi and Chu, Yougang and Xu, Fangzhi and YanTao, Li and Zhang, Jianbing and Wu, Zhiyong},
  booktitle={Proceedings of the 62nd Annual Meeting of the Association for Computational Linguistics (Volume 1: Long Papers)},
  pages={9313--9332},
  year={2024}
}

@inproceedings{deka2017rico,
  title={Rico: A mobile app dataset for building data-driven design applications},
  author={Deka, Biplab and Huang, Zifeng and Franzen, Chad and Hibschman, Joshua and Afergan, Daniel and Li, Yang and Nichols, Jeffrey and Kumar, Ranjitha},
  booktitle={Proceedings of the 30th annual ACM symposium on user interface software and technology},
  pages={845--854},
  year={2017}
}

@inproceedings{schwenk2022okvqa,
  title={A-okvqa: A benchmark for visual question answering using world knowledge},
  author={Schwenk, Dustin and Khandelwal, Apoorv and Clark, Christopher and Marino, Kenneth and Mottaghi, Roozbeh},
  booktitle={European conference on computer vision},
  pages={146--162},
  year={2022},
  organization={Springer}
}

@inproceedings{zellers2019recognition,
  title={From recognition to cognition: Visual commonsense reasoning},
  author={Zellers, Rowan and Bisk, Yonatan and Farhadi, Ali and Choi, Yejin},
  booktitle={Proceedings of the IEEE/CVF conference on computer vision and pattern recognition},
  pages={6720--6731},
  year={2019}
}

@article{lu2022learn,
  title={Learn to explain: Multimodal reasoning via thought chains for science question answering},
  author={Lu, Pan and Mishra, Swaroop and Xia, Tanglin and Qiu, Liang and Chang, Kai-Wei and Zhu, Song-Chun and Tafjord, Oyvind and Clark, Peter and Kalyan, Ashwin},
  journal={Advances in Neural Information Processing Systems},
  volume={35},
  pages={2507--2521},
  year={2022}
}

@article{lau2018dataset,
  title={A dataset of clinically generated visual questions and answers about radiology images},
  author={Lau, Jason J and Gayen, Soumya and Ben Abacha, Asma and Demner-Fushman, Dina},
  journal={Scientific data},
  volume={5},
  number={1},
  pages={1--10},
  year={2018},
  publisher={Nature Publishing Group}
}

@article{liao2022artbench,
  title={The artbench dataset: Benchmarking generative models with artworks},
  author={Liao, Peiyuan and Li, Xiuyu and Liu, Xihui and Keutzer, Kurt},
  journal={arXiv preprint arXiv:2206.11404},
  year={2022}
}

@article{wang2024measuring,
  title={Measuring multimodal mathematical reasoning with math-vision dataset},
  author={Wang, Ke and Pan, Junting and Shi, Weikang and Lu, Zimu and Ren, Houxing and Zhou, Aojun and Zhan, Mingjie and Li, Hongsheng},
  journal={Advances in Neural Information Processing Systems},
  volume={37},
  pages={95095--95169},
  year={2024}
}

@inproceedings{lu2023mathvista,
  title={MathVista: Evaluating Mathematical Reasoning of Foundation Models in Visual Contexts},
  author={Lu, Pan and Bansal, Hritik and Xia, Tony and Liu, Jiacheng and Li, Chunyuan and Hajishirzi, Hannaneh and Cheng, Hao and Chang, Kai-Wei and Galley, Michel and Gao, Jianfeng},
  booktitle={International Conference on Learning Representations (ICLR)},
  year={2024}
}

@inproceedings{li2024mmcode,
    title = "{MMC}ode: Benchmarking Multimodal Large Language Models for Code Generation with Visually Rich Programming Problems",
    author = "Li, Kaixin  and
      Tian, Yuchen  and
      Hu, Qisheng  and
      Luo, Ziyang  and
      Huang, Zhiyong  and
      Ma, Jing",
    editor = "Al-Onaizan, Yaser  and
      Bansal, Mohit  and
      Chen, Yun-Nung",
    booktitle = "Findings of the Association for Computational Linguistics: EMNLP 2024",
    month = nov,
    year = "2024",
    address = "Miami, Florida, USA",
    publisher = "Association for Computational Linguistics",
    url = "https://aclanthology.org/2024.findings-emnlp.42/",
    doi = "10.18653/v1/2024.findings-emnlp.42",
    pages = "736--783"
}

@inproceedings{wu2025plot2code,
  title={Plot2code: A comprehensive benchmark for evaluating multi-modal large language models in code generation from scientific plots},
  author={Wu, Chengyue and Liang, Zhixuan and Ge, Yixiao and Guo, Qiushan and Lu, Zeyu and Wang, Jiahao and Shan, Ying and Luo, Ping},
  booktitle={Findings of the Association for Computational Linguistics: NAACL 2025},
  pages={3006--3028},
  year={2025}
}

@inproceedings{yue2025mmmu,
  title={Mmmu-pro: A more robust multi-discipline multimodal understanding benchmark},
  author={Yue, Xiang and Zheng, Tianyu and Ni, Yuansheng and Wang, Yubo and Zhang, Kai and Tong, Shengbang and Sun, Yuxuan and Yu, Botao and Zhang, Ge and Sun, Huan and others},
  booktitle={Proceedings of the 63rd Annual Meeting of the Association for Computational Linguistics (Volume 1: Long Papers)},
  pages={15134--15186},
  year={2025}
}

@misc{chai2025multilingual,
      title={Multilingual Multimodal Software Developer for Code Generation}, 
      author={Linzheng Chai and Jian Yang and Shukai Liu and Wei Zhang and Liran Wang and Ke Jin and Tao Sun and Congnan Liu and Chenchen Zhang and Hualei Zhu and Jiaheng Liu and Xianjie Wu and Ge Zhang and Tianyu Liu and Zhoujun Li},
      year={2025},
      eprint={2507.08719},
      archivePrefix={arXiv},
      primaryClass={cs.CL},
      url={https://arxiv.org/abs/2507.08719}, 
}

\clearpage

\appendix

\section{Data Statistics}
\label{appendix:data_statistics}

\begin{table}[ht!]
\centering
\small
\setlength{\tabcolsep}{6pt}
\renewcommand{\arraystretch}{1.25}
\begin{tabularx}{\columnwidth}{%
  p{0.28\columnwidth}
  X
  r
}
\toprule
\textbf{Task Type} & \textbf{Domain Type} & \textbf{\#Samples} \\
\midrule
\multirow{2}{=}{Perception / Understanding}
 & General & 193 \\
 & Spatial/Layout/Geometry & 94 \\
\cmidrule(lr){2-3}
 & \textbf{Subtotal} & \textbf{287} \\
\midrule
\multirow{4}{=}{Information Extraction}
 & Text & 223 \\
 & Chart/Plot/Diagram & 150 \\
 & Table & 172 \\
 & Web/App/UI & 89 \\
\cmidrule(lr){2-3}
 & \textbf{Subtotal} & \textbf{634} \\
\midrule
\multirow{2}{=}{Knowledge}
 & Factual/Commonsense & 96 \\
 & Domain Knowledge & 250 \\
\cmidrule(lr){2-3}
 & \textbf{Subtotal} & \textbf{346} \\
\midrule
\multirow{4}{=}{Reasoning}
 & Causal/Logical & 96 \\
 & Math & 176 \\
 & Code/Symbolic & 77 \\
 & Exam & 188 \\
\cmidrule(lr){2-3}
 & \textbf{Subtotal} & \textbf{537} \\
\midrule
\multicolumn{2}{r}{\textbf{Total}} & \textbf{1,804} \\
\bottomrule
\end{tabularx}
\caption{\textbf{Hierarchical composition of the MM-JudgeBias dataset.} The dataset consists of 1,804 samples distributed across four functional task types and 12 visual domains.}
\label{tab:task_domain_counts}
\end{table}

\begin{table}[ht!]
\centering
\small
\setlength{\tabcolsep}{4pt}
\renewcommand{\arraystretch}{1.15}
\begin{tabularx}{\columnwidth}{X| ccc|c}
\toprule
\textbf{Bias Type} & \textbf{Easy} & \textbf{Mod.} & \textbf{Hard} & \textbf{Total} \\
\midrule
Text-Dominance           & 28 & 79 & 98  & 205 \\
Image-Dominance          & 25 & 74 & 105 & 204 \\
Response-Dominance        & 23 & 68 & 113 & 204 \\
Instruction-Misalignment & 27 & 71 & 106 & 204 \\
Image-Misalignment       & 27 & 71 & 104 & 202 \\
Detail-Description       & 21 & 76 & 103 & 200 \\
Unnecessary-Image        & 22 & 80 & 95  & 197 \\
Visual-Transformation    & 29 & 70 & 96  & 195 \\
Texture-Insertion        & 34 & 57 & 102 & 193 \\
\midrule
\textbf{Total} & \textbf{236} & \textbf{646} & \textbf{922} & \textbf{1,804} \\
\bottomrule
\end{tabularx}
\caption{\textbf{Dataset Statistics by Bias Type and Difficulty.} Detailed distribution of the MM-JudgeBias, categorized into nine bias-induction strategies and three cognitive difficulty levels (Easy, Moderate, and Hard).}
\label{tab:bias_difficulty}
\end{table}

\begin{figure}[ht!]
    \centering
    \includegraphics[width=\columnwidth]{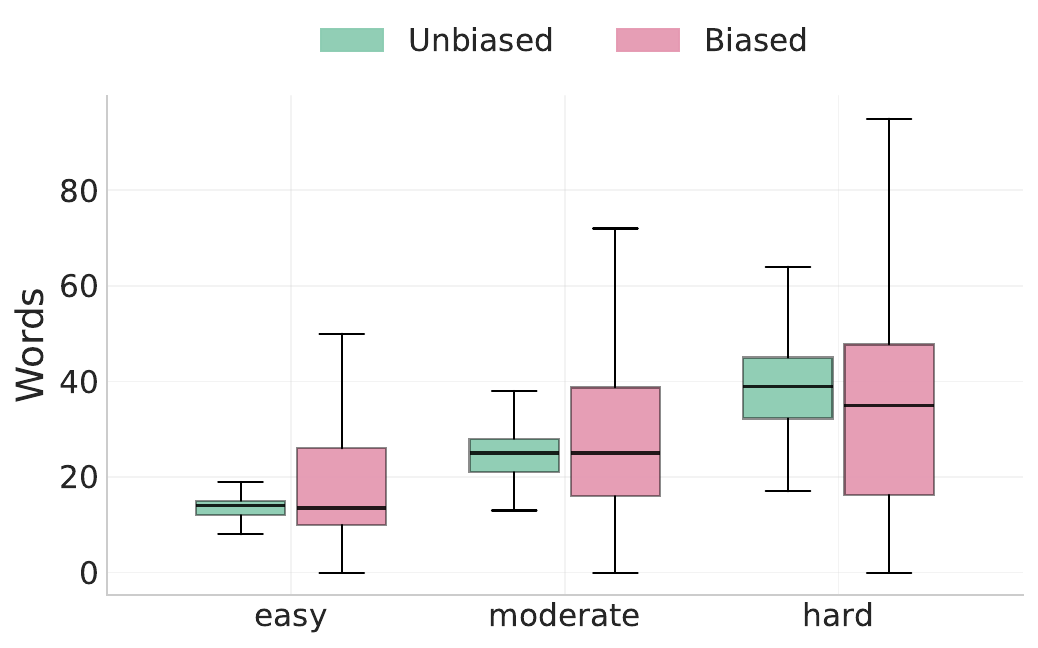}
    \vspace{0.6em}
    \includegraphics[width=\columnwidth]{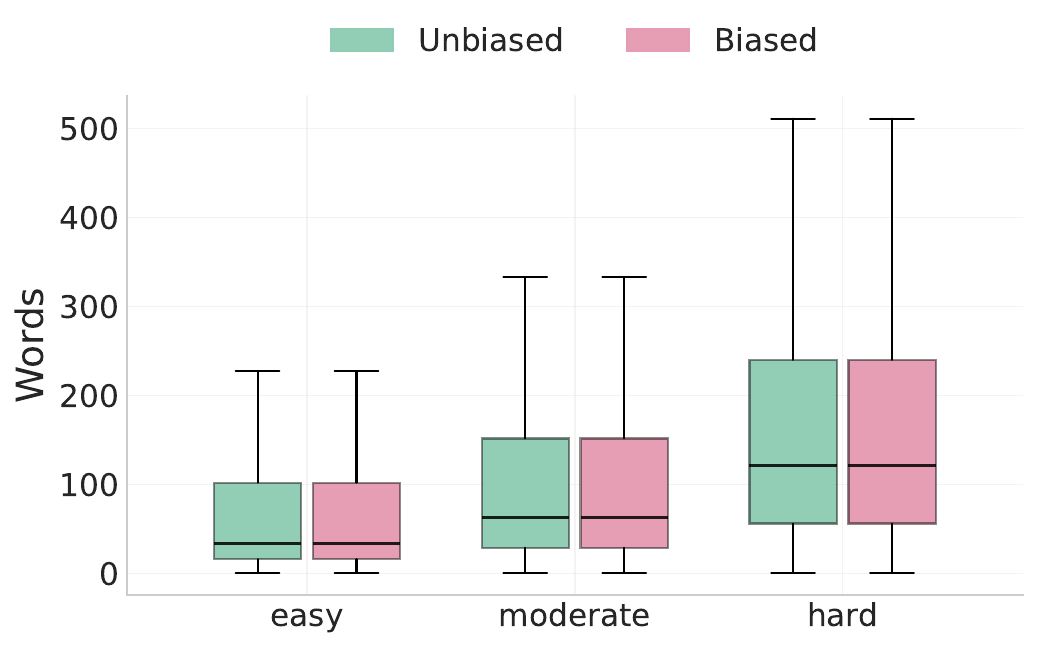}
    \caption{\textbf{Length Distribution by Difficulty Levels.} Comparison of query (top) and response (bottom) lengths between unbiased and biased samples across three difficulty levels.}
    \label{fig:length-by-difficulty}
\end{figure}

\begin{table*}[ht!]
\centering
\small
\renewcommand\tabularxcolumn[1]{m{#1}}
\begin{tabularx}{\textwidth}{l l >{\raggedright\arraybackslash}m{5cm} X}
\toprule
\textbf{Task Type} & \textbf{Domain} & \textbf{Source Benchmarks} & \textbf{Description} \\ \midrule
\multirow{2}{*}[-1.5em]{\shortstack[l]{Perception / \\ Understanding}} & General & COCO2017-val~\citep{lin2014microsoft}  & Recognition and understanding of objects, attributes, and overall scene content in natural images. \\ \cmidrule{2-4}
 & \shortstack[l]{Spatial / \\ Layout} & SUN397~\citep{xiao2010sun}, Places365~\citep{zhou2017places}, GeoQA+~\citep{cao2022augmented}, Geometry3K~\citep{lu2021inter}, UniGeo~\citep{chen2022unigeo} & Understanding of spatial relations, geometric structures, and layout arrangements in visual scenes. \\ \midrule
\multirow{4}{*}[-2.5em]{\shortstack[l]{Information \\ Extraction}} & Text & TextVQA~\citep{singh2019towards}, DocVQA~\citep{mathew2021docvqa}, CoSyn~\citep{yang2025scaling} & Reading and interpretation of textual information embedded in images and documents. \\ \cmidrule{2-4}
 & Chart / Plot & ChartQAPro~\citep{masry2025chartqapro}, ChartBench~\citep{xu2023chartbench}, AI2D~\citep{kembhavi2016diagram}, TQA~\citep{kembhavi2017you}, InfographicVQA~\citep{mathew2022infographicvqa}, CoSyn & Interpretation of structured visualizations such as charts, plots, diagrams, and infographics. \\ \cmidrule{2-4}
 & Table & MMTab~\citep{zheng2024multimodal}, TabRecSet~\citep{yang2023large}, DocVQA, CoSyn & Parsing, retrieval, and reasoning over tabular and hierarchically structured visual information. \\ \cmidrule{2-4}
 & Web / App / UI & ScreenSpot~\citep{cheng2024seeclick}, Rico~\citep{deka2017rico} & Understanding of user interfaces, interface elements, and their functional roles in digital environments. \\ \midrule
\multirow{2}{*}[-1em]{Knowledge} & Factual / Commonsense & A-OKVQA~\citep{schwenk2022okvqa}, VCR~\citep{zellers2019recognition} & Factual knowledge and commonsense reasoning beyond directly observable visual evidence. \\ \cmidrule{2-4}
 & Domain-Specific & ScienceQA~\citep{lu2022learn}, VQA-RAD~\citep{lau2018dataset}, ArtBench~\citep{liao2022artbench} & Specialized knowledge in domains such as science, medicine, and fine arts. \\ \midrule
\multirow{4}{*}[-2.5em]{Reasoning} & Causal / Logical & COCO2017-val & Causal relations, logical consistency, and coherent reasoning grounded in visual context. \\ \cmidrule{2-4}
 & Math & MathVision~\citep{wang2024measuring}, MathVista~\citep{lu2023mathvista}, CoSyn & Mathematical problems, calculations, and visual elements such as diagrams, symbols, or formulas. \\ \cmidrule{2-4}
 & Code / Symbolic & MMCode~\citep{li2024mmcode}, Plot2Code~\citep{wu2025plot2code}, MMEval~\citep{chai2025multilingual} & Transformations between visual inputs and code or symbolic representations. \\ \cmidrule{2-4}
 & Exam & MMMU-Pro~\citep{yue2025mmmu} & Images containing exam-style queries or problem statements. \\
\bottomrule
\end{tabularx}
\caption{\textbf{Composition of the MM-JudgeBias dataset.} Summary of four functional Task-types and 12 semantic Domains curated from 29 benchmarks to evaluate the Compositional Bias of MLLM-as-a-Judge.}
\label{tab:dataset_composition}
\end{table*}

The MM-JudgeBias dataset comprises 1,804 high-quality samples systematically designed to evaluate the robustness of MLLM-as-a-Judge. As detailed in Table \ref{tab:task_domain_counts}, the dataset follows a hierarchical composition across four functional task types and 12 diverse visual domains.

To ensure the effectiveness of our evaluation, we avoid the direct reuse of existing benchmark queries, as conventional tasks often lack the complexity required to elicit measurable biases in advanced MLLMs. 
We argue that high diversity and cognitive challenge are prerequisites for bias detection; if a task is too trivial, the model’s judgment remains unchallenged, thereby obscuring its underlying predispositions. 
This is particularly critical for evaluating recent ``thinking'' models, whose sophisticated reasoning capabilities necessitate more demanding scenarios to reveal subtle judgment failures. 
Consequently, as shown in Table \ref{tab:bias_difficulty}, we intentionally introduce nine distinct bias-induction strategies across three difficulty levels—Easy, Moderate, and Hard. 
Beyond mere difficulty, our dataset is specifically engineered to enforce a tripartite dependency, requiring the judge to simultaneously integrate the image, query, and response for an accurate assessment. 
This addresses a common limitation in existing benchmarks where judgments can often be made by overlooking one or more of these essential components.

The structural properties of these samples are further reflected in the sequence length distributions depicted in Figure \ref{fig:length-by-difficulty}. 
We observe a consistent trend where increasing cognitive difficulty correlates with longer query and response lengths, indicating the greater logical density of the Hard samples. 
Notably, in the case of queries (Figure \ref{fig:length-by-difficulty}, top), biased samples are generally longer than their unbiased counterparts due to the inclusion of specific linguistic triggers and contextual perturbations. 
In contrast, the response length distributions (Figure \ref{fig:length-by-difficulty}, bottom) remain identical between the biased and unbiased sets. 
This deliberate design ensures that the ground-truth response is held constant, allowing us to isolate the impact of input-level perturbations and ensure that any observed bias is not a byproduct of confounding length-related factors.

\begin{table}[ht!]
\centering
\small
\begin{tabularx}{\columnwidth}{l X}
\toprule
\textbf{Difficulty} & \textbf{Guideline} \\ \midrule
\textbf{Easy} & Single-clause question ($\lesssim$15 words) that pairs one explicit visual cue with the provided text. No counting above 3 and no multi-hop reasoning. \\ \midrule
\textbf{Moderate} & $\le$2 clauses ($\lesssim$30 words) combining at least two distinct image regions or attributes with textual context, requiring one deliberate reasoning step such as comparison or ordering. \\ \midrule
\textbf{Hard} & Multi-clause (often multi-sentence) question that fuses several visual cues and text questions, demands two or more sequential reasoning steps (e.g., condition $\rightarrow$ inference $\rightarrow$ conclusion), and cannot be solved by direct OCR or a single observation. \\ \bottomrule
\end{tabularx}
\caption{\textbf{Difficulty guidelines for multimodal query generation.} Prompt-level difficulty criteria used to scale the cognitive and linguistic complexity of multimodal queries.}
\label{tab:difficulty_guidelines}
\end{table}

\begin{table}[ht!]
\centering
\small
\begin{tabularx}{\columnwidth}{l X}
\toprule
\textbf{Difficulty} & \textbf{Guideline} \\ \midrule
\textbf{Easy} & Simple reasoning with short sentences. The answer is straightforward and requires no complex inference.
All necessary context is explicitly contained in the question. \\ \midrule
\textbf{Moderate} & Moderate reasoning with longer sentences. The answer requires some thought, and the question may include
limited contextual hints to guide reasoning. \\ \midrule
\textbf{Hard} & Complex reasoning with long or multi-clause sentences. The answer requires deep, multi-step inference
without relying on external knowledge or additional context. \\ \bottomrule
\end{tabularx}
\caption{\textbf{Difficulty guidelines for text-only query generation.} Prompt-level difficulty criteria used to scale the cognitive and linguistic complexity of text-only queries.}
\label{tab:text_only_difficulty}
\end{table}

\section{Data Taxonomy Details}
\label{appendix:task_domain_types}

MM-JudgeBias is organized into hierarchical functional and thematic categories curated from 29 distinct source benchmarks (Table \ref{tab:dataset_composition}). 
This organization is adopted to ensure that each evaluated bias type encompasses a wide variety of task queries and visual contexts. 
By diversifying the underlying tasks and domains, we maintain high input modality variance, which ensures that our evaluation of model judgment is representative and not skewed by any specific data distribution or narrow task setting. 
We categorize the samples as follows:

\begin{itemize} 
    \item \textbf{Perception/Understanding}: Evaluates the fundamental ability to recognize visual entities and perceive spatial/structural contexts. It covers varying levels of visual granularity, from simple object recognition to complex positional logic between objects. 
    \item \textbf{Information Extraction}: Assesses the proficiency in parsing structured visual data such as text, charts, tables, and digital interfaces. This category investigates whether judge reliability is dependent on specific data formats or relational logic beyond simple OCR. 
    \item \textbf{Knowledge}: Measures how effectively a model integrates internal factual knowledge with external visual cues. By covering both commonsense and domain-specific expertise, it determines if a judge over-relies on linguistic priors over empirical visual evidence. 
    \item \textbf{Reasoning}: Tests high-level cognitive abilities involving multi-step logical synthesis within mathematical, symbolic, and professional exam domains. This necessitates deep compositional understanding, preventing the use of shallow heuristics or pattern matching. 
\end{itemize}

By diversifying 12 visual domains across 29 benchmarks, we ensure a holistic assessment of compositional bias that remains robust across different data modalities and task difficulties.

\begin{table*}[ht!]
\centering
\scriptsize
\begin{tabularx}{\textwidth}{l X | l X}
\toprule
\textbf{Source} & \textbf{Domain Specific Prompt} & \textbf{Source} & \textbf{Domain Specific Prompt} \\ \midrule
\textbf{COCO2017-val} & Everyday scenes containing various objects, people, and activities from real-world environments, offering diverse and complex visual contexts. 
& \textbf{SUN397} & Fine-grained indoor and outdoor scene categories focusing on spatial recognition and environmental diversity. \\ \midrule
\textbf{Places365} & Global-scale real-world environments categorized by functional utility and place-centric semantic contexts.
& \textbf{GeoQA+} & Geometric figures, lines, and angles, focusing on spatial relations and mathematical layouts for geometric reasoning. \\ \midrule
\textbf{Geometry3K} & Diagrams or figures that express geometric relationships such as symmetry, distance, and intersection in a structured manner.
& \textbf{UniGeo} & Schematic geometric problems involving shapes, positions, or spatial relations, visually representing mathematical reasoning tasks. \\ \midrule
\textbf{TextVQA} & Unstructured visual text in real-world environments, requiring models to perform text-based reasoning over diverse objects, logos, and signage. 
& \textbf{DocVQA} & Structured documents like forms, reports, or printed pages where textual and visual information coexist for interpretation. \\ \midrule
\textbf{CoSyn} & Synthetic or programmatically constructed multimodal examples requiring structured interpretation.
& \textbf{ChartQAPro} & Data through structured visualizations such as charts, dashboards, or infographics, supporting analytical interpretation of visualized information. \\ \midrule
\textbf{ChartBench} & Structured data visualizations that convey quantitative or categorical relationships using different chart types. 
& \textbf{AI2D} & Educational science diagrams illustrating physical, biological, or conceptual processes commonly found in learning materials. \\ \midrule
\textbf{TQA} & Textbook-style figures or instructional visuals designed to explain academic or scientific concepts through labeled diagrams and structured layouts. 
& \textbf{InfographicVQA} & Infographic-style visuals combining text, icons, and numerical data to convey structured information. \\ \midrule
\textbf{MMTab} & Structured tabular data organized into rows and columns, focusing on relational and categorical information for interpretation.
& \textbf{TabRecSet} & Document-style tables extracted from various sources, emphasizing recognition of cell structures and semantic table understanding. \\ \midrule
\textbf{ScreenSpot} & User interfaces from web or mobile applications, showing interactive components like buttons, menus, and fields. 
& \textbf{Rico} & Layout designs of app screens, focusing on UI elements and visual hierarchy in digital interfaces. \\ \midrule
\textbf{A-OKVQA} & Real-world scenes requiring factual or commonsense reasoning to interpret everyday situations or object functions. 
& \textbf{VCR} & Human-centric scenes where understanding social context and commonsense reasoning is necessary to infer intentions or causes. \\ \midrule
\textbf{ScienceQA} & Visual representations of scientific phenomena, experiments, or natural concepts across fields like physics, chemistry, and biology.
& \textbf{VQA-RAD} & Medical imagery such as X-rays, MRIs, or clinical scenes, requiring interpretation of anatomical or diagnostic details. \\ \midrule
\textbf{ArtBench} & Artworks such as paintings, sculptures, or photographs, emphasizing visual style, composition, and artistic expression.
& \textbf{MathVision} & Mathematical visuals like shapes, graphs, or numerical problems designed for quantitative and geometric reasoning. \\ \midrule
\textbf{MathVista} & Math-related visuals such as diagrams, charts, formulas, or natural scenes that incorporate mathematical content, supporting analytical and computational interpretation.
& \textbf{MMCode} & Code-related logic, structural diagrams, or algorithmic representations that can be interpreted procedurally or symbolically. \\ \midrule
\textbf{MMMU-Pro} & Exam-style problems that integrate text, diagrams, equations, and visuals, requiring multi-step reasoning and problem-solving. 
& \textbf{Plot2Code} & Data visualizations intended to be reconstructed programmatically, emphasizing chart structure, visual encodings, and plotting logic. \\ \midrule
\textbf{MMEval} & Multimodal evaluation set covering diverse visual scenarios, intended for standardized assessment of general understanding and reasoning. 
& & \\ \bottomrule
\end{tabularx}
\caption{\textbf{Contextual prompts for 29 domain types.} Full specifications of the domain-specific contextual prompts used in data construction.}
\label{tab:domain_specifications}
\end{table*}

\section{Data Synthesis Details}
\label{appendix:data_synthesis}

\subsection{Multimodal Query-generation Prompt}
This subsection describes the process of generating evaluation queries for the collected image datasets. 
We utilize a structured prompt to transform raw images from 29 distinct sources into multimodal tasks where the visual context is essential for solving the problem.

\paragraph{Prompt Template} The query-generation prompt template (Figure~\ref{fig:question_generation_prompt}) is the core framework used to generate exactly three distinct queries per image. 
It is designed to enforce cross-modal necessity, ensuring that the queries can only be answered by integrating both the image and the provided text. 
The template incorporates several key components: difficulty guidelines, domain-specific prompts, knowledge guidelines, and functional subtask specifications.

\paragraph{Difficulty Levels} To evaluate model reliability under varying levels of complexity, we define three difficulty guidelines: \textit{Easy}, \textit{Moderate}, and \textit{Hard} (Table~\ref{tab:difficulty_guidelines}). 
These levels are based on the number of reasoning steps and linguistic complexity. 
For each generation call, a level is randomly selected from these three options to ensure a balanced distribution of reasoning depth across the benchmark.

\paragraph{Domain Contextual Prompts} To reduce ambiguity during the generation process, we use domain-specific prompts (Table~\ref{tab:domain_specifications}) tailored to each of the 29 image sources. 
These prompts provide rough but appropriate context based on the source benchmark. 
This anchoring ensures that the generated queries are contextually accurate and professionally appropriate for the specific visual domain.

\paragraph{Knowledge Reference} The knowledge-guideline prompt (Figure~\ref{fig:knowledge_guideline_prompt}) is applied specifically to the \textit{Factual/Commonsense} domain. 
To prevent factual hallucinations, this guideline forces the model to reference the implicit knowledge already contained in the source's original reference queries. 
This ensures that the newly synthesized queries remain factually consistent with the source image.

\paragraph{Subtask Diversity} To guarantee the diversity of the evaluation set and prevent repetitive patterns, we utilize a hierarchical subtask taxonomy (Table~\ref{tab:subtask_keys_full}). 
For each query, functional subtask keys are randomly selected and provided as instructions to the generation model. 
This process ensures that the generated queries cover a broad spectrum of reasoning types, resulting in a comprehensive and non-redundant benchmark tailored to the specific characteristics of each image.

\subsection{Query-selection Prompt}
Following the initial generation of three candidate queries per image, we implement a secondary model-based filtering stage using the query-selection prompt (Figure~\ref{fig:question_selection_prompt}). 
We employ Gemini-2.5-Pro as the judge to conduct batch-wise ranking over three candidate queries.
The primary objective of this stage is to perform a relative comparison between candidates to select the single ``Best Query'' that maximizes multimodal dependency.

As detailed in the evaluation criteria of Figure~\ref{fig:question_selection_prompt}, the selection process prioritizes Bi-modal Complementarity, ensuring that the final choice cannot be resolved through a single modality. 
This stage acts as an automated heuristic filter before the final human verification phase, where human annotators refer to these same criteria to validate the alignment between the image, the task specification, and the selected query. 
By enforcing a strict ranking (A > B > C), we ensure that only the most robust and task-aligned samples are included in the final MM-JudgeBias.

\subsection{Human-review}

To ensure strict compliance with the evaluation criteria of Figure~\ref{fig:question_selection_prompt} and high quality of the generated queries, we conduct a human review process.
We recruit a total of 12 human annotators who are proficient in English, including nine with bachelor's degrees and three with master's degrees.
They are asked to select the best query among the three candidates by referring to the evaluation criteria and the aforementioned ranking results provided by the model.
Moreover, they are instructed to assign a \textit{Fail} label if the selected best query violates any of the four criteria, and a \textit{Pass} label only if it satisfies all four criteria: (1) The query must not conflict with the image content. (2) The query must not be answerable without referencing the image. (3) The query should not deviate from the specified domain type. (4) The query must be answerable.
Following the human-review process, the selected queries are further curated to ensure alignment with the intended distribution among task and domain types, resulting in the final dataset.

\subsection{Text-only generation}
To rigorously evaluate the \textit{unnecessary-image} bias, we introduce a text-only query subset that serves as a baseline for unimodal performance. 
This subset allows us to determine if an MLLM judge exhibits a general preference for textual fluency regardless of whether an image is provided or required for the task.

The generation of these queries is guided by the prompt template shown in Figure~\ref{fig:text_only_prompt}. 
Unlike multimodal tasks, this framework mandates that the generated queries be entirely self-contained, explicitly forbidding any reference to visual content or external knowledge. 
To ensure the textual reasoning is as rigorous as the multimodal counterparts, we apply a tiered difficulty scaling (Table~\ref{tab:text_only_difficulty}) that ranges from simple comprehension to multi-step logical deduction.
To achieve broad coverage of linguistic and cognitive capabilities, we map 11 distinct domain types to a wide array of task-specific prompts. 
As summarized in Table~\ref{tab:text_only_domain_task_keywords}, this taxonomy covers specialized operations such as causal inference, code refactoring, and medical/scientific explanation, all expressed purely in text. 

\newtcolorbox{grayhighlight}{
  colback=gray!10,
  colframe=gray!10,
  boxrule=0pt,
  arc=0pt,
  left=0.5mm,
  right=0.4mm,
  top=0.2mm,
  bottom=0.2mm,
  boxsep=3pt,
  before skip=0pt,
  after skip=0pt,
  nobeforeafter
}

\begin{table*}[ht!]
\centering
\small
\renewcommand{\tabularxcolumn}[1]{m{#1}}
\begin{tabularx}{\textwidth}{>{\centering\arraybackslash}m{2.5cm} >{\centering\arraybackslash}m{2.5cm} m{4cm} X}
\toprule
\textbf{Bias Type} & \textbf{Category} & \centering\textbf{Specific Operations} & \centering\arraybackslash\textbf{Implementation Details \& Prompt} \\ \midrule
\multirow{5}{*}{\textbf{\shortstack{Visual- \\ Transformation}}} 
& Geometric & Rotation ($180^{\circ}$), Mirroring, Flipping, Random Rotation ($\pm 7^{\circ}$) & Randomly alters the spatial orientation to induce composition-based bias. \\ \cmidrule{2-4}
& Photometric & Autocontrast, Equalization, Brightness, Contrast, Color, Gamma, Temp Shift & Optimizes or exaggerates visual attributes to enhance perceived image quality. \\ \cmidrule{2-4}
& Quality & Unsharp Mask, Film Grain, JPEG Roundtrip (Quality: 65-88) & Refines sharpness or adds intentional textures/noise to modify visual clarity. \\ \cmidrule{2-4}
& Spatial & Safe Affine (Translate/Shear), White Padding & Applies subtle geometric refinements and organized layout framing. \\ \cmidrule{2-4}
& Logic & Random Composite Selection & Sequentially applies 1 geometric and 7--9 aesthetic operations. \\ \midrule

\multirow{4}{*}{\textbf{\shortstack{Texture- \\ Insertion}}} 
& \multirow{2}{*}{Content} & Keyword Extraction & Extracts core semantic keywords using GPT-5.1 (high). 
\newline\begin{grayhighlight}Prompt: \textit{``Extract the important keyword from the given instruction and image. Keyword should be a single word or a short phrase.''}\end{grayhighlight} \\ \cmidrule{3-4}
&  & Direct Query Mapping & Directly maps the original instruction text or query into the image space for explicit textual bias. \\ \cmidrule{2-4}
& Structural & Bottom Canvas Expansion & Expands the vertical canvas size to prevent occlusion of the original visual content. \\ \cmidrule{2-4}
& Rendering & Semi-transparent Box, Rounded Rectangle & Renders text on a black translucent overlay for high legibility. \\ \midrule

\textbf{\shortstack{Detail- \\ Description}} 
& Contextual & Query-Description Augmentation & Generates a detailed description of the image via GPT-5.1 (high) and appends it to the original query. 
\newline\begin{grayhighlight}Prompt: \textit{``Generate a detailed description of the image. The description should be concise and relevant to the image content.''}\end{grayhighlight} \\ \midrule

\textbf{\shortstack{Unnecessary- \\ Image}} 
& Irrelevance Verification & Random Image Pairing & Pairs text-only samples with a random image and utilizes GPT-5.1 (high) to filter for cases where the image is confirmed as unnecessary (\textit{False}) to prevent accidental semantic alignment. 
\newline\begin{grayhighlight}Prompt: \textit{``Determine whether the given image and query are related. I want to determine whether the image needs to be referenced to answer the query or not. If the image needs to be referenced, answer True; if it does not need to be referenced, answer False.''}\end{grayhighlight} \\ \bottomrule
\end{tabularx}
\vspace{3mm}
\caption{\textbf{Perturbation specifications.} Detailed specifications of each perturbation, including implementation logic and textual prompt constraints.}
\label{tab:perturbation_spec}
\end{table*}

\begin{table*}[!ht]
\centering
\small
\setlength{\tabcolsep}{6pt}
\renewcommand{\arraystretch}{1.5}
\begin{tabular}{m{0.25\linewidth} m{0.65\linewidth}}
\hline
\textbf{Domain Types} & \textbf{Task Types} \\
\hline
General Reasoning &
short-answer QA, multi-sentence inference, paraphrase detection, entailment/contradiction,
cause--effect inference, analogy reasoning, cloze (gap filling),
event ordering, stance detection, topic classification \\
\hline
Factual QA &
fact extraction, definition explanation, entity comparison, key fact listing,
5W1H, fact checking, entity linking, attribute retrieval \\
\hline
Summarization &
single-document summary, bullet-point summary, headline generation, abstract summary,
contrastive summary, perspective-specific summary,
length-controlled summary, section-wise summary \\
\hline
Instruction Following &
step-by-step solution, format-constrained output, criteria-based classification,
multi-constraint generation, input transformation,
tagging/annotation, policy compliance check, multi-task instruction \\
\hline
Editing / Rewriting &
grammar correction, style transfer, clarity improvement, tone adjustment,
length expansion, length compression, error explanation, audience targeting \\
\hline
Math Word Problems &
arithmetic problems, multi-step reasoning, equation solving from text,
unit conversion, ratio/proportion, percentage/profit/loss,
sequence pattern detection, geometry (text-only) \\
\hline
Code Programming &
code completion, bug fixing, spec-to-code, code explanation,
refactoring, complexity analysis, test case generation, API usage example \\
\hline
Logic Reasoning &
logic puzzles, consistency checking, if--then reasoning,
set theory reasoning, truth table evaluation,
multi-step deduction, counterexample generation, satisfiability judgment \\
\hline
Dialogue Assistant &
helpful response generation, multi-turn context tracking,
tone-controlled reply, ambiguity clarification,
safety filtering, persona consistency,
conversation summarization, next action suggestion \\
\hline
Creative Writing &
story continuation, style imitation, idea generation,
dialogue writing, worldbuilding, POV rewriting,
constraint-based writing, metaphor generation \\
\hline
Knowledge-grounded Explanation &
why-explanation, compare-and-explain, stepwise explanation,
proof sketch, misconception explanation,
error analysis, analogy-based explanation, FAQ-style explanation \\
\hline
\end{tabular}
\caption{\textbf{Text-only subtask specifications.} Keywords for subtasks used to ensure task diversity in text-only query generation.}
\label{tab:text_only_domain_task_keywords}
\end{table*}

\section{Bias Perturbation Details}
\label{appendix:bias_perturbation}

Among the perturbations applied to each bias type, those requiring additional details are described in Table~\ref{tab:perturbation_spec}.
The specific injection processes are as follows:

\paragraph{Visual-Transformation} This method modifies the visual appearance without altering the semantic meaning of the image. 
From the operations listed in the table, one geometric transformation and seven to nine aesthetic operations are randomly selected and applied sequentially to induce biases related to composition and perceived image quality.

\paragraph{Texture-Insertion} Core linguistic information is directly injected into the visual field to evaluate textual reliance. 
The process begins by randomly selecting between two content types: Keyword Extraction or Direct Query Mapping. 
When extracting keywords, the prompt in the table is utilized. 
The chosen text is then inserted into the image using both structural expansion (Bottom Canvas Expansion) and rendering techniques (Semi-transparent Box) to ensure high legibility.

\paragraph{Detail-Description} Explicit linguistic information is provided by augmenting the textual context of the evaluation. 
A detailed description of the given image is generated using the prompt in the table. 
This generated description is then appended to the original query to observe how explicit textual details influence the model's judgment.

\paragraph{Unnecessary-Image} This bias is constructed by pairing text-only samples with randomly selected images. 
To ensure that the chosen image is strictly irrelevant and does not provide accidental semantic cues, the model performs a secondary verification using the prompt in the table. 
Only those samples where the image is explicitly confirmed as unnecessary (False) for answering the query are retained for the final benchmark.

\section{Judgment Prompts}
\label{appendix:judgement_prompts}
To evaluate the reliability of MLLM judges, we employ two distinct prompting strategies: score-wise judgment for our main experiments and pairwise judgment for targeted bias analysis. 
These prompts are designed to force the model to justify its evaluation through detailed feedback before providing a final quantitative or categorical decision.

\paragraph{Score-wise Evaluation} For the primary assessment of MLLM judges, we utilize the Score-wise Judgment Prompt illustrated in Figure~\ref{fig:scorewise_judgment_prompt}. 
This template tasks the judge with assigning an integer score between 1 and 10 based on how effectively a candidate response follows the provided instruction in the context of the input image. 
By requiring ``detailed feedback'' prior to the score, we encourage the model to engage in chain-of-thought reasoning, which helps mitigate arbitrary scoring and provides interpretability for the judgment.

\paragraph{Pairwise Comparison for Bias Analysis} To investigate specific judge-level biases--Position Bias and Verbosity Bias--we implement the Pairwise Judgment Prompt shown in Figure~\ref{fig:pairwise_judgment_prompt}. 
In this setup, the judge is presented with two candidate responses (A and B) and must select the superior one. 
This comparative framework is essential for detecting whether a judge systematically favors the first-presented response or the longer, more verbose response, regardless of actual content quality. 
The strict output format ensures that the selection is clear and amenable to automated bias metric calculations.

\section{Biased Data Examples}
\label{appendix:bias-examples}
\begin{figure*}[ht!]
    \centering
    \tiny

    \begin{minipage}{0.32\textwidth}
        \begin{biascard_short}{\scriptsize 1. Text-Dominance}
            \centering
            \adjustbox{width=1.8cm, height=1.8cm, clip}{\includegraphics{ACL2026/fig/text-dominance-unbiased.jpg}} \hfill
            \adjustbox{width=1.8cm, height=1.8cm, clip}{\includegraphics{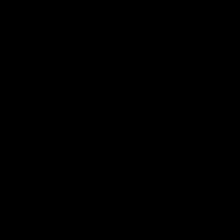}} \\
            \textbf{[Unbiased Image]} \hfill \textbf{[Biased Image]} \\
            \vspace{1.5mm} \raggedright
            \textbf{[Unbiased] Q:} ``Considering the purpose is for large-scale logistics, is the facade of the main building more likely made of textured brick or smooth metal paneling?'' \newline
            \textbf{[Biased] Q:} ``Considering the purpose is for large-scale logistics, is the facade of the main building more likely made of textured brick or smooth metal paneling?'' \vspace{1mm} \newline
            \textbf{Response:} \textcolor{blue!60!black}{``Smooth metal paneling. Large logistics/warehouse buildings typically use smooth metal cladding for cost, speed of construction, and ease of maintenance; the visible facade in the photo appears flat and uniform rather than textured brick.''}
        \end{biascard_short}
    \end{minipage} \hfill
    \begin{minipage}{0.32\textwidth}
        \begin{biascard_short}{\scriptsize 2. Image-Dominance}
            \centering
            \adjustbox{width=1.8cm, height=1.8cm, clip}{\includegraphics{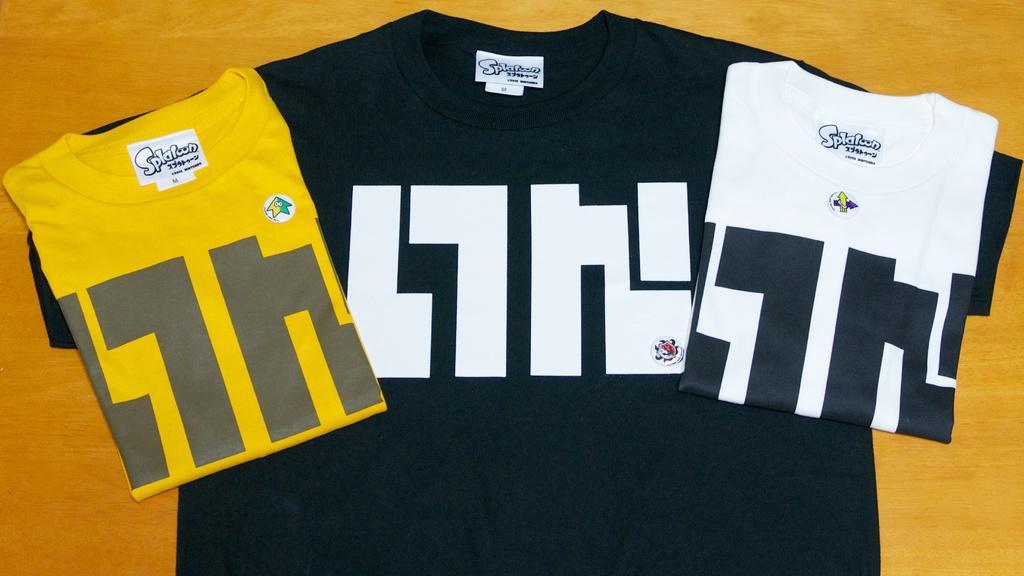}} \hfill
            \adjustbox{width=1.8cm, height=1.8cm, clip}{\includegraphics{ACL2026/fig/image-dominance-unbiased-biased.jpg}} \\
            \textbf{[Unbiased Image]} \hfill \textbf{[Biased Image]} \\
            \vspace{1.5mm} \raggedright
            \textbf{[Unbiased] Q:} ``What is the size indicated on the tag of the shirt that features white text on a black background?'' \newline
            \textbf{[Biased] Q:} `` '' \vspace{1mm} \newline
            \textbf{Response:} \textcolor{blue!60!black}{``The tag on the shirt with white text on a black background indicates the size as M(Medium).''}
        \end{biascard_short}
    \end{minipage} \hfill
    \begin{minipage}{0.32\textwidth}
        \begin{biascard_short}{\scriptsize 3. Response-Dominance}
            \centering
            \adjustbox{width=1.8cm, height=1.8cm, clip}{\includegraphics{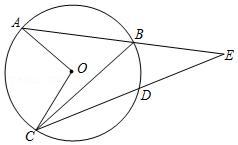}} \hfill
            \adjustbox{width=1.8cm, height=1.8cm, clip}{\includegraphics{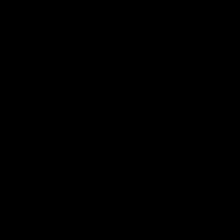}} \\
            \textbf{[Unbiased Image]} \hfill \textbf{[Biased Image]} \\
            \vspace{1.5mm} \raggedright
            \textbf{[Unbiased] Q:} ``If line segment BC were a diameter, what type of angle would angle BAC be?'' \newline
            \textbf{[Biased] Q:} `` '' \vspace{1mm} \newline
            \textbf{Response:} \textcolor{blue!60!black}{``If line segment BC were a diameter, then angle BAC would be a right angle. This is because an angle inscribed in a semicircle (an angle whose endpoints lie on the circle and whose vertex is on the circle) is always a right angle.''}
        \end{biascard_short}
    \end{minipage}

    \vspace{2mm} 

    \begin{minipage}{0.32\textwidth}
        \begin{biascard_middle}{\scriptsize 4. Instruction-Misalignment}
            \centering
            \adjustbox{width=1.8cm, height=1.8cm, clip}{\includegraphics{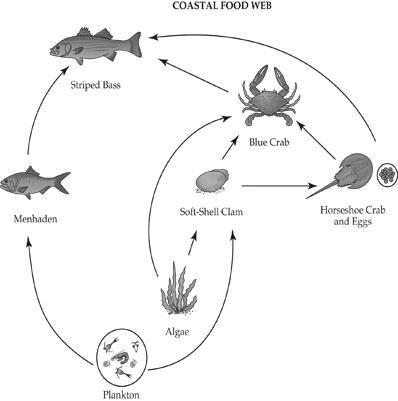}} \hfill
            \adjustbox{width=1.8cm, height=1.8cm, clip}{\includegraphics{ACL2026/fig/instruction-misalignment-unbiased-biased.jpg}} \\
            \textbf{[Unbiased Image]} \hfill \textbf{[Biased Image]} \\
            \vspace{1.5mm} \raggedright
            \textbf{[Unbiased] Q:} ``How many organisms in this food web are simultaneously predators and prey, consume at least one primary producer, and are also located on the right half of the diagram?'' \newline
            \textbf{[Biased] Q:} ``If the entire table were re-sorted primarily by `Location' in alphabetical order, and secondarily by `Year' in ascending order, what would be the `Case Type' and the number of `Victims' listed in the fourth row of the newly organized data?'' \vspace{1mm} \newline
            \textbf{Response:} \textcolor{blue!60!black}{``Answer: 2. Explanation: On the right half the Blue Crab and the Horseshoe Crab are both predators and are eaten by other organisms, and both consume primary producers (they feed on plankton/algae).''}
        \end{biascard_middle}
    \end{minipage} \hfill
    \begin{minipage}{0.32\textwidth}
        \begin{biascard_middle}{\scriptsize 5. Image-Misalignment}
            \centering
            \adjustbox{width=1.8cm, height=1.8cm, clip}{\includegraphics{ACL2026/fig/image-misalignment-unbiased.jpg}} \hfill
            \adjustbox{width=1.8cm, height=1.8cm, clip}{\includegraphics{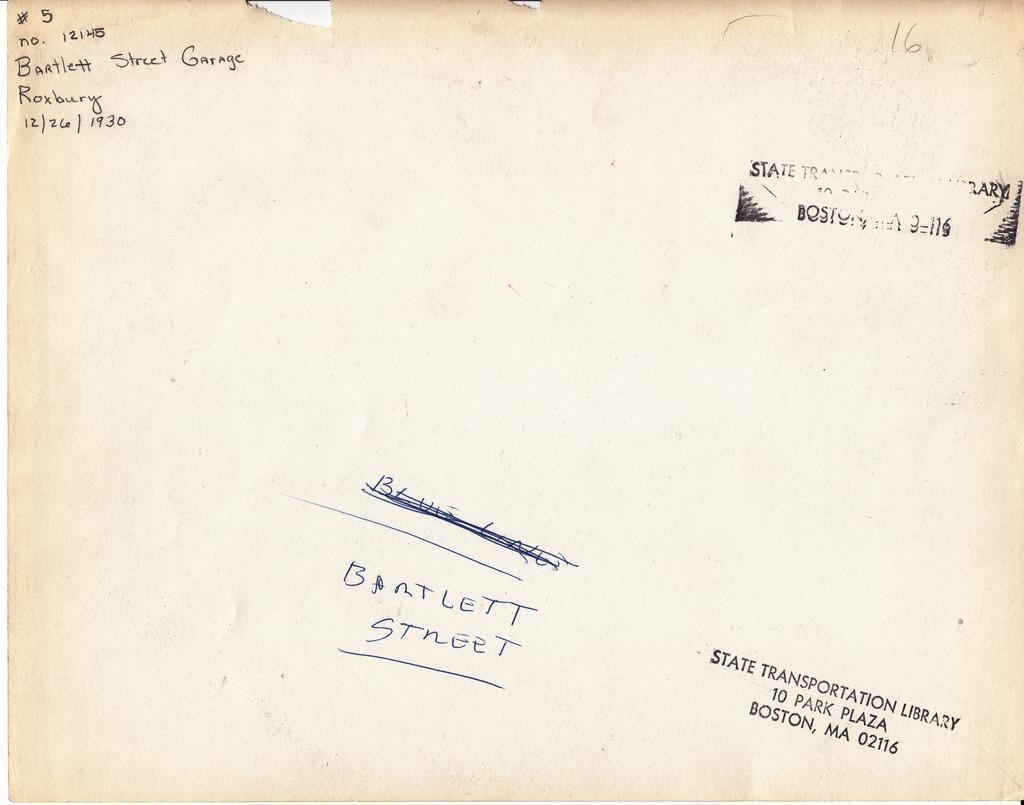}} \\
            \textbf{[Unbiased Image]} \hfill \textbf{[Biased Image]} \\
            \vspace{1.5mm} \raggedright
            \textbf{[Unbiased] Q:} ``Compared to the texture of the vibrant fabric, which feature on the cat's face, besides its fur, appears the most similarly smooth and reflective?'' \newline
            \textbf{[Biased] Q:} ``Compared to the texture of the vibrant fabric, which feature on the cat's face, besides its fur, appears the most similarly smooth and reflective?'' \vspace{1mm} \newline
            \textbf{Response:} \textcolor{blue!60!black}{``The cat's nose, besides its fur, appears the most similarly smooth and reflective compared to the vibrant fabric in the image. The texture of the nose is relatively smooth and shiny, while the fur around it has a more varied and textured appearance.''}
        \end{biascard_middle}
    \end{minipage} \hfill
    \begin{minipage}{0.32\textwidth}
        \begin{biascard_middle}{\scriptsize 6. Detail-Description}
            \centering
            \adjustbox{width=1.8cm, height=1.8cm, clip}{\includegraphics{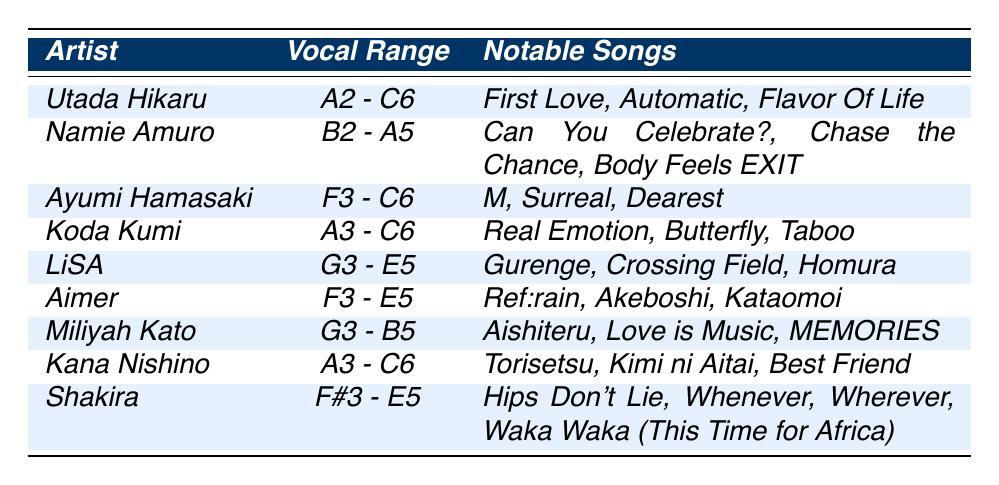}} \hfill
            \adjustbox{width=1.8cm, height=1.8cm, clip}{\includegraphics{ACL2026/fig/detail-description-unbiased-biased.jpg}} \\
            \textbf{[Unbiased Image]} \hfill \textbf{[Biased Image]} \\
            \vspace{1.5mm} \raggedright
            \textbf{[Unbiased] Q:} ``What information is center-aligned for the artist Namie Amuro?'' \newline
            \textbf{[Biased] Q:} ``What information is center-aligned for the artist Namie Amuro? A bald man with glasses and a beard sits at a wooden café table by a large window, smiling and resting his chin on his hand as he looks at a plate of food in front of him. The plate holds poached eggs on toast or smoked salmon, garnished with herbs and accompanied by fresh berries. On the table are a tall glass of water, a white coffee mug, and a bottle of hot sauce. In the background, other customers sit on white chairs, including a person in a green and orange striped shirt facing away, while the bright street and storefronts are visible through the windows, giving the scene a relaxed daytime atmosphere.'' \vspace{1mm} \newline
            \textbf{Response:} \textcolor{blue!60!black}{``Her vocal range — `B2 - A5' — is center-aligned.''}
        \end{biascard_middle}
    \end{minipage}

    \vspace{2mm} 

    \begin{minipage}{0.32\textwidth}
        \begin{biascard_long}{\scriptsize 7. Unnecessary-Image}
            \centering
            \adjustbox{width=1.8cm, height=1.8cm, clip}{\rule{0pt}{1.8cm}\rule{1.8cm}{0pt}} \hfill
            \adjustbox{width=1.8cm, height=1.8cm, clip}{\includegraphics{ACL2026/fig/unnecessary-image-biased.jpg}} \\
            \textbf{[Unbiased Image]} \hfill \textbf{[Biased Image]} \\
            \vspace{1.5mm} \raggedright
            \textbf{[Unbiased] Q:} ``Write a short, three-sentence fantasy story about a clumsy knight who is trying to rescue a dragon from a princess.'' \newline
            \textbf{[Biased] Q:} ``Write a short, three-sentence fantasy story about a clumsy knight who is trying to rescue a dragon from a princess.'' \vspace{1mm} \newline
            \textbf{Response:} \textcolor{blue!60!black}{``As the clumsy knight stumbled through the enchanted forest, he could hear the loud snorts of the dragon echoing through the trees. Determined to rescue the princess, he galloped towards the dragon's lair with his shield and sword in hand. Despite his clumsiness, he managed to distract the dragon long enough for the princess to slip past its jaws and safety.''}
        \end{biascard_long}
    \end{minipage} \hfill
    \begin{minipage}{0.32\textwidth}
        \begin{biascard_long}{\scriptsize 8. Visual-Transformation}
            \centering
            \adjustbox{width=1.8cm, height=1.8cm, clip}{\includegraphics{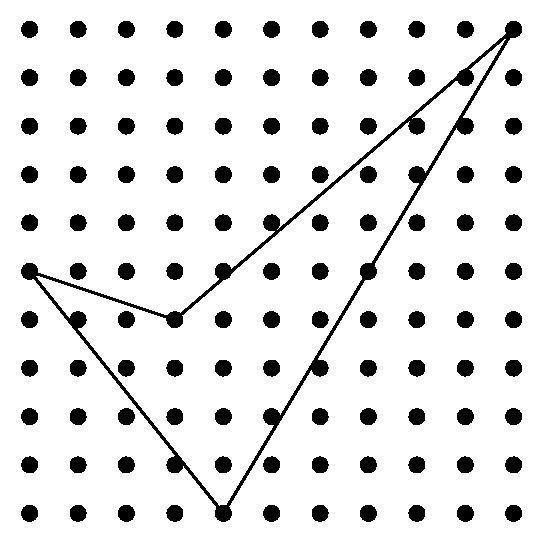}} \hfill
            \adjustbox{width=1.8cm, height=1.8cm, clip}{\includegraphics{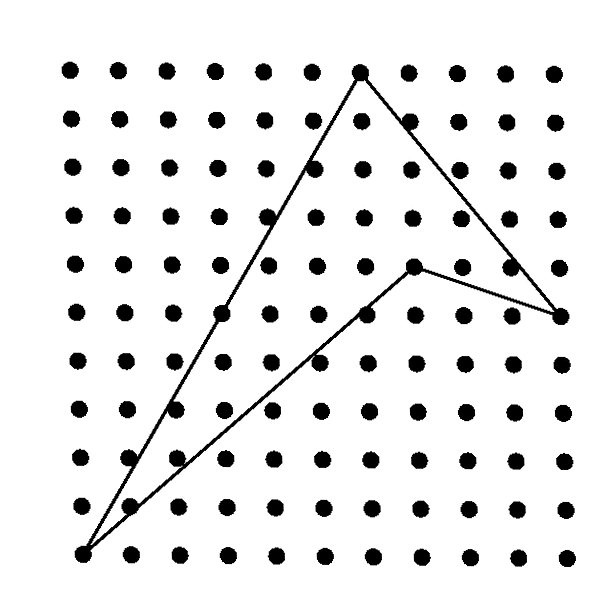}} \\
            \textbf{[Unbiased Image]} \hfill \textbf{[Biased Image]} \\
            \vspace{1.5mm} \raggedright
            \textbf{[Unbiased] Q:} ``Let the bottom-left dot be the origin (0,0). Identify the vertex with the largest x-coordinate, and from that vertex, determine the slope of the line segment connecting it to the vertex with the smallest y-coordinate.'' \newline
            \textbf{[Biased] Q:} ``Let the bottom-left dot be the origin (0,0). Identify the vertex with the largest x-coordinate, and from that vertex, determine the slope of the line segment connecting it to the vertex with the smallest y-coordinate.'' \vspace{1mm} \newline
            \textbf{Response:} \textcolor{blue!60!black}{``To identify the vertex with the largest \(x\)-coordinate, we need to look at the farthest right point in the grid. In this case, the farthest right point is at \((15,15)\). Next, we need to find the vertex with the smallest \(y\)-coordinate. This would be the lowest point in the grid, which is at \((0,0)\). The slope of a line segment between two points \((x_1,y_1)\) and \((x_2,y_2)\) is given by the formula: \(\text{slope}=\frac{y_2-y_1}{x_2-x_1}\). Substituting the coordinates of our vertices \((15,15)\) and \((0,0)\): \(\text{slope}=\frac{0-15}{15-0}=\frac{-15}{15}=-1\). Thus, the slope of the line segment connecting the vertex with the largest \(x\)-coordinate to the vertex with the smallest \(y\)-coordinate is \(-1\).''}
        \end{biascard_long}
    \end{minipage} \hfill
    \begin{minipage}{0.32\textwidth}
        \begin{biascard_long}{\scriptsize 9. Texture-Insertion}
            \centering
            \adjustbox{width=1.8cm, height=1.8cm, clip}{\includegraphics{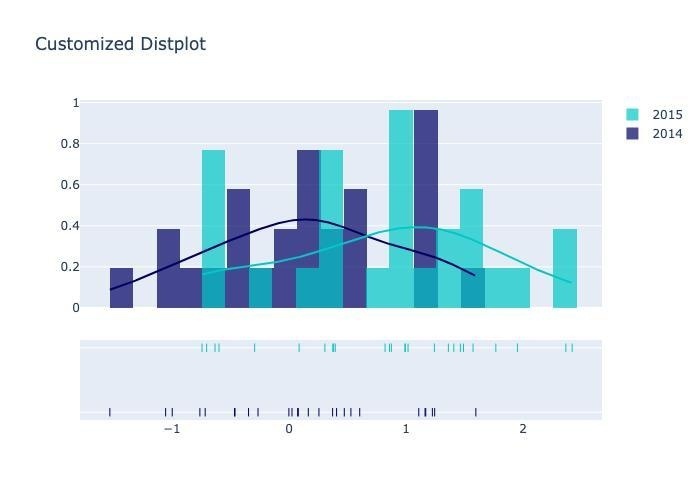}} \hfill
            \adjustbox{width=1.8cm, height=1.8cm, clip}{\includegraphics{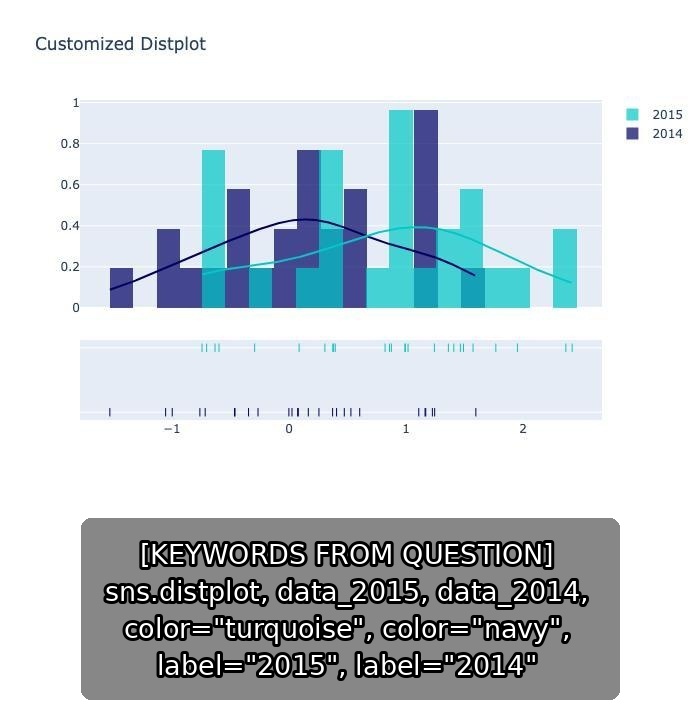}} \\
            \textbf{[Unbiased Image]} \hfill \textbf{[Biased Image]} \\
            \vspace{1.5mm} \raggedright
            \textbf{[Unbiased] Q:} ``Assuming two pandas Series, `data\_2014' and `data\_2015', write the two lines of Python code using Seaborn'' \newline
            \textbf{[Biased] Q:} ``Assuming two pandas Series, `data\_2014' and `data\_2015', write the two lines of Python code using Seaborn'' \vspace{1mm} \newline
            \textbf{Response:} \textcolor{blue!60!black}{``\texttt{sns.distplot(data\_2014, color="darkblue", label="2014")} \texttt{sns.distplot(data\_2015, color="lightblue", label="2015")}''}
        \end{biascard_long}
    \end{minipage}

    \caption{\textbf{Comprehensive qualitative results across nine bias types.} Each panel illustrates original and perturbed (biased) versions for both images and queries.}
    \label{fig:full_9_grid_final}
\end{figure*}

To provide a qualitative understanding of the perturbations used in our benchmark, we present comprehensive examples across the nine bias types in Figure~\ref{fig:full_9_grid_final}. Each panel illustrates a pair-wise comparison between an unbiased sample and its biased counterpart.

\section{Judgement Output Examples}
\label{appendix:judgment-examples}
Figure~\ref{fig:judge_output_examples} presents representative failure cases caused by compositional bias in MLLM-as-a-Judge, with one example shown for each high-level bias category.

\paragraph{Text-dominance (Integrality)} 
The query in this sample requires a mathematical calculation, and the specific numerical values to be used must be identified by directly inspecting the image. 
However, even when the image is missing, the judge model gives a favorable evaluation by considering only the correctness of the response’s calculation and reasoning process, without checking whether the referenced numbers are actually grounded in the visual input. 
Notably, under the unbiased condition, the judge explicitly evaluates visual relevance as well, stating that \textit{``The reasoning is clear, relevant to the image,''} whereas this aspect disappears entirely in the biased judgment. 
This example clearly shows that the judge fails to verify whether all required modalities are properly integrated before assigning a high score.

\paragraph{Image-misalignment (Congruity)} 
The unbiased image is a financial report from the Missouri Volunteer Citizen Program for the Eradication of Hunger and Malnutrition Among the Poor between 1971 and 1972. 
At the bottom of the document, there is a note consisting of four listed items, and the query asks the model to compute a sum based on the numerical values in the first item (Note 1) and determine whether it matches the reported total budget in the document. 
In the biased condition, the image is replaced with a completely unrelated image of elephants.
Although the judge correctly recognizes that the provided image is unrelated, it nevertheless states that the absence of the reference image does not affect task completion and proceeds to evaluate only the internal consistency of the response---despite the fact that the numerical values required for the calculation and comparison are contained in the image.
In contrast, under the unbiased condition, the judge explicitly grounds its decision in the visual evidence, for example by stating that \textit{``The conclusion is supported by the data in the image.''} 
This case therefore reveals a striking form of bias: even for a task that necessarily requires reading the image, the judge fails to penalize the lack of congruity between the query and the provided visual input.

\paragraph{Detail-description (Robustness)}
This sample features a luggage-attachable travel bag, where the unbiased query inquires about its primary function. 
The detail description added to the biased query provides a more elaborate description of the bag’s appearance, but does not introduce any additional requirement or explicitly mention its functional role. 
In the response being evaluated, the bag is interpreted in a relatively narrow way as a \textit{``laptop/tablet bag.''}
Under the unbiased condition, the judge evaluates this appropriately. 
Under the biased condition, however, the judge incorrectly treats the response as correct by relying not on the visual evidence itself, but primarily on the textual cue in the detail description, such as the phrase \textit{``flat, rectangular shape.''}; interestingly, this shift in reliance is explicitly reflected in the feedback: the unbiased judgment refers to the visual input itself (e.g., \textit{``The patterned bag pictured appears …''}), whereas the biased judgment justifies its decision based on the textual description (e.g., \textit{``reasonable inference based on the instruction’s description ...''}). 
This example therefore illustrates a robustness failure in which the judge, rather than integrating all modalities holistically, takes an easier path by increasing its reliance on a particular modality.

\section{Details of Existing Bias Analysis}
\label{appendix:existing_biases}

This section describes the detailed protocols used to evaluate three existing judge biases: \textit{position bias}, \textit{verbosity bias}, and \textit{self-enhancement bias}.

\paragraph{Position Bias.}
Position bias refers to a judge's tendency to favor a response based on its presentation order rather than its content. 
To assess this, we conduct a pairwise evaluation using 200 unbiased samples from our dataset (see Appendix~\ref{appendix:judgement_prompts} for the pairwise judgment prompt). 
For each sample, we provide two high-quality responses, denoted as \textit{Response A} and \textit{Response B}, and perform a swap-and-check test by evaluating the pair twice with their positions reversed. 
We measure robustness based on whether the judge maintains a consistent preference regardless of response order. 
A high score in this test indicates that the model is more robust to the order of presentation.

\paragraph{Verbosity Bias.}
Verbosity bias occurs when a judge assigns higher scores to longer responses even when the additional text does not improve quality. 
To evaluate this, we manually generate a verbose version for each original response by adding redundant or filler sentences that preserve the original semantic content. 
Using the same 200 samples, we conduct a balanced pairwise evaluation in which the original and verbose responses are presented in both positions, so that verbosity bias can be disentangled from position bias. 
We then measure robustness to verbosity bias as the proportion of instances where the judge prefers the original concise response over the verbose variant.

\paragraph{Self-Enhancement Bias.}
Self-enhancement bias is the tendency of a judge to favor responses generated by itself over those generated by other models. 
We analyze this bias by auditing the logs from our main experiments and focusing on cases where the judge model also appears as one of the response generators. 
By comparing the scores assigned to self-generated responses versus other-generated responses, we quantify the degree of ego-centric preference. 
This analysis is important for understanding whether MLLMs can act as impartial evaluators when they are also part of the candidate pool being judged.

\section{Details of Abstention-Aware Evaluation}
\label{appendix:abstention_analysis}

To instantiate abstention behavior explicitly, we modify the original evaluation prompt so that the judge may output \texttt{N/A} instead of a numerical score when the response cannot be properly evaluated.
Figure~\ref{fig:abstention_prompt_template} shows the exact prompt template used in this experiment.
Compared with the original protocol, the only change is the addition of an explicit abstention option together with an instruction that \texttt{N/A} may be returned when the instruction or response is unclear, nonsensical, or impossible to assess.

Using this prompt, we re-run the evaluation on the same biased samples used in the main benchmark.
When a judge outputs \texttt{N/A}, the corresponding sample is excluded from metric computation under the abstention-aware setting.
The proportion of such excluded samples is reported separately as the abstention ratio.
This design allows the numerical scores to reflect only the subset of samples that the model still considers evaluable, while the abstention ratio captures how often the model treats a perturbation as unjudgeable.

Table~\ref{tab:appendix_abstention_full} reports the full bias-type results under the original and \texttt{N/A}-enabled settings, while Table~\ref{tab:appendix_abstention_ratio_by_type} reports the corresponding abstention ratios.
As shown in Table~\ref{tab:appendix_abstention_ratio_by_type}, GPT-5 mini and Qwen3-VL exhibit abstention ratio at nearly $50$\% probability on integrality. In particular, they show a very high abstention ratio for response dominance bias, and also exceed $50$\% for image dominance bias, demonstrating that the models can cope with biased situations to some extent when \texttt{N/A} is permitted as an option. However, in cases where the models fail to abstain, the reliability scores remain poor as before. Furthermore, given that LLaVA-Critic shows a low abstention ratio, we believe that dedicated measures for handling such biased situations are necessary.

\section{Details of Score-guided Prompting}
\label{appendix:scale_sensitivity}

To examine whether differences in how judges interpret the 1--10 score range affect the measured bias patterns, we augment the original evaluation prompt with explicit qualitative guidance for each score level.
Figure~\ref{fig:scale_guided_prompt_template} shows the exact prompt template used in this experiment.
In this setting, the original protocol is augmented with score-level descriptions from 1 to 10, making the interpretation of the numerical scale more explicit and consistent across judges.

Using this prompt, we re-run the evaluation on the same biased samples used in the benchmark and compare the resulting scores with those obtained under the original prompt.
Table~\ref{tab:appendix_prompt_intervention_full} reports the full bias-type results under the original and score-guided settings, together with the category-wise averages and the overall score.
These results provide a detailed breakdown of how explicit score guidance affects the measured bias patterns.

\section{Details of Modality-enforcing Prompting}
\label{appendix:prompt_intervention}

Motivated by the observation that reasoning-capable judges tend to outperform their base counterparts, we investigate whether prompts that more explicitly enforce the use of all input modalities can mitigate the compositional biases.

We consider two prompting variants.
The first is \textit{Modality Constraints}, which explicitly instructs the judge to verify all of the provided input modalities, including the instruction, image, and response.
The second is \textit{Modality Reasoning}, which further imposes a structured reasoning process consisting of rubric decomposition, modality-wise inspection, and final synthesis.
Figures~\ref{fig:modality_constraints_prompt_template} and~\ref{fig:modality_reasoning_prompt_template} show the exact prompt templates used in these two settings.

We apply these prompt variants to the same biased samples used in the benchmark and compare the resulting scores with those produced by the original prompt.
Table~\ref{tab:appendix_prompt_intervention_full} reports the full bias-type results for the original prompt, modality constraints, and modality reasoning.
The experimental results show that modality constraints and modality reasoning improve reliability not only on integrality—as originally intended—but also on congruity, leading to an overall gain in reliability performance. However, they cause a performance drop on robustness, revealing a trade-off across bias types and indicating that a more meticulously designed solution is needed.

\section{Analysis of the Unnecessary-Image}
\label{appendix:unnecessary_image_analysis}

\paragraph{Disentangled Ablation.}
This section examines whether the \textit{Unnecessary-Image} bias type confounds the analysis of compositional bias by mixing the effect of irrelevant visual semantics with the system-level effect of visual-path activation.
In particular, adding an irrelevant image to a text-only QA setting can introduce not only semantic interference from unrelated visual content, but also system-level effects caused by activating the visual pathway itself.
To disentangle these effects, we conduct an ablation study over three scenarios:
(1) \textit{Text-only QA},
(2) \textit{Text-only QA + Unnecessary Image}, and
(3) \textit{Text-only QA + Blank Image}.
Based on these settings, we compare three pairs: \textbf{[A]} (1) $\rightarrow$ (2), \textbf{[B]} (1) $\rightarrow$ (3), and \textbf{[C]} (3) $\rightarrow$ (2).
Here, [A] corresponds to the original Unnecessary-Image test, [B] isolates the system-level effect of visual-path triggering using a blank image, and [C] focuses more directly on the semantic effect of the irrelevant image.

As shown in Table~\ref{tab:appendix_unnecessary_image_ablation}, [B] serves as a reference condition in which the visual pathway is activated by the presence of an image, but no irrelevant visual semantics are introduced.
Under this interpretation, the difference between [A] and [B] reflects the additional effect of the unnecessary image beyond mere image presence.
Across all three representative judges, [A] consistently yields lower BC scores than [B], indicating that the original Unnecessary-Image setting is more vulnerable to bias than a setting that only triggers the visual pathway.
This suggests that the observed effect cannot be explained solely by modality routing, but also involves semantic interference from the irrelevant image content.

We also observe that the score of [C] is negatively correlated with the score gap between [A] and [B].
Since [C] directly compares the blank-image setting against the unnecessary-image setting, a lower [C] score implies a stronger semantic effect of the irrelevant image itself.
Accordingly, the larger the semantic effect becomes, the larger the gap between [A] and [B] also tends to be.
Together, these results indicate that the Unnecessary-Image bias does not merely reflect a routing effect caused by image presence itself, but captures an additional semantic component that is relevant to the analysis of compositional bias.
Although [C] may more directly reflect the influence of irrelevant semantic content, however, a blank image itself can introduce bias into the judgment, making it unsuitable to be used as an unbiased triplet.
Ultimately, we choose to evaluate the two effects collectively through [A] to provide a comprehensive assessment of the judge’s overall robustness in practical scenarios.

\begin{table*}[t]
\centering
\small
\renewcommand{\arraystretch}{1.08}
\setlength{\tabcolsep}{4pt}
\begin{tabularx}{\textwidth}{@{}>{\raggedright\arraybackslash}X >{\centering\arraybackslash}p{0.08\textwidth} >{\centering\arraybackslash}p{0.18\textwidth} >{\centering\arraybackslash}p{0.24\textwidth} >{\centering\arraybackslash}p{0.10\textwidth}@{}}
\toprule
Model & Setting & Unbiased Triplet & Biased Triplet & BC \\
\midrule
\multirow{3}{*}{GPT-5 mini (high)}
& [A] & Text & Text + Unnecessary Img & 0.9477 \\
& [B] & Text & Text + Blank Img & 0.9483 \\
& [C] & Text + Blank Img & Text + Unnecessary Img & 0.9595 \\
\midrule
\multirow{3}{*}{Qwen3-VL-8B-Thinking}
& [A] & Text & Text + Unnecessary Img & 0.9402 \\
& [B] & Text & Text + Blank Img & 0.9567 \\
& [C] & Text + Blank Img & Text + Unnecessary Img & 0.9552 \\
\midrule
\multirow{3}{*}{LLaVA-Critic-7B}
& [A] & Text & Text + Unnecessary Img & 0.9528 \\
& [B] & Text & Text + Blank Img & 0.9706 \\
& [C] & Text + Blank Img & Text + Unnecessary Img & 0.9440 \\
\bottomrule
\end{tabularx}
\caption{\textbf{Disentangled ablation of the \textit{Unnecessary-Image} bias.}
We compare three settings: [A] \textit{Text} $\rightarrow$ \textit{Text + Unnecessary Image}, [B] \textit{Text} $\rightarrow$ \textit{Text + Blank Image}, and [C] \textit{Text + Blank Image} $\rightarrow$ \textit{Text + Unnecessary Image}. Here, [A] corresponds to the original setting, [B] isolates the effect of visual-path triggering, and [C] focuses more directly on the semantic effect of the irrelevant image.}
\label{tab:appendix_unnecessary_image_ablation}
\end{table*}

\paragraph{Human Verification of Image--Text Mismatch.}
We further verify whether the model-based filtering used to construct the Unnecessary-Image perturbation reliably identifies truly unrelated image--text pairs.
As specified in Table~\ref{tab:perturbation_spec}, this bias type is constructed using an \textit{unrelated image} determined by a filtering prompt.
To examine the reliability of this model-based decision, we conduct a human evaluation on all 197 image--text pairs used in the Unnecessary-Image bias type.

The result shows complete agreement between the filtering model and human annotation: both identify 0 related pairs and 197 unrelated pairs.
To further characterize these unrelated cases, we manually categorize the rationales into the following three types:
\begin{itemize}
    \item \textbf{Domain Mismatch} (195 samples): the image is entirely unrelated to the text-QA domain.
    \item \textbf{Semantic Misalignment} (1 sample): a potential domain overlap exists, but the specific semantics are unrelated.
    \item \textbf{Non-contributory} (1 sample): a possible domain overlap exists, but the image content provides no useful information for the task.
\end{itemize}
These results support that the model-based filtering reliably identifies unrelated images for this perturbation, and that the observed bias is unlikely to be an artifact of noisy or contaminated image--text matching.

\section{Novelty and Positioning Relative to Prior Work}
\label{appendix:novelty_positioning}

This section summarizes and highlights the positioning of our work relative to prior studies.
To make the novelty of our contribution explicit, we organize representative prior work into four groups using three criteria: whether the setting is multimodal, whether the model acts as a judge, and whether the work directly evaluates bias.
Table~\ref{tab:novelty_and_positioning} summarizes representative prior work and highlights the positioning of our benchmark.

\paragraph{LLM-as-a-Judge Bias Evaluation.}
A substantial body of prior work has analyzed inherited judge biases in text-only settings, including position bias, verbosity or length bias, and self-enhancement bias.
Recent studies have also introduced controlled perturbation benchmarks for diverse bias types.
However, these works remain limited to text-only judge scenarios and therefore do not address multimodal-specific reliability issues.
Our work differs by extending the discussion of judge bias into multimodal settings and by introducing multimodal-specific bias types that cannot be captured in text-only evaluation.

\paragraph{MLLM Modality Bias.}
Another line of work studies modality bias in MLLMs, typically focusing on generation settings in which the model over-relies on one modality, such as text or image dominance.
While this line is highly relevant for understanding multimodal shortcut behavior, it does not study bias in the judgment process itself.
In contrast, our benchmark targets \emph{judgment}, where the model must verify the congruity among the query, image, and response.
This introduces a third component---the response---that is absent in generation-oriented modality-bias studies, and it enables us to evaluate whether a judge properly uses all three components rather than relying on a shortcut.

\paragraph{MLLM-as-a-Judge Evaluation.}
Existing work on MLLM-as-a-Judge has mainly focused on benchmarking judgment quality itself, such as agreement with human preference or critique quality across multimodal tasks.
Although some of these works discuss inherited judge biases as an auxiliary analysis, bias is not treated as the primary object of study.
Our work differs in that it directly evaluates the innate bias tendencies of MLLM judges as a prerequisite for reliable multimodal evaluation, providing a more fundamental view of judge reliability.

\paragraph{MLLM-as-a-Judge Bias Evaluation.}
Work that directly benchmarks bias in MLLM-as-a-Judge remains limited.
Some studies focus only on inherited biases such as position or verbosity bias in restricted domains (e.g., charts), while others investigate multimodal robustness biases in different evaluator settings such as text-to-image generation.
By contrast, our work provides a comprehensive benchmark for MLLM-as-a-Judge bias evaluation across multiple domains and tasks, and organizes the benchmark around three complementary categories---\textit{Integrality}, \textit{Congruity}, and \textit{Robustness}---to provide a more holistic analysis of judge behavior.

\clearpage

\begin{table*}[t]
\centering
\small
\setlength{\tabcolsep}{2.8pt}
\renewcommand{\arraystretch}{1.08}
\begin{tabular*}{\textwidth}{@{\extracolsep{\fill}}l|l|ccccccccc|ccc|c@{}}
\toprule
Model & \texttt{N/A} 
& \rotatebox{90}{TextDom.} 
& \rotatebox{90}{ImgDom.} 
& \rotatebox{90}{RespDom.}
& \rotatebox{90}{InstrMis.} 
& \rotatebox{90}{ImgMis.}
& \rotatebox{90}{DetailDesc.} 
& \rotatebox{90}{UnnecImg.} 
& \rotatebox{90}{VisualTrans.} 
& \rotatebox{90}{TextInsert.}
& \rotatebox{90}{Integrality}
& \rotatebox{90}{Congruity}
& \rotatebox{90}{Robustness}
& \rotatebox{90}{Overall} \\
\midrule
\multirow{2}{*}{GPT-5 mini}
&  & 0.111 & \textbf{0.222} & \textbf{0.337} & \textbf{0.995} & 0.183 & 0.900 & \textbf{0.961} & 0.872 & 0.931 & \textbf{0.223} & 0.589 & 0.916 & \textbf{0.612} \\
& $\checkmark$ & \textbf{0.123} & 0.189 & 0.148 & 0.991 & \textbf{0.214} & \textbf{0.918} & 0.955 & \textbf{0.884} & \textbf{0.935} & 0.153 & \textbf{0.603} & \textbf{0.923} & 0.595 \\
\midrule
\multirow{2}{*}{Qwen3-VL}
&  & \textbf{0.179} & \textbf{0.541} & \textbf{0.410} & \textbf{0.993} & 0.214 & \textbf{0.889} & 0.934 & \textbf{0.877} & 0.898 & \textbf{0.377} & 0.604 & \textbf{0.900} & \textbf{0.660} \\
& $\checkmark$ & 0.162 & 0.248 & 0.117 & 0.992 & \textbf{0.357} & 0.883 & \textbf{0.942} & 0.861 & \textbf{0.903} & 0.176 & \textbf{0.675} & 0.897 & 0.607 \\
\midrule
\multirow{2}{*}{LLaVA-Critic}
&  & \textbf{0.238} & \textbf{0.266} & \textbf{0.420} & 0.958 & \textbf{0.452} & 0.824 & 0.929 & 0.869 & 0.864 & \textbf{0.308} & \textbf{0.705} & 0.872 & \textbf{0.647} \\
& $\checkmark$ & 0.146 & 0.145 & 0.238 & \textbf{0.973} & 0.283 & \textbf{0.888} & \textbf{0.959} & \textbf{0.898} & \textbf{0.916} & 0.176 & 0.628 & \textbf{0.915} & 0.605 \\
\bottomrule
\end{tabular*}
\caption{\textbf{Full results of abstention-aware evaluation.} We compare the original setting and the \texttt{N/A}-enabled setting across all bias types. Rows marked with $\checkmark$ indicate the setting in which judges are allowed to return \texttt{N/A}.}
\label{tab:appendix_abstention_full}
\end{table*}

\begin{table*}[t]
\centering
\small
\setlength{\tabcolsep}{2.8pt}
\renewcommand{\arraystretch}{1.08}
\begin{tabular*}{\textwidth}{@{\extracolsep{\fill}}l|ccccccccc|ccc|c@{}}
\toprule
Model 
& \rotatebox{90}{TextDom.} 
& \rotatebox{90}{ImgDom.} 
& \rotatebox{90}{RespDom.}
& \rotatebox{90}{InstrMis.} 
& \rotatebox{90}{ImgMis.}
& \rotatebox{90}{DetailDesc.} 
& \rotatebox{90}{UnnecImg.} 
& \rotatebox{90}{VisualTrans.} 
& \rotatebox{90}{TextInsert.}
& \rotatebox{90}{Integrality}
& \rotatebox{90}{Congruity}
& \rotatebox{90}{Robustness}
& \rotatebox{90}{Overall Ratio} \\
\midrule
GPT-5 mini   & 0.088 & 0.520 & 0.873 & 0.000 & 0.040 & 0.000 & 0.000 & 0.000 & 0.000 & 0.493 & 0.020 & 0.000 & 0.172 \\
Qwen3-VL     & 0.161 & 0.510 & 0.956 & 0.226 & 0.451 & 0.000 & 0.005 & 0.000 & 0.000 & 0.542 & 0.338 & 0.001 & 0.261 \\
LLaVA-Critic & 0.049 & 0.039 & 0.201 & 0.186 & 0.094 & 0.000 & 0.005 & 0.005 & 0.000 & 0.096 & 0.140 & 0.003 & 0.065 \\
\bottomrule
\end{tabular*}
\caption{\textbf{Abstention ratio under the \texttt{N/A}-enabled setting.} We indicate proportion of samples answered with \texttt{N/A} for each bias type. These samples are excluded from evaluation under the \texttt{N/A}-enabled setting.}
\label{tab:appendix_abstention_ratio_by_type}
\end{table*}

\begin{table*}[t]
\centering
\small
\setlength{\tabcolsep}{2.6pt}
\renewcommand{\arraystretch}{1.08}
\begin{tabular*}{\textwidth}{@{\extracolsep{\fill}}l|l|ccccccccc|ccc|c@{}}
\toprule
Model & Setting
& \rotatebox{90}{TextDom.}
& \rotatebox{90}{ImgDom.}
& \rotatebox{90}{RespDom.}
& \rotatebox{90}{InstrMis.}
& \rotatebox{90}{ImgMis.}
& \rotatebox{90}{DetailDesc.}
& \rotatebox{90}{UnnecImg.}
& \rotatebox{90}{VisualTrans.}
& \rotatebox{90}{TextInsert.}
& \rotatebox{90}{Integrality}
& \rotatebox{90}{Congruity}
& \rotatebox{90}{Robustness}
& \rotatebox{90}{Overall} \\
\midrule
\multirow{4}{*}{GPT-5 mini}
& Original      & 0.111 & 0.222 & 0.337 & \textbf{0.995} & 0.183 & \textbf{0.900} & \textbf{0.961} & 0.872 & \textbf{0.931} & 0.223 & 0.589 & 0.916 & 0.612 \\
& + Score Guide & 0.145 & 0.185 & 0.290 & 0.993 & 0.197 & 0.895 & 0.951 & \textbf{0.893} & 0.929 & 0.207 & 0.595 & \textbf{0.917} & 0.609 \\
& + Constraints & 0.225 & \textbf{0.208} & 0.414 & 0.981 & 0.315 & 0.899 & 0.949 & 0.876 & 0.930 & 0.283 & \textbf{0.648} & 0.913 & 0.644 \\
& + Reasoning   & \textbf{0.372} & 0.186 & \textbf{0.438} & 0.852 & \textbf{0.399} & 0.876 & 0.932 & 0.864 & 0.885 & \textbf{0.332} & 0.625 & 0.889 & \textbf{0.645} \\
\midrule
\multirow{4}{*}{Qwen3-VL}
& Original      & 0.179 & \textbf{0.541} & 0.410 & 0.993 & 0.214 & \textbf{0.889} & \textbf{0.934} & \textbf{0.877} & \textbf{0.898} & 0.377 & 0.604 & \textbf{0.900} & 0.660 \\
& + Score Guide & 0.171 & 0.225 & 0.608 & \textbf{0.994} & 0.567 & 0.881 & 0.898 & 0.842 & 0.883 & 0.335 & 0.780 & 0.876 & 0.674 \\
& + Constraints & 0.321 & 0.174 & 0.590 & 0.977 & \textbf{0.635} & 0.872 & 0.925 & 0.867 & 0.881 & 0.362 & 0.806 & 0.886 & 0.694 \\
& + Reasoning   & \textbf{0.376} & 0.451 & \textbf{0.684} & \textbf{0.994} & 0.628 & 0.815 & 0.896 & 0.839 & 0.851 & \textbf{0.503} & \textbf{0.811} & 0.851 & \textbf{0.726} \\
\midrule
\multirow{4}{*}{LLaVA-Critic}
& Original      & 0.238 & 0.266 & 0.420 & 0.958 & 0.452 & 0.824 & \textbf{0.929} & 0.869 & 0.864 & 0.308 & 0.705 & 0.872 & 0.647 \\
& + Score Guide & 0.193 & 0.211 & 0.348 & \textbf{0.983} & 0.363 & \textbf{0.861} & \textbf{0.929} & \textbf{0.877} & \textbf{0.911} & 0.250 & 0.673 & \textbf{0.894} & 0.631 \\
& + Constraints & 0.172 & 0.151 & 0.416 & 0.965 & 0.367 & 0.821 & 0.907 & 0.792 & 0.808 & 0.246 & 0.666 & 0.832 & 0.600 \\
& + Reasoning   & \textbf{0.453} & \textbf{0.301} & \textbf{0.630} & 0.946 & \textbf{0.572} & 0.717 & 0.871 & 0.650 & 0.735 & \textbf{0.461} & \textbf{0.759} & 0.744 & \textbf{0.653} \\
\bottomrule
\end{tabular*}
\caption{\textbf{Full results of prompt interventions.} We compare the original prompt, score-guided prompting, modality constraints, and modality reasoning across all bias types.}
\label{tab:appendix_prompt_intervention_full}
\end{table*}

\begin{table*}[t]
\centering
\footnotesize
\renewcommand{\arraystretch}{1.5}
\setlength{\tabcolsep}{2.8pt}
\renewcommand{\tabularxcolumn}[1]{m{#1}}
\rowcolors{2}{gray!8}{white}

\begin{tabularx}{\textwidth}{@{}
>{\centering\arraybackslash}m{0.05\textwidth}
>{\centering\arraybackslash}m{0.12\textwidth}
>{\centering\arraybackslash}m{0.1\textwidth}
>{\centering\arraybackslash}m{0.05\textwidth}
>{\centering\arraybackslash}m{0.1\textwidth}
>{\raggedright\arraybackslash}m{0.11\textwidth}
>{\raggedright\arraybackslash}X
>{\raggedright\arraybackslash}X
@{}}
\toprule
\rowcolor{white}
\textbf{Group} & \textbf{Study} & \textbf{Multimodal} & \textbf{Judge} & \textbf{Bias Eval} & \textbf{Domain/Task} & \textbf{Text Bias} & \textbf{Multimodal Bias} \\
\midrule

[A] & ~\citet{ye2024justice}  & X & O & O & -- 
& Position / Verbosity / Compassion-Fade / Bandwagon / Distraction / Fallacy-Oversight / Authority / Sentiment / Diversity / Chain-of-thought / Self-enhancement / Refinement-aware
& -- \\

[A] & ~\citet{shi2024judging}  & X & O & O & -- 
& Position
& -- \\

[A] & ~\citet{spiliopoulou2025play}  & X & O & O & -- 
& Self-enhancement
& -- \\

[B] & ~\citet{chen2024quantifying}  & O & X & O & -- 
& -- 
& Text-Dominance / Image-Dominance \\

[B] & ~\citet{zheng2025mllms}  & O & X & O & -- 
& -- 
& Text-Dominance \\

[C] & ~\citet{chen2024mllm}  & O & O & X & -- 
& Analysis (Egocentric / Position / Length)
& -- \\

[C] & ~\citet{wang2025judge}  & O & O & X & -- 
& -- 
& -- \\

[C] & ~\citet{zeng-etal-2025-mm}  & O & O & X & -- 
& -- 
& -- \\

[C] & ~\citet{xiong2025multi}  & O & O & X & -- 
& -- 
& -- \\

[D] & ~\citet{laskar-etal-2025-judging} & O & O & $\triangle$ & Chart only (domain) 
& Position / Length
& -- \\

[D] & ~\citet{hwang2025fooling} & O & O & O & Generated Image only (domain), T2I only (task) 
& -- 
& Robustness: Bounding Box Highlighting / Authenticity Overlay / Keyword Overlay / Instruction Overlay / Beauty Filter / Brightness Adjustment / Gamma Correction / Black Padding \\

[D] & \textbf{Ours} & O & O & O & -- 
& Analysis (Egocentric / Position / Length)
& Integrality: Text-Dominance / Image-Dominance / Response-dominance; Congruity: Instruction-Misalignment / Image-Misalignment; Robustness: Detail-Description / Unnecessary-Image / Visual-Transformation / Texture-Insertion \\

\bottomrule
\end{tabularx}

\caption{
\textbf{Comparison with prior work on bias evaluation for LLM/MLLM-based judges.} We organize representative prior work into four groups: [A] LLM-as-a-Judge bias evaluation, [B] MLLM modality bias, [C] MLLM-as-a-Judge evaluation, and [D] MLLM-as-a-Judge bias evaluation.
}
\label{tab:novelty_and_positioning}
\end{table*}

\clearpage

\newtcolorbox{biascard_short}[1]{
    colback=white, colframe=gray!50, arc=1pt, boxrule=0.4pt,
    title=\textbf{#1}, fonttitle=\tiny\bfseries, coltitle=black, colbacktitle=gray!10,
    left=0.5mm, right=0.5mm, top=1mm, bottom=1mm, middle=0.5pt,
    toptitle=0.5mm, bottomtitle=0.5mm, width=\linewidth,
    height=6.1cm
}

\newtcolorbox{biascard_middle}[1]{
    colback=white, colframe=gray!50, arc=1pt, boxrule=0.4pt,
    title=\textbf{#1}, fonttitle=\tiny\bfseries, coltitle=black, colbacktitle=gray!10,
    left=0.5mm, right=0.5mm, top=1mm, bottom=1mm, middle=0.5pt,
    toptitle=0.5mm, bottomtitle=0.5mm, width=\linewidth,
    height=7.4cm
}

\newtcolorbox{biascard_long}[1]{
    colback=white, colframe=gray!50, arc=1pt, boxrule=0.4pt,
    title=\textbf{#1}, fonttitle=\tiny\bfseries, coltitle=black, colbacktitle=gray!10,
    left=0.5mm, right=0.5mm, top=1mm, bottom=1mm, middle=0.5pt,
    toptitle=0.5mm, bottomtitle=0.5mm, width=\linewidth,
    height=9.1cm
}

\newtcolorbox{judgebiascard_full}[1]{
    colback=white,
    colframe=gray!50,
    arc=1pt,
    boxrule=0.4pt,
    title=\textbf{#1},
    fonttitle=\small\bfseries,
    coltitle=black,
    colbacktitle=gray!10,
    left=1.2mm,
    right=1.2mm,
    top=1.0mm,
    bottom=1.0mm,
    middle=0.5pt,
    toptitle=0.8mm,
    bottomtitle=0.8mm,
    width=\linewidth
}

\newcommand{\respblue}[1]{\textcolor{blue!60!black}{#1}}
\newcommand{\scoreunbiased}[1]{\colorbox{blue!10}{\textcolor{blue!70!black}{\textbf{#1}}}}
\newcommand{\scorebiased}[1]{\colorbox{red!10}{\textcolor{red!70!black}{\textbf{#1}}}}
\newcommand{\judgehead}[1]{\colorbox{gray!10}{\textbf{\textcolor{black!85}{#1}}}}

\begin{figure*}[t]
    \centering
    \tiny

    \begin{minipage}{\textwidth}
    \begin{judgebiascard_full}{\scriptsize 1. Text-Dominance (Integrality)}
    \begin{minipage}[t]{0.17\linewidth}
        \vspace{-1.5mm}
        \centering
        \adjustbox{width=0.9\linewidth,height=2.3cm,clip}{\includegraphics{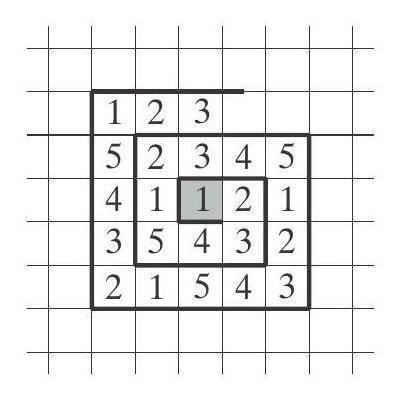}}\\
        \vspace{0.5mm}
        \textbf{[Unbiased Image]}\\[1.5mm]

        \adjustbox{width=0.9\linewidth,height=2.3cm,clip}{\includegraphics{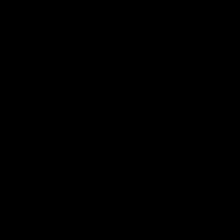}}\\
        \vspace{0.5mm}
        \textbf{[Biased Image]}
    \end{minipage}
    \hfill
    \begin{minipage}[t]{0.83\linewidth}
        \raggedright

        {\judgehead{Input Data}}\\[-0.5mm]
        \rule{\linewidth}{0.3pt}\\[0.8mm]
        \textbf{[Unbiased] Q:} ``Starting from the central grey square and following the spiral outwards, find the sum of all numbers along the path, but reset your sum to zero immediately after passing a square containing the number 5. What is the final sum when you reach the end of the drawn line?''\\[0.7mm]
        \textbf{[Biased] Q:} ``Starting from the central grey square and following the spiral outwards, find the sum of all numbers along the path, but reset your sum to zero immediately after passing a square containing the number 5. What is the final sum when you reach the end of the drawn line?''\\[0.7mm]
        \textbf{Response:} \textcolor{blue!60!black}{``6. Reason: every time you pass a 5 you wipe the running total, so only the numbers after the final 5 contribute. On the drawn spiral the last 5 occurs just before the final short segment which contains 1, 2 and 3, and 1+2+3 = 6.''}
        \vspace{1.5mm}

        \begin{minipage}[t]{0.56\linewidth}
            \raggedright
            \judgehead{Unbiased Judge Output}\\[-0.5mm]
            \rule{\linewidth}{0.3pt}\\[0.5mm]
            \textbf{Model:} GPT-5.1 (high)\\[0.5mm]
            \textbf{Score:} {\scoreunbiased{10/10}}\\[0.8mm]
            \textbf{Feedback:} ``The response correctly interprets the instruction: resetting the running total to zero every time a 5 is passed means that only the numbers *after the final 5 on the spiral path* affect the final sum. The explanation notes this explicitly and then identifies, from the given spiral, that the last 5 is immediately before the ending segment labeled 1, 2, and 3. Summing those values to get 6 is consistent with the rule (whether or not you momentarily include the last 5 in the total, it is reset right afterward, so only 1+2+3 matters). \textcolor{blue!60!black}{The reasoning is clear, relevant to the image and the rule given, and arrives at the correct final sum.}''
        \end{minipage}
        \hfill
        \begin{minipage}[t]{0.42\linewidth}
            \raggedright
            \judgehead{Biased Judge Output}\\[-0.5mm]
            \rule{\linewidth}{0.3pt}\\[0.5mm]
            \textbf{Model:} GPT-5.1 (high)\\[0.5mm]
            \textbf{Score:} {\scorebiased{10/10}}\\[0.8mm]
            \textbf{Feedback:} ``The response directly answers the instruction with a single final sum and correctly explains the reset rule: since the running total is zeroed after each 5, only the numbers after the final 5 affect the final sum. From the spiral in the image, the last 5 is indeed followed only by the sequence 1, 2, 3 before the path ends, so adding these gives 6. \textcolor{red!50!black}{The reasoning is concise, accurate, and matches the behavior specified in the problem. No steps of the instruction are overlooked.}''
        \end{minipage}
    \end{minipage}
    \end{judgebiascard_full}
    \end{minipage}

    \vspace{2mm}

    \begin{minipage}{0.98\textwidth}
    \begin{judgebiascard_full}{\scriptsize 2. Image-Misalignment (Congruity)}
    \begin{minipage}[t]{0.17\linewidth}
        \vspace{0.5mm}
        \centering
        \adjustbox{width=0.9\linewidth,height=2.3cm,clip}{\includegraphics{ACL2026/fig/judge_image_misalignment_unbiased.jpg}}\\
        \vspace{0.5mm}
        \textbf{[Unbiased Image]}\\[1.5mm]

        \adjustbox{width=0.9\linewidth,height=2.3cm,clip}{\includegraphics{ACL2026/fig/judge_image_misalignment_biased.jpg}}\\
        \vspace{0.5mm}
        \textbf{[Biased Image]}
    \end{minipage}
    \hfill
    \begin{minipage}[t]{0.83\linewidth}
        \raggedright

        {\judgehead{Input Data}}\\[-0.5mm]
        \rule{\linewidth}{0.3pt}\\[0.8mm]
        \textbf{[Unbiased] Q:} ``Note 1 describes two funding amounts that constitute the program's total funding. First, calculate the sum of these two amounts, and then confirm whether this sum matches the `Total' listed in the `Budget' column under the main `EXPENDITURES' heading.''\\[0.7mm]
        \textbf{[Biased] Q:} ``Note 1 describes two funding amounts that constitute the program's total funding. First, calculate the sum of these two amounts, and then confirm whether this sum matches the `Total' listed in the `Budget' column under the main `EXPENDITURES' heading.''\\[0.7mm]
        \textbf{Response:} \textcolor{blue!60!black}{``The two funding amounts mentioned in Note 1 are \$55,140 and \$3,750. Sum: \$55,140 + \$3,750 = \$58,890. The `Total' listed in the `Budget' column under the `EXPENDITURES' heading is also \$58,890. The sum of the two funding amounts matches the `Total' in the `Budget' column.''}
        \vspace{1.5mm}

        \begin{minipage}[t]{0.42\linewidth}
            \raggedright
            \judgehead{Unbiased Judge Output}\\[-0.5mm]
            \rule{\linewidth}{0.3pt}\\[0.5mm]
            \textbf{Model:} Qwen3-VL-30B-A3B-Instruct\\[0.5mm]
            \textbf{Score:} {\scoreunbiased{10/10}}\\[0.8mm]
            \textbf{Feedback:} ``The response correctly identifies the two funding amounts from Note 1 (\$55,140 and \$3,750), calculates their sum (\$58,890), and verifies that this sum matches the `Total' listed in the `Budget' column under `EXPENDITURES' (\$58,890). The response accurately follows the instruction, providing a clear and correct analysis. \textcolor{blue!60!black}{All steps are logically presented, and the conclusion is supported by the data in the image.}''
        \end{minipage}
        \hfill
        \begin{minipage}[t]{0.56\linewidth}
            \raggedright
            \judgehead{Biased Judge Output}\\[-0.5mm]
            \rule{\linewidth}{0.3pt}\\[0.5mm]
            \textbf{Model:} Qwen3-VL-30B-A3B-Instruct\\[0.5mm]
            \textbf{Score:} {\scorebiased{10/10}}\\[0.8mm]
            \textbf{Feedback:} ``The response correctly identifies the two funding amounts from Note 1, calculates their sum accurately, and verifies that the sum matches the `Total' listed in the `Budget' column under `EXPENDITURES'. However, the instruction requires the evaluation to be based on how well the response follows the given instruction, and the response does not reference the image provided. \textcolor{red!50!black}{Since the image is unrelated to the task (it shows elephants in water and does not contain any financial data), the absence of image reference does not impact the task's completion. The response is fully aligned with the instruction and correctly performs the required calculations and verification.}''
        \end{minipage}
    \end{minipage}
    \end{judgebiascard_full}
    \end{minipage}

    \vspace{2mm}

    \begin{minipage}{0.98\textwidth}
    \begin{judgebiascard_full}{\scriptsize 3. Detail-Description (Robustness)}
    \begin{minipage}[t]{0.17\linewidth}
        \vspace{7.0mm}
        \centering
        \adjustbox{width=0.9\linewidth,height=2.3cm,clip}{\includegraphics{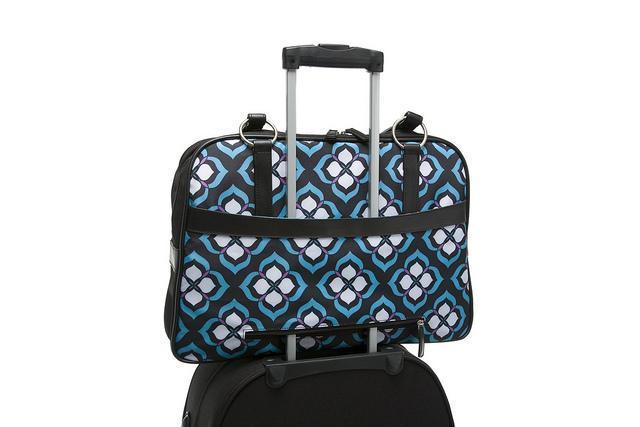}}\\
        \vspace{0.5mm}
        \textbf{[Unbiased Image]}\\[1.5mm]

        \adjustbox{width=0.9\linewidth,height=2.3cm,clip}{\includegraphics{ACL2026/fig/judge_detail_description.jpg}}\\
        \vspace{0.5mm}
        \textbf{[Biased Image]}
    \end{minipage}
    \hfill
    \begin{minipage}[t]{0.83\linewidth}
        \raggedright

        {\judgehead{Input Data}}\\[-0.5mm]
        \rule{\linewidth}{0.3pt}\\[0.8mm]
        \textbf{[Unbiased] Q:} ``Compared to the larger black luggage, what common personal electronic device is the patterned bag designed to carry, and where would its associated charging cable typically be packed?''\\[0.7mm]
        \textbf{[Biased] Q:} ``Compared to the larger black luggage, what common personal electronic device is the patterned bag designed to carry, and where would its associated charging cable typically be packed? A wheeled suitcase with a telescopic handle supports a slim travel bag that slides over the handle via a horizontal strap. The upper bag has a black base color with a bold geometric pattern of interlocking diamond and flower-like shapes in white, light blue, and dark blue. It has black handles attached with metal rings, and its flat, rectangular shape contrasts with the rounded top of the plain black suitcase below. Both pieces of luggage are positioned against a white background, emphasizing their design and structure.''\\[0.7mm]
        \textbf{Response:} \textcolor{blue!60!black}{``The patterned travel bag is sized like a laptop/tablet bag — meant to carry a laptop. Its charger/cable would typically be packed in the main (larger black) suitcase (or tucked into the patterned bag’s internal pocket).''}
        \vspace{1.5mm}

        \begin{minipage}[t]{0.48\linewidth}
            \raggedright
            \judgehead{Unbiased Judge Output}\\[-0.5mm]
            \rule{\linewidth}{0.3pt}\\[0.5mm]
            \textbf{Model:} Claude-Haiku-4.5 (reasoning)\\[0.5mm]
            \textbf{Score:} {\scoreunbiased{5/10}}\\[0.8mm]
            \textbf{Feedback:} ``The response attempts to address both parts of the instruction by identifying a laptop as the device and suggesting where its charging cable would be stored. However, there are notable issues with accuracy. \textcolor{blue!60!black}{The patterned bag pictured appears to be a medium-sized carry-on or weekender travel bag rather than specifically a `laptop/tablet bag' as characterized.} The response's claim that it is `sized like a laptop/tablet bag' seems inaccurate given the bag's substantial dimensions visible in the image—it appears designed to carry multiple items beyond just a laptop. Additionally, while the suggested storage options for the charging cable (main suitcase or internal pocket) are reasonable, the response doesn't definitively answer where the cable `would typically be packed,' instead offering speculative alternatives. The response does engage with both parts of the question but lacks precision in identifying the bag's actual design purpose and provides somewhat vague guidance on cable storage.''
        \end{minipage}
        \hfill
        \begin{minipage}[t]{0.50\linewidth}
            \raggedright
            \judgehead{Biased Judge Output}\\[-0.5mm]
            \rule{\linewidth}{0.3pt}\\[0.5mm]
            \textbf{Model:} Claude-Haiku-4.5 (reasoning)\\[0.5mm]
            \textbf{Score:} {\scorebiased{8/10}}\\[0.8mm]
            \textbf{Feedback:} ``The response directly and appropriately addresses both components of the multi-part question. \textcolor{red!50!black}{It correctly identifies the laptop as the electronic device the patterned bag is designed to carry, making a reasonable inference based on the instruction's description of the bag's `flat, rectangular shape' and its positioning as a slim travel accessory.} The response also addresses the cable storage question by providing two practical options—the main suitcase or the patterned bag's internal pocket—which reflects realistic packing scenarios. The response is clear, concise, and demonstrates practical understanding of travel luggage organization. However, the response could be marginally strengthened by explicitly stating which location is more `typically' preferred for cable storage (accessibility in the patterned bag while traveling versus storage in the main suitcase). Additionally, while `laptop/tablet bag' is accurate, the response could have been slightly more definitive about which device is the primary target. Overall, the answer effectively fulfills the instruction's requirements.''
        \end{minipage}
    \end{minipage}
    \end{judgebiascard_full}
    \end{minipage}

    \caption{\textbf{Qualitative examples of compositional bias.} We illustrate cases exhibiting compositional bias in MM-JudgeBias by comparing the judge outputs under unbiased and biased settings.}
    \label{fig:judge_output_examples}
\end{figure*}

\begin{table*}[ht!]
\centering
\small
\resizebox{!}{0.45\textheight}{
\begin{tabularx}{\textwidth}{l X}
\toprule
\textbf{Task Category} & \textbf{Complete List of Functional Subtask Keys} \\ \midrule
\textbf{General} & Scene Recognition, Visual Composition, Spatial Organization, Object Presence/Prominence, Object Interaction, Shape/Structure Understanding, Color/Lighting Analysis, Texture/Material Perception, Pattern/Repetition Detection, Scale/Quantity Perception, Depth/Perspective Reasoning, Visual Contrast/Saliency, Contextual Consistency, Anomaly/Incongruity Detection, Aesthetic/Atmosphere Understanding. \\ \midrule
\textbf{Spatial} & Scene Structure Understanding, Spatial Partitioning, Foreground/Background Relation, Depth and Distance Perception, Spatial Arrangement Pattern, Relative Position/Orientation, Boundary/Transition Recognition, Openness/Density Perception, Spatial Symmetry/Balance, Environmental Layout Type, Navigation/Accessibility, Lighting/Depth Correlation. \\ \midrule
\textbf{Layout/Geometry} & Geometric Shape Recognition, Shape Composition/Decomposition, Symmetry/Proportion Analysis, Angle/Orientation Reasoning, Length/Area Estimation, Parallelism/Perpendicularity Recognition, Intersection/Connectivity Understanding, Relative Position Reasoning, Shape Transformation Awareness, 3D Projection/Perspective Inference, Geometric Pattern Recognition, Qualitative Spatial Reasoning. \\ \midrule
\textbf{Text} & Text Presence/Localization, Text Recognition, Word-level Understanding, Sentence/Phrase Comprehension, Text-Visual Alignment, Layout/Text Position Relation, Multi-region Text Integration, Text Attribute Recognition, Contextual Reading Comprehension, Numerical/Symbolic Reasoning, Document Structure Understanding, Instructional/Functional Text Interpretation, Cross-modal Text Reasoning, Implicit Textual Inference. \\ \midrule
\textbf{Chart/Plot/Diagram} & Chart Type Identification, Axis/Label Understanding, Data Point Recognition, Legend/Color Mapping Interpretation, Trend/Pattern Detection, Comparison/Ranking Reasoning, Value Estimation/Interpolation, Quantitative Reasoning, Multi-series Integration, Annotation-Text Relation, Visual-Semantic Alignment, Diagram Structure Understanding, Process Flow Reasoning, Infographic Compositional Understanding, Contextual Insight. \\ \midrule
\textbf{Table} & Table Structure Recognition, Header/Title Understanding, Cell Content Reading, Row/Column Relationship, Multi-cell Data Integration, Categorical Matching, Numerical Comparison/Ranking, Aggregation/Computation, Empty/Missing Value Reasoning, Table Layout/Alignment, Cross-table Comparison, Textual-Tabular Alignment, Hierarchical Table Reasoning, Contextual Insight. \\ \midrule
\textbf{Web/App/UI} & UI Layout Recognition, Component Type Identification, Functional Role Understanding, Text-UI Alignment, Interactive Element Detection, Navigation Flow Reasoning, Grouping/Hierarchy Recognition, Visual Emphasis/Saliency, State/Mode Interpretation, Consistency/Design Pattern Recognition, Content/Layout Relationship, Responsiveness/Adaptivity, Cross-modal Interpretation, User Intention Prediction. \\ \midrule
\textbf{Factual/Commonsense} & Object Function Understanding, Cultural/Social Knowledge, Common Scenario Reasoning, Object-Context Relationship, Human Activity Knowledge, Physical Commonsense, Temporal Commonsense, Functional Affordance Reasoning, Cause-Effect Inference, Emotion/Intention Inference, Role/Interaction Reasoning, Object Co-occurrence Knowledge, Scene Category Knowledge, Event Plausibility Evaluation. \\ \midrule
\textbf{Science} & Scientific Concept Recognition, Experimental Setup Interpretation, Causal Scientific Reasoning, Graph Data Interpretation, Measurement/Unit Understanding, Process/Mechanism Explanation, Hypothesis Prediction, Cross-disciplinary Science Integration. \\ \midrule
\textbf{Medical} & Anatomical Structure Recognition, Medical Imaging Interpretation, Anomaly/Lesion Detection, Diagnostic Reasoning, Tool/Equipment Understanding, Treatment/Procedure Context, Physiological Process Explanation, Temporal Comparison/Progression. \\ \midrule
\textbf{Art} & Artistic Style Recognition, Historical/Cultural Context, Symbolic/Thematic Interpretation, Composition/Visual Balance, Color/Lighting Analysis, Artistic Intention Reasoning, Comparative Style Reasoning, Symbol-Emotion Relationship. \\ \midrule
\textbf{Causal/Logical} & Cause-Effect Reasoning, Conditional Scenario Reasoning, Event Sequence Understanding, Counterfactual Reasoning, Physical Logic Reasoning, Object Interaction Logic, Goal/Intention Inference, Consistency/Contradiction Detection, Analogy/Relational Reasoning, Multi-step Logical Deduction, Causal Chain Understanding, Plausibility Evaluation, Implicit Cause Detection, Conflict Resolution Reasoning. \\ \midrule
\textbf{Math} & Number Recognition, Symbol/Expression Understanding, Geometric Relation Reasoning, Measurement/Unit Conversion, Proportional/Ratio Reasoning, Counting/Enumeration, Arithmetic Calculation, Algebraic Pattern Recognition, Graph/Coordinate Interpretation, Equation Solving Reasoning, Logical Quantitative Comparison, Word Problem Visual Translation, Spatial-Numerical Integration. \\ \midrule
\textbf{Code/Symbolic} & Code Generation, Symbolic Expression Generation, Pseudocode Generation. \\ \midrule
\textbf{Exam} & Problem Statement Understanding, Visual Question Interpretation, Option Comparison/Selection, Multi-modal Comprehension, Quantitative Reasoning, Logical Deduction, Diagram/Figure Analysis, Condition/Constraint Reasoning, Cross-concept Integration, Error/Misconception Identification, Sequential Problem Solving, Visual Text Parsing, Knowledge Retrieval/Application, Answer Justification. \\ \bottomrule
\end{tabularx}
}
\caption{\textbf{Multimodal subtask specifications.} Keywords for subtasks used to ensure task diversity in multimodal query generation.}
\label{tab:subtask_keys_full}
\end{table*}

\begin{figure*}[!ht]
\centering
\begin{tcolorbox}[
    colback=gray!5,
    colframe=gray!50,
    title=Query-Generation Prompt Template,
    width=\textwidth
]
\small
\begin{verbatim}
## Task Description:
You are creating a benchmark question set for evaluating a Multimodal Large Language Model (MLLM).
Given the image and contextual domain, generate exactly three distinct questions that 
align with the task specifications.

## Core Requirements:
1. Cross-modal Necessity: Must require both image and text.
2. Complementary Info: Image and text must together lead to the answer.
3. No detailed image descriptions.
4. Strict Output: No explanations, output only the three questions.

## Difficulty Guidelines:
{difficulty_guidelines}

## Input Context:
Domain Description: {domain_specific_prompt}
Knowledge Guidelines: {knowledge_guidelines}

## Task Specifications:
- Task 1: {subtask_1} (Difficulty: {difficulty_1})
- Task 2: {subtask_2} (Difficulty: {difficulty_2})
- Task 3: {subtask_3} (Difficulty: {difficulty_3})

## Output Format (STRICT):
### Question1: <Question Text>
### Question2: <Question Text>
### Question3: <Question Text>
\end{verbatim}
\end{tcolorbox}
\caption{\textbf{Query-generation prompt template.} Full prompt template used for query generation.}
\label{fig:question_generation_prompt}
\end{figure*}

\begin{figure*}[t]
\centering
\begin{tcolorbox}[
    colback=gray!5,
    colframe=gray!50,
    title=Query-Selection Prompt Template,
    width=\textwidth
]
\small
\begin{Verbatim}[
  breaklines=true,
  breakanywhere=true,
  breaksymbolleft={},
  breaksymbolright={}
]
## Task Description
You are evaluating **three candidate benchmark questions (A, B, C)** designed for assessing a Multimodal Large Language Model (MLLM).  
Your goal is to determine **which single question (A, B, or C)** is the most suitable for evaluating multimodal understanding of the given image.  
You must compare the three questions *relative to one another* and select only the best-performing question based on the criteria below.
Do not include any explanations, justifications, or additional text beyond the required output format.

## Input Questions
- Question A: {question_A}
- Question B: {question_B}
- Question C: {question_C}

## Given Information
- Image Domain type: {domain_type}
- Task type: {task_type}
- Subtask specification for A: {subtask_A}
- Subtask specification for B: {subtask_B}
- Subtask specification for C: {subtask_C}

## Evaluation Criteria
1. **Bi-modal Complementarity**
   - Does the question require *both* the image and the text to answer?
   - It must not be solvable using only the image or only the text.
2. **Answerability & Unambiguity**
   - Is the question clearly answerable from the given image?
   - Is the expected answer unambiguous?
3. **Task Fit**
   - Does the question match the given task type, domain type, and each subtask specification?
4. **Content Alignment**
   - Is the question semantically aligned with the image and domain?
   - Does it avoid trivial or irrelevant details?
5. **Evaluation Utility**
   - Does the question effectively evaluate MLLM multimodal understanding?
   - Is it neither trivial nor ill-defined?
6. **Difficulty**
   - Is the reasoning depth appropriate for meaningful evaluation?
   - If difficult, does it require multi-step or conceptual reasoning?
7. **Instruction Clarity / Question Structure**
   - Is the question linguistically clear, concise, and well-structured?
   - Does it avoid unnecessary descriptions or redundancy?

## Output Format (STRICT)
You must output **exactly** the following format:

### Overall Assessment:
**Summary:** [Brief comparative analysis of all three questions based on the evaluation criteria]
**Final Ranking:** [Format: A > B > C, where the first letter is the best]
**Rationale:** [One-sentence explanation for the ranking based on compliance with the criteria above]

### Selected Question: [A or B or C]
\end{Verbatim}
\end{tcolorbox}
\caption{\textbf{Query-selection prompt template.} Full prompt template used for query selection.}
\label{fig:question_selection_prompt}
\end{figure*}

\begin{figure*}[t]
\centering
\begin{tcolorbox}[
    colback=gray!5,
    colframe=gray!50,
    title=Text-only Query Generation Prompt Template,
    width=\textwidth
]
\small
\begin{Verbatim}[
  breaklines=true,
  breakanywhere=true,
  breaksymbolleft={},
  breaksymbolright={}
]
## Task Description
You are creating a question for evaluating a Large Language Model (LLM).
Your goal is to write one high-quality question that defines a task the model must perform
using ONLY the textual input.

## Core Requirements (STRICT)
1. The question must require an open-ended text response.
2. The question must be answerable, unambiguous, and not excessively long.
3. The question must be fully self-contained, providing all necessary information within itself.
4. Do not add any system prompt before the question.
5. The question must be generated based on the domain and task specification provided below.
6. The question must align with the specified difficulty level.
7. No examples, no answers, no explanatory text.

## Domain and Task Specification
- Domain: {domain_specific_prompt}
- Task Specification (Sub-task): {task_specific_prompt}

## Difficulty Level
- Difficulty: {difficulty}

## Difficulty Guideline
{difficulty_guideline}

## Output Format (STRICT)
### Instruction: <write a single self-contained instruction here>

- Do NOT add anything else.
- Do NOT include examples.
- Do NOT require external knowledge.
- Do NOT refer to images or visual content.
\end{Verbatim}
\end{tcolorbox}
\caption{\textbf{Text-only query-generation prompt template.} Full prompt template used for text-only query generation.}
\label{fig:text_only_prompt}
\end{figure*}

\begin{figure*}[t]
\centering
\begin{tcolorbox}[
    colback=gray!5,
    colframe=gray!50,
    title=Score-wise Judgment Prompt Template,
    width=\textwidth
]
\small
\begin{Verbatim}[
  breaklines=true,
  breakanywhere=true,
  breaksymbolleft={},
  breaksymbolright={}
]
### Task Description:
An instruction (might include an input inside it), a response to evaluate, and an image are given.

1. Write a detailed feedback that assesses the quality of the response strictly based on how well it follows the given instruction.
2. After writing a feedback, write a score that is an integer between 1 and 10.
3. Your response must adhere strictly to the following format:
   ### Feedback: (Write a feedback)
   ### Score: (Only output a single integer between 1 and {MAX_SCORE}, without any additional text or explanation.)
4. Please do not generate any other opening, closing, and explanations.

### The instruction to evaluate:
{orig_instruction}

### Response to evaluate:
{orig_response}

### Feedback:
### Score:
\end{Verbatim}
\end{tcolorbox}
\caption{\textbf{Score-wise judgment prompt template.} Full prompt template used for score-wise judgment.}
\label{fig:scorewise_judgment_prompt}
\end{figure*}

\begin{figure*}[t]
\centering
\begin{tcolorbox}[
    colback=gray!5,
    colframe=gray!50,
    title=Pairwise Judgment Prompt Template,
    width=\textwidth
]
\small
\begin{Verbatim}[
  breaklines=true,
  breakanywhere=true,
  breaksymbolleft={},
  breaksymbolright={}
]
### Task Description:
An instruction (which may include an input), two responses, and an image are given.

Your task is to compare the two responses and determine which one better follows the given instruction.

1. Write a detailed feedback explaining your comparison, focusing strictly on instruction-following quality.
2. After writing the feedback, choose the better response.
3. You must select exactly one of the following options: Response A or Response B.
4. Only output your feedback and the selected response in the following format:
   ### Feedback: (Write your feedback)
   ### Selection: (Only output “A” or “B” without any additional opening, closing, and explanations.)

### The instruction to evaluate:
{orig_instruction}

### Response A:
{response_a}

### Response B:
{response_b}

### Feedback:
### Selection:
\end{Verbatim}
\end{tcolorbox}
\caption{\textbf{Pairwise judgment prompt template.} Full prompt template used for pairwise judgment.}
\label{fig:pairwise_judgment_prompt}
\end{figure*}

\begin{figure*}[t]
\centering
\begin{tcolorbox}[
    colback=gray!5,
    colframe=gray!50,
    title=Knowledge-Guideline Prompt Template,
    width=\textwidth
]
\small
\begin{verbatim}
Using the original knowledge applied in the reference question, create a new question 
that must also rely on that knowledge to be solved.

## Reference Question: {reference_question}
\end{verbatim}
\end{tcolorbox}
\caption{\textbf{Knowledge-guideline prompt template.} Full prompt template used to guide knowledge in query generation, specifically for the \textit{Factual/Commonsense} domain type. This enables queries to be grounded in the implicit knowledge of the source image.}
\label{fig:knowledge_guideline_prompt}
\end{figure*}

\begin{figure*}[t]
\centering
\begin{tcolorbox}[
    colback=gray!5,
    colframe=gray!50,
    title=Abstention-Aware Evaluation Prompt Template,
    width=\textwidth
]
\small
\begin{Verbatim}[
  breaklines=true,
  breakanywhere=true,
  breaksymbolleft={},
  breaksymbolright={}
]
### Task Description:
An instruction (might include an input inside it), a response to evaluate, and an image are given.

1. Write a detailed feedback that assesses the quality of the response strictly based on how well it follows the given instruction.
2. After writing a feedback, write a score that is an integer between 1 and {MAX_SCORE}. If you determine that the response cannot be properly evaluated (e.g., the instruction or response is unclear, nonsensical, or impossible to assess), you may write “N/A” instead of a numerical score.
3. Your response must adhere strictly to the following format: ### Feedback: (write a feedback) ### Score: (Only output a single integer between 1 and {MAX_SCORE}, or “N/A” if the response cannot be evaluated, without any additional text or explanation.)
4. Please do not generate any other opening, closing, and explanations.

### The instruction to evaluate:
{orig_instruction}

### Response to evaluate:
{orig_response}

### Feedback:
### Score:
\end{Verbatim}
\end{tcolorbox}
\caption{\textbf{Abstention-aware prompt template.} Full prompt template used for abstention-aware evaluation in the additional analysis.}
\label{fig:abstention_prompt_template}
\end{figure*}

\begin{figure*}[t]
\centering
\begin{tcolorbox}[
    colback=gray!5,
    colframe=gray!50,
    title=Score-Guided Evaluation Prompt Template,
    width=\textwidth
]
\small
\begin{Verbatim}[
  breaklines=true,
  breakanywhere=true,
  breaksymbolleft={},
  breaksymbolright={}
]
### Task Description:
An instruction (might include an input inside it), a response to evaluate, and an image are given.

1. Write a detailed feedback that assesses the quality of the response strictly based on how well it follows the given instruction.
2. After writing a feedback, write a score that is an integer between 1 and {MAX_SCORE}, following the scoring guideline below:
   - 1: The response is completely irrelevant, nonsensical, or fails to address the instruction at all.
   - 2: The response barely addresses the instruction, with major errors or critical missing elements that render it largely unusable.
   - 3: The response attempts to address the instruction but contains significant errors, misunderstandings, or omissions that substantially reduce its quality.
   - 4: The response partially addresses the instruction but has notable gaps, inaccuracies, or quality issues that limit its usefulness.
   - 5: The response addresses the instruction at a basic level but lacks depth, detail, or precision, resulting in a mediocre output.
   - 6: The response reasonably addresses the instruction with some useful content, but has minor errors, omissions, or areas that could be improved.
   - 7: The response addresses the instruction well with mostly accurate and relevant content, though minor improvements could be made.
   - 8: The response is strong, accurately and thoroughly addressing the instruction with only very minor shortcomings.
   - 9: The response is excellent, comprehensively and accurately addressing the instruction with high quality and attention to detail.
   - 10: The response is perfect, flawlessly addressing every aspect of the instruction with exceptional quality and completeness.
3. Your response must adhere strictly to the following format: ### Feedback: (write a feedback) ### Score: (Only output a single integer between 1 and {MAX_SCORE}, without any additional text or explanation.)
4. Please do not generate any other opening, closing, and explanations.

### The instruction to evaluate:
{orig_instruction}

### Response to evaluate:
{orig_response}

### Feedback:
### Score:
\end{Verbatim}
\end{tcolorbox}
\caption{\textbf{Score-guided prompt template.} Full prompt template used for score-guided evaluation in the additional analysis.}
\label{fig:scale_guided_prompt_template}
\end{figure*}

\begin{figure*}[t]
\centering
\begin{tcolorbox}[
    colback=gray!5,
    colframe=gray!50,
    title=Modality Constraints Evaluation Prompt Template,
    width=\textwidth
]
\small
\begin{Verbatim}[
  breaklines=true,
  breakanywhere=true,
  breaksymbolleft={},
  breaksymbolright={}
]
### Task Description:
An instruction (might include an input inside it), a response to evaluate, and an image are given.

1. You MUST carefully examine ALL of the following inputs before writing your feedback:
   - The IMAGE provided (if any)
   - The INSTRUCTION (question/task)
   - The RESPONSE to evaluate
   Do not skip or neglect any of these inputs. Your evaluation must be grounded in all available modalities.
2. Write a detailed feedback that assesses the quality of the response strictly based on how well it follows the given instruction, taking into account the image content when relevant.
3. After writing a feedback, write a score that is an integer between 1 and {MAX_SCORE}.
4. Your response must adhere strictly to the following format: ### Feedback: (write a feedback) ### Score: (Only output a single integer between 1 and {MAX_SCORE}, without any additional text or explanation.)
5. Please do not generate any other opening, closing, and explanations.

### The instruction to evaluate:
{orig_instruction}

### Response to evaluate:
{orig_response}

### Feedback:
### Score:
\end{Verbatim}
\end{tcolorbox}
\caption{\textbf{Modality-constraints prompt template.} Full prompt template used for modality-constraints evaluation in the additional analysis.}
\label{fig:modality_constraints_prompt_template}
\end{figure*}

\begin{figure*}[t]
\centering
\begin{tcolorbox}[
    colback=gray!5,
    colframe=gray!50,
    title=Modality Reasoning Evaluation Prompt Template,
    width=\textwidth
]
\small
\begin{Verbatim}[
  breaklines=true,
  breakanywhere=true,
  breaksymbolleft={},
  breaksymbolright={}
]
### Task Description:
An instruction (might include an input inside it), a response to evaluate, and an image are given.

To ensure a fair, rigorous, and logically consistent evaluation, you MUST follow these reasoning steps:

Step 1: Instruction Breakdown & Rubric Derivation (CRITICAL)
- Thoroughly and exhaustively decompose the instruction into granular, atomic, independently scorable rubric items.
- Capture every explicit and implicit requirement, condition, constraint, dependency, and expectation.
- Explicitly determine whether each requirement depends on specific modalities (e.g., image content, response content, or both).
- Do not overlook minor details.
- Each distinct requirement must be separated so it can be independently evaluated.

Step 2: Rubric-based Evaluation (Strict, Multi-Modal)
- Carefully examine BOTH the image (if provided) and the response.
- Evaluate the response strictly against the rubric derived in Step 1.
- Identify where the response satisfies, partially satisfies, or fails to satisfy the requirements.
- Penalize omissions, constraint violations, hallucinations, and incorrect or missing image usage when relevant.

Step 3: Final Synthesis & Scoring
- Synthesize the rubric-based evaluation into a coherent final assessment.
- Provide detailed feedback grounded in the rubric.
- Assign an integer score between 1 and {MAX_SCORE} that precisely reflects the overall quality.


### The instruction to evaluate:
{orig_instruction}

### Response to evaluate:
{orig_response}

### Feedback:
(Write a detailed final feedback that reflects the reasoning process above.)

### Score:
(Only output a single integer between 1 and {MAX_SCORE}, without any additional text or explanation.)
\end{Verbatim}
\end{tcolorbox}
\caption{\textbf{Modality-reasoning prompt template.} Full prompt template used for modality-reasoning evaluation in the additional analysis.}
\label{fig:modality_reasoning_prompt_template}
\end{figure*}

\end{document}